\documentclass[twoside,11pt]{article}
\usepackage{subfigure}
\usepackage{float}
\usepackage[all]{xy}
\usepackage{enumitem}
\usepackage{algorithmic} 
\setlist[enumerate]{itemsep=0mm}
\usepackage{color}
\usepackage{jmlr2e}
\usepackage{graphics}
\usepackage{epsfig}
\usepackage[ruled]{algorithm2e}
\usepackage{booktabs}
\usepackage{url}
\usepackage{psfrag}
\usepackage{relsize}
\usepackage{amsmath,amsfonts,mathrsfs,mathtools,amssymb}
\usepackage[colorlinks,linkcolor=blue,anchorcolor=green,citecolor=green]{hyperref}
\usepackage[citecolor=green]{hyperref}
\usepackage{tikz,pgfplots}
\usepackage{tkz-tab}
\usepackage[format=hang,justification=justified,singlelinecheck=false,font=footnotesize]{caption}
\usepackage[T1]{fontenc}
\usepackage[utf8]{inputenc}
\usepackage[font=small,labelfont=bf,tableposition=top]{caption}
\usepackage{booktabs}
\usepackage{threeparttable} 
\usepackage{multirow}

\newcommand{\eins}{\boldsymbol{1}}
\DeclareSymbolFont{wideparensymbol}{OMX}{yhex}{m}{n}
\DeclareMathAccent{\wideparen}{\mathord}{wideparensymbol}{"F3}

\newcommand{\argmin}{\operatornamewithlimits{arg \, min}}

\makeatletter

\newcommand{\Rmnum}[1]{\expandafter\@slowromancap\romannumeral #1@}
\makeatother

\jmlrheading{}{}{~}{~}{~}{2021 Hanyuan Hang, Tao Huang, Yuchao Cai, Hanfang Yang, and Zhouchen Lin}

\usepackage{tikz,pgfplots}
\pgfplotsset{compat=newest}
\pgfplotsset{plot coordinates/math parser=false,trim axis left}
\newlength\figureheight
\newlength\figurewidth

\begin{document}

\title{Gradient Boosted Binary Histogram Ensemble \\ for Large-scale Regression}

\author{\name Hanyuan Hang \email h.hang@utwente.nl \\
\addr Department of Applied Mathematics, University of Twente \\
7522 NB Enschede, The Netherlands
\vspace{0.2cm}
\\
\name Tao Huang \email tao.huang2018@ruc.edu.cn \\ 
\addr Institute of Statistics and Big Data,
Renmin University of China \\
100872 Beijing, China 
\vspace{0.2cm}
\\
\name Yuchao Cai \email  yccai@ruc.edu.cn\\
\name Hanfang Yang \email  hyang@ruc.edu.cn \\
\addr School of Statistics, Renmin University of China \\
100872 Beijing, China 
\vspace{0.2cm}
\\
\name Zhouchen Lin \email zlin@pku.edu.cn \\
\addr Key Lab.~of Machine Perception (MOE), School of EECS,
Peking University \\
100871 Beijing, China
}


\allowdisplaybreaks

\maketitle

\begin{abstract}In this paper, we propose a gradient boosting algorithm for large-scale regression problems called \textit{Gradient Boosted Binary Histogram Ensemble} (GBBHE) based on binary histogram partition and ensemble learning. From the theoretical perspective, by assuming the H\"{o}lder continuity of the target function, we establish the statistical convergence rate of GBBHE in the space $C^{0,\alpha}$ and $C^{1,0}$, where a lower bound of the convergence rate for the base learner demonstrates the advantage of boosting. Moreover, in the space $C^{1,0}$, we prove that the number of iterations to achieve the fast convergence rate can be reduced by using ensemble regressor as the base learner, which improves the computational efficiency. In the experiments, compared with other state-of-the-art algorithms such as gradient boosted regression tree (GBRT), Breiman's forest, and kernel-based methods, our GBBHE algorithm shows promising performance with less running time on large-scale datasets.
\end{abstract}


\section{Introduction} \label{sec::Introduction}

Over the past two decades, boosting has become one of the most successful algorithms in the machine learning community \citep{buhlmann2003boosting}. After the idea of iterative utilization of \textit{weak learners} from a certain function space to generate a strong one, which is called \textit{boosting}, first came out in \citet{schapire1990strength,freund1995boosting}, it gains a lot of attention, and a wealth of literature has applied it to a large number of datasets. During this period, many boosting algorithms with impressive performance have been proposed. The first boosting algorithm dates back to the Adaboost for classification by \citet{schapire1995decision,freund1997decision}. Another important boosting algorithm for regression called \textit{Gradient Boosted Regression Tree} (GBRT) was proposed by \citet{friedman2001greedy}. GBRT takes advantage of tree-based learners to capture complex data structures.

In addition to the great success of these boosting algorithms, a lot of attempts have been made to establish their theoretical foundations. For example, from the perspective of statistical analysis, \citet{buhlmann2003boosting} derived an exponential bias-variance trade-off for linear regression to illustrate the almost resistance to overfitting for $L_2$-boosting in a fixed design setting. Moreover, \citet{park2009l2} and \citet{lin2019boosted} established the theoretical analysis of boosting methods using Nadaraya-Watson kernel estimates and kernel ridge regression estimates as base learners, respectively. However, these methods are of little value in practice since they fail to capture the complex data dependencies in applications. In the prior work of this paper \citep{cai2020boosted}, we proposed a boosting algorithm for regression problems called boosted histogram transform for regression (BHTR) with sound theoretical guarantees and satisfactory empirical performance. However, since it utilizes the ordinary histogram transforms as base learners, the number of partition cells grows exponentially with the dimension, and thus is hard to adapt to high-dimensional data.

Moreover, despite the success in achieving desirable performance, boosting procedures can encounter heavy computational costs in the computations of the features and the selection of the weak learner, both of which depend on the number of features and the number of training examples. Therefore, boosting algorithms cannot be directly applied to large-scale high-dimensional data. Efforts have been made to further enhance the efficiency of boosting algorithms in large-scale scenarios. For example, \citet{dubout2014adaptive} utilized an adaptive sampling approach to sample features at every boosting step, so as to reduce computational cost. Other works focus on enhancing the efficiency of boosting through engineering optimizations: \citet{chen2016xgboost} came up with the \textit{eXtreme Gradient Boosting} (XGBoost), which achieves excellent experimental performance. \citet{ke2017lightgbm} speeds up the training process of conventional GBRT by up to over 20 times. More recently, \citet{biau2019accelerated} incorporates the Nesterov’s accelerated gradient descent technique \citep{nesterov27method} to accelerate GBRT. Unfortunately, none of the above-mentioned boosting works for large-scale regression presents a satisfactory statistical theoretical guarantee.

Under such background, this paper aims to establish a new boosting algorithm for large-scale and high-dimensional regression, which is not only with satisfactory performance but also with solid theoretical foundations. 
To be specific, motivated by the random rotation ensemble algorithms \citep{lopez2013histogram, blaser2016random} and binary histogram partition \citep{biau2012analysis}, we propose \textit{gradient boosted binary histogram ensemble} (GBBHE) for regression, where we for the first time combine two ensemble methods, i.e. gradient boosting and base learner aggregation, to enhance effectiveness and computational efficiency.
First of all, we apply a random rotation to the input space and apply the binary histogram partition, with the help of which we obtain piecewise constant base learners. 
Then at each iteration, we generate several independent base learners and ensemble them by taking the arithmetic average, where we call the ensemble estimator \textit{rotated binary histogram ensembles}. 
The iterative process to fit residuals is started with the help of a sequence of rotated binary histogram ensembles by a natural adaption of gradient descent boosting algorithm. 

It is worth mentioning that GBBHE enjoys four advantages. 
Firstly, our obtained regression function can be globally smooth thanks to the diversity of different base learners resulting from the randomness of both random rotations and binary histogram partitions. 
Secondly, under binary histogram partitions, all cells are split to the same depth, which ensures the fast convergence of our algorithm. 
Thirdly, compared with ordinary histograms, our binary histogram partition split at the midpoint of only one selected side at each iteration, so the cells will not be too small and thus it applies better to high dimensional data. 
Finally, GBBHE applies well to the large-scale scenarios since the more accurate base ensembles, enjoying high computational efficiency with the help of parallel computing, can effectively reduce the number of iterations in boosting. 

As follows are the contributions of this paper.

\textit{(i)} Aiming at solving the large-scale regression problem, we propose a novel boosting algorithm named \textit{Gradient Boosted Binary Histogram ensemble} (GBBHE), where the binary histogram regressors are used as base learners and the ensemble method performed on base learners helps improve its computational efficiency by reducing the number of boosting iterations. 
We claim that the binary histogram partition is adapted well to high-dimensional data, compared with ordinary histogram partitions. 

\textit{(ii)} From the theoretical perspective, we first of all investigate a special case of GBBHE, that is, GBBHE without ensemble (abbreviated as GBBH), where the number of base learners in an ensemble is set to be $1$.
Under the assumption that the target function resides in $C^{0,\alpha}$ and $C^{1,0}$, respectively, by decomposing the error term into approximation error and sample error, we establish the fast convergence rates of GBBH in the space $C^{0,\alpha}$. Moreover, for the subspace $C^{1,0}$ consisting of smoother functions, we are able to show that GBBH can attain the asymptotical convergence rate $O(n^{-1/(2+d\log 2)})$ whereas the lower bound of the convergence rates for its base learner binary histogram is merely of the order $O(n^{\log(1-0.75/d)/(\log 2-\log (1-0.75/d))})$. As a result, we succeed to prove that the boosted regressor GBBH can achieve faster convergence rates than the base learner in the subspace $C^{1,0}$ when $d \geq 6$, which confirms the benefits of the proposed boosting procedure.

\textit{(iii)}  We further prove that GBBHE has the same convergence rates as that of GBBH in both $C^{0,\alpha}$ and $C^{1,0}$. 
However, compared with the results for GBBH, where the number of iterations $T_n$ is required to be of the order $n^{1/(8+4d\log2)}$ to achieve the convergence rate, for GBBHE, we only require that $T_nK_n$ is of the same order $n^{1/(8+4d\log2)}$, where $K$ denotes the number of ensembles at each iteration. 
In other words, we can reduce the number of iterations $T$ by enlarging the number of base learners for ensemble $K$. 
It is well known that boosting algorithms with a large number of iterations can be quite time-consuming, since acceleration techniques such as parallel computing are not directly available. 
Therefore, by combining ensemble methods, which enjoy high computational efficiency with the help of parallel computing, we successfully save the running time of GBBHE by reducing the number of iterations $T$.

\textit{(iv)} In the experiments, several numerical experiments are designed to study the parameters including the depth of the binary histogram transform $p$, the number of iterations of boosting $T$, the number of binary histograms in each iteration $K$, and learning rate $\rho$. And we empirically verify the theoretical results that ensemble methods can reduce the number of iterations $T$ to achieve the fast convergence rate. Moreover, we compare our GBBHE with state-of-the-art large-scale regressors including GBRT, Breiman's forest, and LiquidSVM on both moderate-sized and large-scale real datasets. It turns out that our GBBHE shows comparable or even better performance than the compared methods, while enjoys higher computational efficiency with less running time required.

This paper is organized as follows. In Section \ref{sub::methodology}, we introduce the methodology of this paper, where we construct our main algorithm GBBHE. 
Then in Section \ref{sec::Thorecialresults}, we present the theoretical results built for GBBHE and its special case GBBH in H\"{o}lder space consisting of $(k,\alpha)$-H\"{o}lder continuous functions. 
We conduct numerical experiments including parameter analysis and real data comparisons with other state-of-the-art large-scale regression algorithms in Section \ref{sec::numerical}. 
Finally, we present our comments and discussions in Section \ref{sec::comments}. 
We put the error analysis for Section \ref{sec::Thorecialresults} in Section \ref{sec::ErrorAnalysis} and the related proofs in Section \ref{sec::proofs}.

\section{Methodology} \label{sub::methodology}

In this section, we build the main algorithm \textit{Gradient Boosted Binary Histogram Ensemble} (GBBHE) for regression. 
We first introduce some notations in Section \ref{sub::notations} and show the basics of least square regression in Section \ref{sub::LSRegression}. Then in Section \ref{sub::histogram}, we introduce the (rotated) binary histogram, which is an essential part in establishing the main algorithms. In Section \ref{sub::boosting}, we show a special case of GBBHE where the number of ensembles equals to $1$, called GBBH. Finally, in Section \ref{sec::GBBHE}, we present the main algorithm of GBBHE.

\subsection{Notations} \label{sub::notations}

Regression is to predict the value of an unobserved output variable $Y$ based on the observed input variable $X$, based on a dataset $D := \{ (x_1, y_1), \ldots, (x_n, y_n) \}$ consisting of i.i.d.~observations drawn from an unknown probability measure $\mathrm{P}$ on $\mathcal{X} \times \mathcal{Y}$. Throughout this paper, we assume that $\mathcal{X} \subset \mathbb{R}^d$ and $\mathcal{Y} \subset \mathbb{R}$ are compact and non-empty.

For any fixed $r > 0$, we denote $B_r$ as the centered hyper-cube of $\mathbb{R}^d$ with size $2r$, that is, $B_r := [-r, r]^d := \{ x = (x_1, \ldots, x_d) \in \mathbb{R}^d : x_i \in [-r, r], i = 1, \ldots, d \}$. Recall that for $1 \leq p < \infty$, the $L_p$-norm of $x = (x_1, \ldots, x_d)$ is defined by $\|x\|_p := (|x_1|^p + \cdots + |x_d|^p)^{1/p}$, and the $L_{\infty}$-norm is defined by $\|x\|_{\infty} := \max_{i=1,\ldots,d} |x_i|$.

In the sequel, the notation $a_n \asymp b_n$ means that there exists some positive constant $c \in (0, 1)$, such that $a_n \geq c b_n$ and $a_n \leq c^{-1} b_n$, for all $n \in \mathbb{N}$. Similarly, the notation $a_n \lesssim b_n$ denotes that there exists some positive constant $c \in (0, 1)$, such that $a_n \leq c b_n$ and $a_n \gtrsim b_n$ denotes that there exists some positive constant $c \in (0, 1)$, such that $a_n \geq c^{-1} b_n$. For any $x \in \mathbb{R}$, let $\lfloor x \rfloor$ denote the largest integer less than or equal to $x$. Moreover, the following multi-index notations are used frequently. For any vector $x=(x_i)^d_{i=1}\in \mathbb{R}^d$, we write $\lfloor x \rfloor:=(\lfloor x_i \rfloor )_{i=1}^d$, $x^{-1}:=(x_i^{-1})_{i=1}^d$, $\log (x):=(\log x_i)^d_{i=1}$, $\overline{x}=\max_{i=1,\ldots,d} x_i$, and  $\underline{x}=\min_{i=1,\ldots,d} x_i$.

\subsection{Least Square Regression} \label{sub::LSRegression}

It is legitimate to consider the least square loss $L : \mathcal{X} \times \mathcal{Y} \to [0, \infty)$ defined by $L(x, y, f(x)) := (y - f(x))^2$ for our target of regression. Then, for a measurable decision function $f : \mathcal{X} \to \mathcal{Y}$, the risk is defined by $\mathcal{R}_{L,\mathrm{P}}(f) := \int_{\mathcal{X} \times \mathcal{Y}} L(x, y, f(x)) \, d\mathrm{P}(x,y)$ and the empirical risk is defined by $\mathcal{R}_{L,\mathrm{D}}(f):=\frac{1}{n}\sum^n_{i=1} L(x_i,y_i,f(x_i))$. The Bayes risk, which is the smallest possible risk with respect to $\mathrm{P}$ and $L$, is given by $\mathcal{R}_{L, \mathrm{P}}^{*}:=\inf \{\mathcal{R}_{L, \mathrm{P}}(f) | f: \mathcal{X} \to \mathcal{Y} \text{ measurable} \}$.

In what follows, it is sufficient to consider predictors with values in $[-M, M]$. To this end, we introduce the concept of \textit{clipping} for the decision function, see also Definition 2.22 in \citet{StCh08}. 
Let $\wideparen{t}$ be the \textit{clipped} value of $t\in \mathbb{R}$ at $\pm M$ defined by $- M$ if $t < - M$,  $t$ if $t \in [-M, M]$, and $M$ if $t > M$. 
Then, a loss is called \textit{clippable} at $M > 0$ if, for all $(y, t) \in \mathcal{Y} \times \mathbb {R}$, there holds $L(x, y, \wideparen{t}) \leq L(x, y, t)$. 
According to Example 2.26 in \citet{StCh08}, the least square loss $L$ is \textit{clippable} at $M$ with the risk reduced after clipping, i.e., $\mathcal{R}_{L, \mathrm{P}}(\wideparen{f}) \leq \mathcal{R}_{L, \mathrm{P}}(f)$. Therefore, in the following, we only consider the clipped version $\wideparen{f}_{\mathrm{D}}$ of the decision function as well as the risk $\mathcal{R}_{L, \mathrm{P}}(\wideparen{f}_{\mathrm{D}})$.

\subsection{Binary Histogram for Regression} \label{sub::histogram}

In the following, we use the tilde notation to distinguish between the transformed space and the original input space.

\subsubsection{Random Rotation}

To give a clear description of one possible construction procedure of rotated random histograms, we start with the random rotation matrix $R$, which is a real-valued $d \times d$ orthogonal square matrix with unit determinant, that is
\begin{align}\label{RotationMatrix}
R^{\top} = R^{-1} \quad \text{ and } \quad \det(R) = 1.
\end{align}
Then we define the rotation transformation $H : \mathcal{X} \to \mathcal{X}$ by 
\begin{align}\label{equ::RBH}
H(x) := R \cdot x. 
\end{align}
In the following we will assume that each individual histogram is constructed in the transformed space $\tilde{\mathcal{X}} := H(\mathcal{X})$.

Here we describe a practical method for the construction of the random rotation transformations we are confined to in this study. Starting with a $d \times d$ square matrix $M$, consisting of $d^2$ independent univariate standard normal random variates, a Householder $Q R$ decomposition is applied to obtain a factorization of the form $M = R \cdot W$, with orthogonal matrix $R$ and upper triangular matrix $W$ with positive diagonal elements. The resulting matrix $R$ is orthogonal by construction and can be shown to be uniformly distributed. Unfortunately, if $R$ does not feature a positive determinant then it is not a proper rotation matrix according to definition (\ref{RotationMatrix}). In this case, we can change the sign of the first column of $R$ to construct a new rotation matrix $R^+$ that satisfies the condition \eqref{RotationMatrix}.

\subsubsection{Binary Histogram Partition}

It is well known that the classical histogram partition is an effective algorithm by grouping samples into the bins with the same shape. However, ordinary histogram partition is plagued by the curse of dimensionality. That is, the number of bins grows exponentially with the dimension $d$, in which case many bins will contain few or even no samples, leading to unacceptable and unnecessary computational costs. Therefore, we propose the \textit{binary partitioning} technique to build the high-dimensional histogram, namely \textit{binary histogram}.

To be specific, Let $\tilde{A}_0 = \{ A_0^1 := [-r, r]^d \} \subset \tilde{\mathcal{X}}$ be the rectangular cell in the transformed space $\tilde{\mathcal{X}}$ and $p$ be a deterministic parameter, fixed beforehand by the user, and possibly depending on $n$. In the first step, we choose one of the coordinates $X = (X_1, \ldots, X_d)$ with the $j$-th feature having a probability $1/d$ of being selected, and then split $[-r, r]^d$ into two rectangular cells along the midpoint of the chosen side. In other words, there exist $1 \leq \ell \leq d$ such that $[-r, r]^d = \tilde{A}_1^1 \cup \tilde{A}_1^2$, where $\tilde{A}_1^1 = \{ (x, y) \in [-r, r]^d : x_{\ell} \leq 0 \}$ and $\tilde{A}_1^2 = [-r, r]^d \setminus A_1^1$. In this way, we get a partition with two rectangular cells. Moreover, we denotes $\tilde{A}_1 := \{ \tilde{A}_1^1, \tilde{A}_1^2\}$ as the collection of partitions with two rectangular cells. Note that the total number of possible partitions after the first step is equal to the dimension $d$. Suppose after $i-1$ steps of the recursion, $1 \leq i \leq p$ we have obtained a partition $\tilde{A}_{i-1}$ of $[-r, r]^d$ with $2^{i-1}$ rectangular cells. In the $p$-th step, further partitioning of the region is defined as follows:
\begin{itemize}
\item[(i)]
For each rectangular cell $\tilde{A}_{i-1}^j$, $1 \leq j \leq 2^{i-1}$, a coordinate of $X = (X_1, \ldots, X_d)$, namely $Z_{i,j}$ is selected, with the $\ell$-th feature having a probability $1/d$ to be chosen, that is,
\begin{align}\label{equ::pzij}
\mathrm{P}(Z_{i,j} = \ell) = 1/d, 
\qquad
\text{ for } 1 \leq \ell \leq d.
\end{align}
\item[(ii)] 
For each rectangular cell $\tilde{A}_{i-1}^j$, $1 \leq j \leq 2^{i-1}$, once the coordinate is selected, the split is at the midpoint of the chosen side. As a result, each rectangular cell $\tilde{A}_{i-1}^j$ is divided into two new ones, namely $\tilde{A}_i^{2j-1}$ and $\tilde{A}_i^{2j}$. We denote the set of all these cells $\{\tilde{A}_i^j,\, 1\leq j\leq 2^i\}$ by $\tilde{A}_i$.
\end{itemize}

\begin{figure}[htb]
\centering
\subfigure[Rotated binary histogram induced by $H_1$.]{
\includegraphics[width=0.45\textwidth, trim=0 70 0 0,clip]{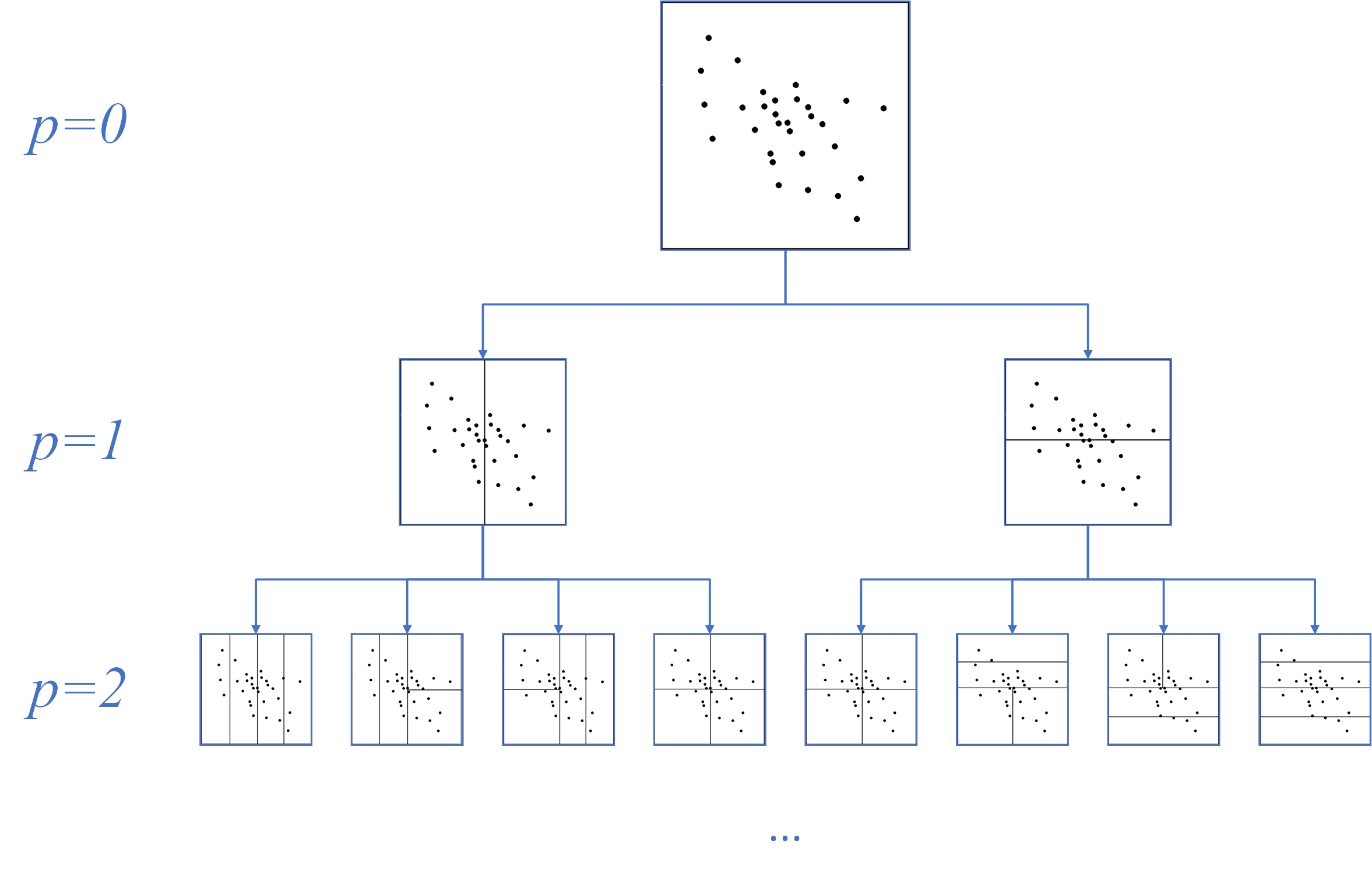}
}
\hspace{5mm}
\subfigure[Rotated binary histogram induced by $H_2$]{
\includegraphics[width=0.45\textwidth, trim=0 70 0 0,clip]{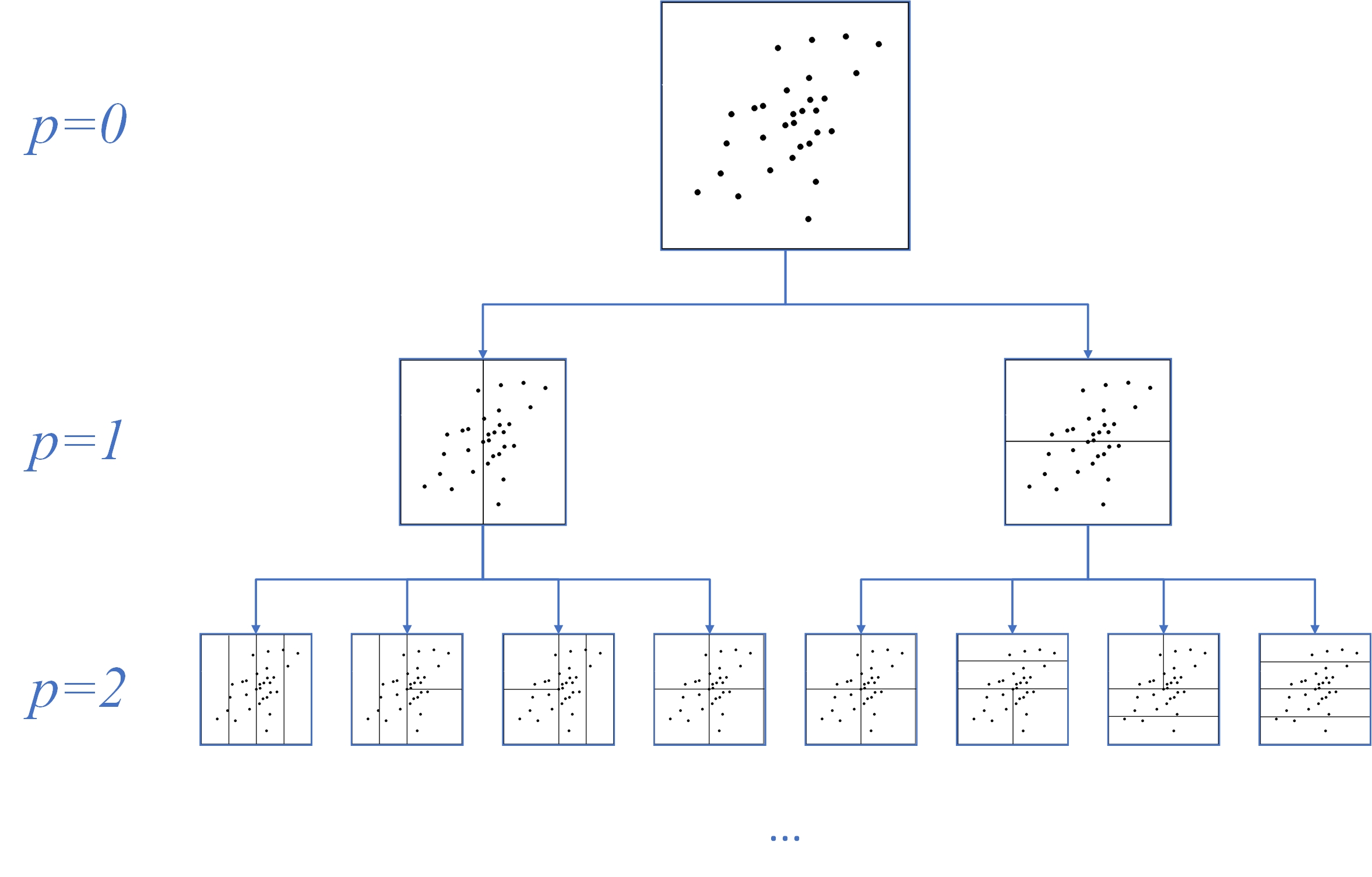}
}
\captionsetup{justification=centering}
\caption{Illustrations of rotated binary histogram partition with $p=2$.}
\label{fig::binarypartition}
\end{figure}

Such a partition obtained by $p$ recursive steps is called a \textit{binary histogram} partition of $[-r, r]^d$ with depth $p$ in the transformed space, and we say
\begin{align}\label{equ::defap}
A_p 
:= \{ A_p^j := H^{-1}(\tilde{A}_p^j), \, 1 \leq j \leq 2^p \} 
\cup \{ A_p^0 := \mathcal{X} \setminus H^{-1}([-r, r]^d) \}
\end{align}
an \textit{rotated binary histogram} with depth $p$ in the input space $\mathcal{X}$, where $H^{-1}(A) := \{ x \in \mathcal{X} : H(x) \in A \}$ for the set $A \subset \tilde{\mathcal{X}}$. The complete process is presented in Algorithm \ref{alg::AdaptivePartitioning} and an illustration is shown in Figure \ref{fig::binarypartition}.

\begin{algorithm}[h]
\caption{Binary Histogram with Rotation}
\label{alg::AdaptivePartitioning}
\begin{algorithmic}
\STATE {\bfseries Input:}
Depth of the binary histogram $p$; \\
\quad\quad\, \, \, 
The random rotation transformation $H(x) = R \cdot x$; \\
\quad\quad\, \, \,
$\tilde{A}_0 = \{ \tilde{A}_0^1 := [-r, r]^d \}$ in the transformed space $\tilde{\mathcal{X}} = H(\mathcal{X})$.
\FOR{$i = 1$ {\bfseries to} $p$}
\FOR{$j=1$ {\bfseries to} $2^{i-1}$}
\STATE For rectangular cell $\tilde{A}_{i-1}^j$, randomly choose one dimension coordinate $Z_{i,j}$ whose probability distribution is given by \eqref{equ::pzij}; \\
\STATE Divide the cell $\tilde{A}_{i-1}^j$ into two subregions, that is, $\tilde{A}_{i-1}^j = \tilde{A}_i^{2j-1} \cup \tilde{A}_i^{2j}$, along the midpoint of the dimension $Z_{i,j}$; 
\ENDFOR
\STATE Get $\tilde{A}_i =\{\tilde{A}_i^j,\, 1\leq j\leq 2^i\}$.
\ENDFOR
\STATE {\bfseries Output:}  Rotated binary histogram $A_p = \{ A_p^j = H^{-1}(\tilde{A}_p^j), \, 1 \leq j \leq 2^p \}$, where $H^{-1}(A) := \{ x \in \mathcal{X} : H(x) \in A \}$ for the set $A \subset \tilde{\mathcal{X}}$.
\end{algorithmic}
\end{algorithm}

For any $x \in H^{-1}(B_r)$, the histogram bin containing $x \in \mathcal{X}$ in the input space is 
\begin{align}\label{equ::InputBin}
A_p(x) := \{ x' \ | \exists 1 \leq i \leq 2^p \text{ such that } H(x) \in \tilde{A}_p^i, \, H(x') \in \tilde{A}_p^i \}
\end{align}
and we further denote all the bins as $\{ A_p(x) : x \in \mathcal{X} \}$ with the repetitive bin counted only once, and $\mathcal{I}_p$ as the index set. In other words, the set $\pi_{p} = \{ A_j \}_{j\in \mathcal{I}_p}$ forms a partition of $H^{-1}(B_r)$.

\subsubsection{Binary Histogram Regressor}

We consider the following function set $\mathcal{F}^p$ defined by
\begin{align}\label{equ::functionFn}
\mathcal{F}^p := \biggl\{ \sum_{j \in \mathcal{I}_p} c_j \eins_{A_j} : c_j \in [-M, M] \biggr\}.
\end{align}
In order to constrain the complexity of $\mathcal{F}^p$, we penalize on the depth $p$ of the partition $\pi_p$. Then the Binary Histogram with rotation can be produced by the regularized empirical risk minimization (RERM) over $\mathcal{F}^p$, i.e.
\begin{align}\label{equ::fD}
(f_{\mathrm{D}},h^*)
= \argmin_{f \in \mathcal{F}^p, \, p \in \mathbb{N}_+} \Omega(p) + \mathcal{R}_{L,\mathrm{D}}(f),
\end{align}
where $\Omega(p) :=\lambda \cdot 4^p$. It is clear to see that the penalty on the depth $p$ of the partition helps to control the bin width of the rectangular cells.

We mention that when the rotation matrix $R$ is deterministic as the identity matrix, we call the algorithm in Algorithm \ref{alg::AdaptivePartitioning} and the regressor in \eqref{equ::fD} \textit{binary histogram without rotation}.

\subsection{Gradient Boosted Binary Histogram (GBBH) for Regression}\label{sub::boosting}

Boosting is the task of converting inaccurate weak learners into a single accurate predictor. To be specific, we define a restricted family of functions $\mathcal{F}$ be a set of base learners and a general boosting algorithm is combining a sequence of functions $\{ f_t\}_{t=1}^T$ from $\mathcal{F}$ to minimize a certain empirical loss. 
Then the final predictor can be represented as 
\begin{align*}
F = \sum^T_{t=1} w_t f_t,
\end{align*}
where $w_t \geq 0$, $t = 1, \ldots, T$, are weights and $f_t \in \mathcal{F}$, $t = 1, \ldots,T$. From a functional gradient descent viewpoint in statistics \citep{friedman2001greedy}, boosting is reformulated as a stage-wise optimization problem with different loss functions. In this scenario, gradient boosting requires computing the negative \textit{functional gradient} as the response 
\begin{align*}
U_i=-\frac{\partial L(y_i,f(x_i))}{\partial f(x_i)}\bigg |_{f(x_i)=\hat{f}(x_i)}
\end{align*}
and select a particular model from the allowable class of functions at each boosting iteration to update the predictor.

In this work, we mainly focus on the boosting algorithm equipped with (rotated) binary histogram regressors as base learners since they are weak predictors and enjoy computational efficiency in high dimensional spaces. 
Before we proceed, we need to introduce the function space that we are most interested in to establish our learning theory. Assume that $\{ R_t \}_{t=1}^T$ is an i.i.d.~sequence of random rotation matrix drawn from some probability measure $\mathrm{P}_R$ and $\{ Z^t_{i,j}, 1 \leq i \leq p, 1 \leq j \leq 2^{i-1}, 1 \leq t \leq T\}$ is an i.i.d.~sequence of selected coordinate to split drawn from the probability measure  $\mathrm{P}_Z$ given by \eqref{equ::pzij}. For $1 \leq t \leq T$, given rotation transformation $H_t$ and select coordinates $Z_t := \{ Z^t_{i,j}, 1 \leq i \leq p, 1 \leq j \leq 2^{i-1} \}$, we define the function space $\mathcal{F}_t := \mathcal{F}^{p}_t$, $t = 1, \ldots, T$ according to \eqref{equ::functionFn}. Then for $r > 0$, we define the function space $E$ by
\begin{align}\label{equ::En}
E := \biggl\{ f : B_r \to \mathbb{R} \, \bigg| \, f = \sum_{t=1}^T w_t f_t, \, f_t \in \mathcal{F}_t \biggr\}.
\end{align}
Moreover, for $f \in E$, we define 
\begin{align}\label{equ::fenorm}
\|f\|_E := \inf \biggl\{ \sum_{t=1}^T |w_t|^2 \text{ with } f = \sum_{i=1}^T w_t f_t, \, f_t \in \mathcal{F}_t \biggr\}.
\end{align}
Then for any $f \in E$, by the Cauchy-Schwarz inequality, we immediately get
\begin{align*}
\|f\|_{\infty}
\leq M \sum_{t=1}^T |w_t| 
\leq M (T \|f\|_E)^{1/2}. 
\end{align*}
In fact, $(E, \|\cdot\|_E)$ is a function space that consists of measurable and bounded functions.

As is mentioned above, boosting methods may be viewed as iterative methods for optimizing a convex empirical cost function. To simplify the theoretical analysis, following the approach of \cite{blanchard2003rate}, we ignore the dynamics of the optimization procedure and simply consider minimizers of an empirical cost function to establish the oracle inequalities, which leads to the following definition.

\begin{definition}\label{def::RBRBH}
Let $E$ be the function space \eqref{equ::En} and $L$ be the least square loss. Given $\lambda_1 > 0$, $\lambda_2 > 0$, we call a learning method that assigns to every $D \in (\mathcal{X} \times \mathcal{Y})^n$ a function $f_{\mathrm{D},B} : \mathcal{X} \to \mathbb{R}$ such that
\begin{align}\label{equ::fdlambda}
(f_{\mathrm{D},B}, h^*) 
= \argmin_{f \in E, \, p\in \mathbb{N}_+} \Omega_{\lambda}(f) + \mathcal{R}_{L,\mathrm{D}}(f)
\end{align}
a gradient boosted binary histogram (GBBH) algorithm for regression with respect to $E$, where $\Omega_{\lambda}(f)$ is defined by
\begin{align}\label{equ::omegalambdaf}
\Omega_{\lambda}(f)
:= \lambda_1 \Omega_1(f) + \lambda_2 \Omega_2(f)
:= \lambda_1 \|f\|_E + \lambda_2\cdot 4^p.
\end{align}
\end{definition}

The regularization term defined in \eqref{equ::omegalambdaf} consists of two components. The first term is motivated by the fact the early boosting methods such as Adaboost may overfit in the presence of label noise. It helps control the degree of overfitting by the $L_2$-norm of the weights of the composite estimators and helps achieve the consistency and convergence results. 
The second term is added to control the bin width of the binary histogram, which has been discussed in subsection \ref{sub::histogram}.  In fact, it is equivalent to adding the $L_p$-norm of the base learners $f_t$, since piecewise constant functions are applied on the rectangular cells.

To conduct the theoretical analysis, we also need the infinite sample version of Definition \ref{def::RBRBH}. To this end, we fix a distribution $\mathrm{P}$ on $\mathcal{X} \times \mathcal{Y}$ and let the function space $E$ be as in \eqref{equ::En}. Then every $f_{\mathrm{P},B} \in E$ satisfying
\begin{align*}
\Omega_{\lambda}(f_{\mathrm{P},B}) + \mathcal{R}_{L,\mathrm{P}}(f_{\mathrm{P},B}) 
= \inf_{f \in E} \Omega_{\lambda}(f) + \mathcal{R}_{L,\mathrm{P}}(f)
\end{align*}
is called an \textit{infinite sample version} of GBBH with respect to $E$ and $L$. Moreover, the approximation error function $A(\lambda)$ is defined by
\begin{align}\label{equ::approximationerror}
A(\lambda) 
= \inf_{f \in E} \Omega_{\lambda}(f) + \mathcal{R}_{L,\mathrm{P}}(f) - \mathcal{R}^*_{L,\mathrm{P}}.
\end{align}

With all these preparations, we now present a general form of algorithm for GBBH in Algorithm \ref{alg::BRBH}. 
Indeed, the randomness of rotation matrix and binary histogram splitting rule provides an effective procedure for carrying out boosting. With the help of (rotated) binary histogram regressor, we repeat the least squares fitting of residuals. 
We mention that when the rotation matrix is determined as the identity matrix, we call the algorithm in Algorithm \ref{alg::BRBH} GBBH without rotation. Moreover, we introduce the learning rate $\rho$ to dampen the move on the gradient descent update, which is related to the regularization through shrinkage.

\begin{algorithm}[!htbp]
\caption{Gradient Boosted Binary Histogram (GBBH) for Regression}
\label{alg::BRBH}
\begin{algorithmic}
\STATE {\bfseries Input:} Training data $D := \{ (x_1, y_1), \ldots, (x_n, y_n) \}$;
\\
\quad\quad\, \, \, Depth of binary histogram $p$;
\\
\quad\quad\, \, \, Learning rate $\rho > 0$;
\\
\STATE{\bfseries Initialization:} $U_i=y_i$, $i=1,\ldots,n$.
$\widehat{F}_{0}(x)=0$.
\\
\FOR{$t = 1$ {\bfseries to} $T$}
\STATE Compute the random rotation transformation
\begin{align*}
H_t(x) := R_t \cdot x.
\end{align*}
\STATE Apply rotated binary histogram in Algorithm \ref{alg::AdaptivePartitioning} to the transformed space; 
\STATE Apply constant functions to each cell, i.e., fit residuals with function $f_t$ such that
\begin{align*}
f_{t}=\argmin_{f\in \mathcal{F}_t} \frac{1}{n}\sum^n_{i=1} L(U_i,f(x_i)),
\end{align*}
where $\mathcal{F}_t$ is defined as in \eqref{equ::functionFn} with respect to $H_t$ and selected coordinates $Z_t$.
\STATE {\bfseries Update:} $\widehat{F}_t(x)=\widehat{F}_{t-1}(x)+\rho f_t(x)$.
\STATE Compute residuals $U_i=U_i-\widehat{F}_t(x_i)$, $i = 1, \ldots, n$.
\ENDFOR
\STATE {\bfseries Output:} Gradient boosted binary histogram regressor is
$f_{\mathrm{D},B}(x) = \widehat{F}_T(x)$.
\end{algorithmic}
\end{algorithm}

\begin{remark}
In fact, the Gradient Boosting Algorithm \ref{alg::BRBH} converges to the empirical risk minimizer of the mean squared error with respect to the function space $E$  defined by \eqref{equ::En}
\begin{align}\label{equ::mse}
\frac{1}{n}\sum^n_{i=1}(y_i-F(x_i))^2,
\end{align}
which is illustrated as below:
\begin{enumerate}
\item[(i)] 
In the least-square regression setting, the goal of a gradient boosting algorithm with $T$ stages is to fit a function $F$ of the form $F(x)=\sum^T_{t=1} f_t(x)$ to minimize \eqref{equ::mse}. 
\item[(ii)] 
At stage $t\ (1\leq t\leq T)$, our algorithm should add some new estimator to improve some imperfect model $F_{t-1}$ to correct the errors of its predecessor. For regression problems, we observe that residuals are the negative gradients (with respect to $F(x)$) of the squared error loss function $(y-F(x))^2/2$. Then gradient boosting will fit $f_t$ from the hypothesis space defined as in \eqref{equ::functionFn} to the residuals $U_i=y_i-F_{t-1}(x_i)$. 
\item[(iii)] 
Algorithm \ref{alg::BRBH} does so by starting with a model $F_0(x)=0$, and incrementally expands it in a greedy fashion. The main idea is to apply a (functional gradient) descent step to this minimization problem to solve the computationally infeasible optimization problem in general.
\item[(iv)] 
The regularization of gradient boosting methods is realized by shrinkage which consists of modifying the update rule with learning rate as shown in Algorithm \ref{alg::BRBH} to improve the generalization ability of the model. 
\end{enumerate}
In summary, in accordance with the empirical risk minimization principle, gradient boosting algorithm tries to find an approximation $F(x)$ that minimizes the average value of the loss function on the training set, i.e., minimizes the empirical risk with respect to the space $E$ defined by \eqref{equ::En}. 
\end{remark}

\subsection{Gradient Boosted Binary Histogram Ensemble (GBBHE) for Large-scale Regression}\label{sec::GBBHE}

For large-scale regression problems, however, GBBH may be of low computational efficiency since it requires a large number of boosting iterations in practical applications. As a result, it is important to find some ways to speed up the applications of GBBH. An intuitive idea is to increase the accuracy of the base learner and thus reduce the number of iterations. To be specific, we proposed a modified boosting algorithm using the ensemble of (rotated) binary histogram regressors as base learners. Recall that in Algorithm \ref{alg::BRBH}, at $t$-th round of boosting iteration, gradient boosting requires computing the negative {functional gradient} as the response 
\begin{align}\label{equ::uti}
U^t_i
= - \frac{\partial L(y_i,f(x_i))}{\partial f(x_i)} \bigg|_{f(x_i) = \widehat{F}_{t-1}(x_i)}
\end{align}
and select a particular model to fit residuals to update the current predictor $F^{t-1}(x)$.

Before we proceed, we denote the number of boosting iterations as $T$ and the number of learners for ensemble at each boosting iteration as $K$. Moreover, we assume that $\{ R_t^k, 1 \leq t \leq T, 1 \leq k \leq K \}$ is an i.i.d.~sequence of random rotation matrix drawn from some probability measure $\mathrm{P}_R$ and $\{ Z^{t,k}_{i,j}, 1 \leq i \leq p, 1 \leq j \leq 2^{i-1}, 1 \leq t \leq T, 1 \leq k \leq K \}$ is an i.i.d.~sequence of selected coordinate to split drawn from the probability measure  $\mathrm{P}_Z$ given by \eqref{equ::pzij}. For $1 \leq t \leq T$ and $1 \leq k \leq K$, given the rotation transformation $H_t^k$ and select coordinates $Z_t^k := \{ Z^{t,k}_{i,j}, 1 \leq i \leq p, 1 \leq j \leq 2^{i-1} \}$, we define the function space $\mathcal{F}^k_t$ according to \eqref{equ::functionFn}. In the scenario of gradient boosted binary histogram ensemble for large-scale regression, at $t$-th round of boosting iteration, let $D_t := \{ (x_1, U_1^t), \ldots, (x_n, U_n^t) \}$ be the training data with $U_i^t$, $1 \leq i \leq n$ defined by \eqref{equ::uti}. For $1\leq k\leq K$, we write
\begin{align} \label{def::fDHkt}
f_{\mathrm{D}}^{t,k} := \argmin_{f \in \mathcal{F}^k_t} \mathcal{R}_{L,\mathrm{D}_t}(f).
\end{align}
Then the ensemble of the (rotated) binary histogram regressor at $t$-th round of boosting is defined by
\begin{align*}
\bar{f}_t := \frac{1}{K} \sum_{k=1}^K f^{t,k}_{\mathrm{D}}.
\end{align*}
It is clear to see that $\bar{f}_t$ belongs to the function space $\bar{E}_t$ given by
\begin{align}\label{equ::barft}
\bar{\mathcal{F}}_t := \biggl\{ f : B_r \to \mathbb{R} \, \bigg| \, f = \frac{1}{K}\sum_{k=1}^K f_t^k, \, f_t^k \in \mathcal{F}_t^k \biggr\}.
\end{align}
Then we define the function space $\bar{E}$ by
\begin{align}\label{equ::barEn}
\bar{E} := \biggl\{ f : B_r \to \mathbb{R} \, \bigg| \, f = \sum_{t=1}^T w_t \bar{f}_t, \, \bar{f}_t \in \bar{\mathcal{F}}_t \biggr\}.
\end{align}
Moreover, for $f \in \bar{E}$, we define 
\begin{align*}
\|f\|_{\bar{E}} := \inf \biggl\{ \sum_{t=1}^T |w_t|^2 \text{ with } f = \sum_{i=1}^T w_t \bar{f}_t, \, \bar{f}_t \in \bar{\mathcal{F}}_t \biggr\},
\end{align*}
which is similar as \eqref{equ::fenorm} where (rotated) binary histogram regressors are used as base learners.

To simplify the theoretical analysis, by similar arguments in Section \ref{sub::boosting}, we ignore the dynamics of the optimization procedure and simply consider minimizers of an empirical cost function to establish the oracle inequalities for GBBHE, which leads to the following definition.

\begin{definition}\label{def::RBRBHlarge}
Let $\bar{E}$ be the function space \eqref{equ::barEn} and $L$ be the least square loss. Given $\lambda_1 > 0$, $\lambda_2 > 0$, we call a learning method that assigns to every $D \in (\mathcal{X} \times \mathcal{Y})^n$ a function $\bar{f}_{\mathrm{D},B} : \mathcal{X} \to \mathbb{R}$ such that
\begin{align}\label{equ::barfdlambda}
(\bar{f}_{\mathrm{D},B}, h^*) 
= \argmin_{f \in \bar{E}, \, p\in \mathbb{N}_+} \bar{\Omega}_{\lambda}(f) + \mathcal{R}_{L,\mathrm{D}}(f)
\end{align}
a gradient boosted binary histogram ensemble (GBBHE) algorithm for regression with respect to $E$, where $\bar{\Omega}_{\lambda}(f)$ is defined by
\begin{align}\label{equ::omegalambdaflarge}
\bar{\Omega}_{\lambda}(f)
:= \lambda_1 \bar{\Omega}_1(f) + \lambda_2 \bar{\Omega}_2(f)
:= \lambda_1 \|f\|_{\bar{E}} + \lambda_2\cdot 4^p.
\end{align}
\end{definition}

Note that when the rotation matrix is determined as the identity matrix, we call the algorithm in Definition \ref{def::RBRBHlarge} GBBHE without rotation.

Compared with the regularization term of \eqref{equ::omegalambdaf}, the first component of \eqref{equ::omegalambdaflarge} is replaced by $L_2$-norm of the weights of the composite estimators, that is, the ensemble of the (rotated) binary histogram regressors to achieve the consistency and convergence rates.

To conduct the theoretical analysis, we also need the infinite sample version of Definition \ref{def::RBRBHlarge}. Let the function space $\bar{E}$ be as in \eqref{equ::En}. Then every $\bar{f}_{\mathrm{P},B} \in \bar{E}$ satisfying
\begin{align*}
\bar{\Omega}_{\lambda}(\bar{f}_{\mathrm{P},B}) + \mathcal{R}_{L,\mathrm{P}}(\bar{f}_{\mathrm{P},B}) 
= \inf_{f \in \bar{E}} \bar{\Omega}_{\lambda}(f) + \mathcal{R}_{L,\mathrm{P}}(f)
\end{align*}
is called an infinite sample version of GBBHE with respect to $\bar{E}$ and $L$. Moreover, the approximation error function $\bar{A}(\lambda)$ is defined by
\begin{align}\label{equ::barapproximationerror}
\bar{A}(\lambda) 
= \inf_{f \in \bar{E}} \bar{\Omega}_{\lambda}(f) + \mathcal{R}_{L,\mathrm{P}}(f) - \mathcal{R}^*_{L,\mathrm{P}}.
\end{align}

With all these preparations, we now present a general form of the algorithm for GBBHE in Algorithm \ref{alg::GBBHE}. 
Indeed, the randomness of (rotated) binary histogram provides an effective procedure for carrying out boosting. Moreover, the ensemble of (rotated) binary histogram regressors helps to improve the accuracy of each base estimator. As a result, fewer iteration rounds are required to achieve satisfying performance in practical applications.

\begin{algorithm}[!h]
\caption{Gradient Boosted Binary Histogram Ensemble (GBBHE) for Large-scale Regression}
\label{alg::GBBHE}
\begin{algorithmic}
\STATE {\bfseries Input:} Training data $D := \{ (x_1, y_1), \ldots, (x_n, y_n) \}$;
\\
\quad\quad\, \, \, Depth of binary histogram $p$;
\\
\quad\quad\, \, \, Learning rate $\rho > 0$;
\\
\quad\quad\, \, \, Number of ensembles $K>0$;
\\
\STATE{\bfseries Initialization:} $U_i=y_i$, $i=1,\ldots,n$.
$\widehat{F}_{0}(x)=0$.
\\
\FOR{$t = 1$ {\bfseries to} $T$}
\FOR{$k =1$ {\bfseries to} $K$}
\STATE Compute the random rotation transformation
\begin{align*}
H_t^k(x) := R_t^k \cdot x.
\end{align*}
\STATE Apply rotated binary histogram in Algorithm \ref{alg::AdaptivePartitioning} to the transformed space; 
\STATE Apply constant functions to each cell, i.e., fit residuals with function $f_t$ such that
\begin{align*}
f_{t}^k=\argmin_{f\in \mathcal{F}_t^k} \frac{1}{n}\sum^n_{i=1} L(U_i,f(x_i)),
\end{align*}
where $\mathcal{F}_t^k$ is defined as in \eqref{equ::functionFn} with respect to $H_t^k$ and selected coordinates $Z_t^k$.
\ENDFOR
\STATE Compute
\begin{align*}
\bar{f}_t=\frac{1}{K}\sum^K_{k=1} f_t^k.
\end{align*}
\STATE {\bfseries Update:} $\widehat{F}_t(x)=\widehat{F}_{t-1}(x)+\rho \bar{f}_t(x)$.
\STATE Compute residuals $U_i=U_i-\widehat{F}_t(x_i)$, $i = 1, \ldots, n$.
\ENDFOR
\STATE {\bfseries Output:} Boosted ensemble of rotated binary histogram regressor
$\bar{f}_{\mathrm{D},B}(x) = \widehat{F}_T(x)$.
\end{algorithmic}
\end{algorithm}

\section{Theoretical Results} \label{sec::Thorecialresults}

In this section, we build theoretical results for GBBH and GBBHE in H\"{o}lder space $C^{k,\alpha}$ consisting of $(k,\alpha)$-H\"{o}lder continuous functions of different order of smoothness. To be specific, Section \ref{sec::assump} shows the fundamental assumption. In Section \ref{sec::c0} and \ref{sec::c1}, we show the convergence rates for GBBH in $C^{0,\alpha}$ and $C^{1,0}$, respectively. Then in Section \ref{sec::c0large} and \ref{sec::c1large}, we show the convergence rates for GBBHE in $C^{0,\alpha}$ and $C^{1,0}$, respectively.

\subsection{Fundamental Assumption}\label{sec::assump}

We assume that the Bayes decision function resides in the H\"{o}lder space.

\begin{definition} \label{def::Cp}
We say that a function $f : \mathcal{X} \to \mathbb{R}$ is $\alpha$-H\"{o}lder continuous, namely $f(x) \in C^{0,\alpha}$ for $0 < \alpha \leq 1$, if there exists a finite constant $c_L > 0$ such that for any $x, y \in \mathcal{X}$, we have $|f(x) - f(y)| \leq c_L \|x - y\|^{\alpha}$. Moreover, if $f(x)$ is differentiable at every $x \in \mathcal{X}$ such that $\|\nabla f(x)\| \leq c_L$, we say that $f \in C^{1,0}$.
\end{definition}

From Definition \ref{def::Cp} we see that the functions contained in the space $C^{0,\alpha}$ with larger $\alpha$ enjoy higher level of smoothness. It is worth pointing out that $C^{1,0}$ is a proper subset of $C^{0,1}$ since there exist Lipschitz continuous functions that are not everywhere differentiable.

Note that according to the definition of the function class $\mathcal{F}^p$ in \eqref{equ::functionFn}, given an rotated binary histogram partition and $f \in \mathcal{F}^p$, there holds $f(x) = 0$ for $H(x) \notin [-r, r]^d$. As a result, to derive consistency and convergence rates of boosted estimators, we further assume that $\mathrm{P}_X$ is the uniform distribution on  $B_{r,d} := [-r/\sqrt{d}, r/\sqrt{d}]^d$ in the theoretical results. For the sake of brevity, we write $\mathrm{P}_{R,Z} := \mathrm{P}_R \otimes \mathrm{P}_Z$ in the following sections.

\subsection{Convergence Rates for GBBH in $C^{0,\alpha}$} \label{sec::c0}

\begin{theorem}\label{thm::tree}
Let $f_{\mathrm{D},B}$ be the GBBH regressor defined by \eqref{equ::fdlambda}. Moreover, suppose that the Bayes decision function $f_{L, \mathrm{P}}^* \in C^{0,\alpha}$ and $\mathrm{P}_X$ is the uniform distribution on  $B_{r,d} = [-r/\sqrt{d}, r/\sqrt{d}]^d$. Furthermore, let $\{ \lambda_{1,n} \}$, $\{ \lambda_{2,n} \}$ and $\{ p_n \}$ be chosen as
\begin{align*}
\lambda_{1,n} = n^{-\frac{2(1-4^{-\alpha})}{(4-2\delta)(1-4^{\alpha})+2d\log 2}}, \,
\lambda_{2,n} = n^{-\frac{2(1-4^{-\alpha}+2d\log 2)}{(4-2\delta)(1-4^{-\alpha})+2d\log 2}}, \,
p_n \asymp \frac{2d\log 2\log n}{(4-2\delta)(1-4^{-\alpha})+2d\log 2},
\end{align*}
where $\delta := 1/(d\cdot 2^p)$. Then for sufficiently large $n$, there holds
\begin{align}\label{eq::orderc0}
\mathbb{E}_{\mathrm{P}_{R,Z}} \bigl( \mathcal{R}_{L, \mathrm{P}}(f_{\mathrm{D},B}) - \mathcal{R}_{L,\mathrm{P}}^* \bigr) 
\lesssim  n^{-\frac{2(1-4^{-\alpha})}{(4-2\delta)(1-4^{-\alpha})+2d\log 2}}
\end{align}
with probability $\mathrm{P}^n$ equal to one. 
\end{theorem}

In particular, if $f^*_{L,\mathrm{P}}\in C^{0,1}$, then for sufficiently large $n$, there holds
\begin{align*}
\mathbb{E}_{\mathrm{P}_{R,Z}}\bigl(
\mathcal{R}_{L, \mathrm{P}}(f_{\mathrm{D},B}) - \mathcal{R}_{L,\mathrm{P}}^*\bigr) 
\lesssim  n^{-\frac{3}{3(2-\delta)+4d\log 2}}
\end{align*}
with probability $\mathrm{P}^n$ equal to one.

\begin{remark}[Convergence rate] \label{rem::RateC0alpha}
Under mild assumption that the target function is $\alpha$-H\"{o}lder continuous, i.e., $f_{L, \mathrm{P}}^* \in C^{0,\alpha}$, we derive the convergence rate of GBBH with probability one. In particular, when $\alpha = 1$, the convergence rate is $n^{-0.75/(1.5+d\log 2)}$. We will show in the next subsection that GBBH attains faster convergence rate in the subspace $C^{1,0}$.
\end{remark}

\begin{remark}[Effect of Rotation]\label{rem::rotation}
We remark that the rotation transformation does not affect the order of convergence rates. That is, in Theorem \ref{thm::tree}, when the rotation matrix is determined as the identity matrix, the convergence rate of GBBH is the same order as in \eqref{eq::orderc0}. We mention that the effect of rotation also holds for Theorems \ref{thm::optimalForest}, \ref{thm::c01gbbhte}, and \ref{thm::optimalForestlarge}.
\end{remark}

\begin{remark}[Effect of the number of iterations $T$]
In Theorem \ref{thm::tree}, we establish a uniform upper bound for the excess risk regardless of the number of iterations $T$. In other words, from the perspective of convergence rate, we cannot demonstrate the advantage of the boosting estimator over its base learners. Therefore, in the next subsection, we turn to the subspace $C^{1,0}$ to show the effect of boosting with the help of a lower bound for the convergence rate of the base learners.
\end{remark}

\subsection{Convergence Rates for GBBH in $C^{1,0}$} \label{sec::c1}

\begin{theorem}\label{thm::optimalForest}
Let $f_{\mathrm{D},B}$ be the GBBH regressor defined by \eqref{equ::fdlambda}. Moreover, suppose that the Bayes decision function $f_{L, \mathrm{P}}^* \in C^{1,0}$ and $\mathrm{P}_X$ is the uniform distribution on  $B_{r,d}=[-r/\sqrt{d},r/\sqrt{d}]^d$. Furthermore, let $\{ \lambda_{1,n} \}$, $\{ \lambda_{2,n} \}$, $\{T_n\}$ and $\{ p_n \}$ be chosen as
\begin{align*}
\lambda_{1,n} := n^{-\frac{3}{4(2-\delta+d\log 2)}}, \,
\lambda_{2,n} := n^{-\frac{2\log 2d+1}{2-\delta+d\log 2}}, \, 
T_n := n^{\frac{1}{4(2-\delta+d\log 2)}}, \, 
p_n \asymp \frac{d\log n}{2-\delta+d\log 2},
\end{align*}
where $\delta :=1/(d\cdot 2^p)$. Then for sufficiently large $n$, there holds
\begin{align} \label{UpperBoundEnsemble}
\mathbb{E}_{\mathrm{P}_{R,Z}}\mathcal{R}_{L,\mathrm{P}}(f_{\mathrm{D},B}) - \mathcal{R}_{L,\mathrm{P}}^* 
\lesssim n^{-\frac{1}{2-\delta+d\log 2}}
\end{align}
with probability $\mathrm{P}^n$ equal to one.
\end{theorem}

\begin{remark}[Convergence rate]
As is shown in Theorem \ref{thm::optimalForest}, when $f_{L, \mathrm{P}}^* \in C^{1,0}$, GBBH attains the asymptotically convergence rate $n^{-1/(2+d\log 2)}$. 
Compared with Remark \ref{rem::RateC0alpha}, we find that GBBH converges faster in the subspace $C^{1,0}$ than in the space $C^{0,1}$.
\end{remark}

\begin{remark}[Effect of the number of iterations $T$]
Different from the results in Theorem \ref{thm::tree}, where we derive a uniform upper bound for the excess risk regardless of the change of $T$, the excess risk decreases as $T_n$ increases at first, and then achieves its minimum when the number of iterations $T_n$ attains a certain level when $f_{L, \mathrm{P}}^* \in C^{1,0}$. This indicates that our proposed GBBH tends to perform better when the target function is of higher order of smoothness.
\end{remark}

To demonstrate the advantage of boosting, we show a lower bound for the convergence rate of the base learners in the following theorem. In this case, let $H$ be the identity map in the rotated binary histogram defined by Algorithm \ref{alg::AdaptivePartitioning}, that is, we leave out the randomness of the transform. Moreover, we suppose that $\mathrm{P}_X$ is the uniform distribution on $B_r:=[-r,r]^d $ instead of on $B_{r,d}=[-r/\sqrt{d},r/\sqrt{d}]^d$ mentioned above.

\begin{theorem}\label{prop::counter}
Let the rotated binary histogram $A_p$ be defined by Algorithm \ref{alg::AdaptivePartitioning} with the identity map $H(x) = x$. Moreover, let the binary histogram regressor $f_{\mathrm{D}}$ be defined as in \eqref{equ::fD} and the regression model be defined by $Y := f(X) + \varepsilon$, where $\mathrm{P}_X$ is the uniform distribution on $B_r = [-r, r]^d$ and $\varepsilon$ is independent of $X$ such that $\mathbb{E}(\varepsilon) = 0$ and $\mathrm{Var}(\varepsilon) = \sigma^2 < \infty$. Moreover, assume that $f \in C^{1,0}$ and there exists a constant $\underline{c}_f\in (0,\infty)$ such that $\|\nabla f\|\geq \underline{c}_f$ and $\|f\|_{\infty}\geq \underline{c}_f$. Then we have
\begin{align} \label{LowerBoundNaive}
\mathbb{E}_{\mathrm{P}^n \otimes \mathrm{P}_R} \bigl( \mathcal{R}_{L,\mathrm{P}}(f_{\mathrm{D}}) - R^*_{L,\mathrm{P}} \bigr)
\geq c_0 n^{\frac{\log (1-0.75/d)}{\log 2-\log(1-0.75/d)}}\vee c_1,
\end{align}
where $c_0$ and $c_1$ are constants depending on $r$, $d$, $\underline{c}_f$ and $\sigma$ which will be
specified in the proof.
\end{theorem}

\begin{remark}[Benefits of boosting]
In Theorem \ref{prop::counter}, we show that for some $f \in C^{1,0}$, the excess risk of the binary histogram regressor attains $n^{(\log (1-0.75/d))/(\log 2 - \log(1-0.75/d))}$. In particular, when the dimension $d \to \infty$, the lower bound shown in \eqref{LowerBoundNaive} turns out to be $n^{-0.75/(0.75+d\log 2)}$. Note that if $d\geq 6$, then the upper bound of the convergence rate \eqref{UpperBoundEnsemble} for GBBH will be smaller than the lower bound \eqref{LowerBoundNaive} for binary histogram regression, which explains the benefits of the boosting procedure.
\end{remark}

\subsection{Convergence Rates for GBBHE in $C^{0,\alpha}$}\label{sec::c0large}

\begin{theorem}\label{thm::c01gbbhte}
Let $\bar{f}_{\mathrm{D},B}$ be the GBBHE regressor defined by \eqref{equ::barfdlambda}. Moreover, suppose that the Bayes decision function $f_{L, \mathrm{P}}^* \in C^{0,\alpha}$ and $\mathrm{P}_X$ is the uniform distribution on  $B_{r,d} = [-r/\sqrt{d}, r/\sqrt{d}]^d$. Furthermore, let $\{ \lambda_{1,n} \}$, $\{ \lambda_{2,n} \}$ and $\{ p_n \}$ be chosen as
\begin{align*}
\lambda_{1,n} = n^{-\frac{2(1-4^{-\alpha})}{(4-2\delta)(1-4^{\alpha})+2d\log 2}}, \,
\lambda_{2,n} = n^{-\frac{2(1-4^{-\alpha}+2d\log 2)}{(4-2\delta)(1-4^{-\alpha})+2d\log 2}}, \,
p_n \asymp \frac{2d\log 2\log n}{(4-2\delta)(1-4^{-\alpha})+2d\log 2},
\end{align*}
where $\delta := 1/(d\cdot 2^p)$. Then for sufficiently large $n$, there holds
\begin{align*}
\mathbb{E}_{\mathrm{P}_{R,Z}} \bigl( \mathcal{R}_{L, \mathrm{P}}(\bar{f}_{\mathrm{D},B}) - \mathcal{R}_{L,\mathrm{P}}^* \bigr) 
\lesssim  n^{-\frac{2(1-4^{-\alpha})}{(4-2\delta)(1-4^{-\alpha})+2d\log 2}}
\end{align*}
with probability $\mathrm{P}^n$ equal to one.
\end{theorem}

In particular, if $f^*_{L,\mathrm{P}} \in C^{0,1}$,  then for sufficiently large $n$, there holds
\begin{align*}
\mathbb{E}_{\mathrm{P}_{R,Z}} \bigl( \mathcal{R}_{L, \mathrm{P}}(\bar{f}_{\mathrm{D},B}) - \mathcal{R}_{L,\mathrm{P}}^* \bigr) 
\lesssim  n^{-\frac{3}{3(2-\delta)+4d\log 2}}
\end{align*}
with probability $\mathrm{P}^n$ equal to one.

\begin{remark}[Convergence rate]
When $f_{L, \mathrm{P}}^* \in C^{0,\alpha}$, we derive the convergence rate of GBBHE with probability one. In particular, when $\alpha = 1$, the convergence rate turns out to be of the order $n^{-3/(6+4d\log 2)}$. We notice that GBBHE attains the same convergence rate with the same parameter selections of $\lambda$ and $p$ as GBBH in the space $C^{0,1}$.
\end{remark}

\subsection{Convergence Rates for GBBHE in $C^{1,0}$}\label{sec::c1large}

\begin{theorem}\label{thm::optimalForestlarge}
Let $\bar{f}_{\mathrm{D},B}$ be the GBBHE regressor defined by \eqref{equ::barfdlambda}. Moreover, suppose that the Bayes decision function $f_{L, \mathrm{P}}^* \in C^{1,0}$ and $\mathrm{P}_X$ is the uniform distribution on  $B_{r,d}=[-r/\sqrt{d},r/\sqrt{d}]^d$. Furthermore, let $\{ \lambda_{1,n} \}$, $\{ \lambda_{2,n} \}$, $\{T_n\}$, $\{K_n\}$ and $\{ p_n \}$ be chosen as
\begin{align*}
\lambda_{1,n} := n^{-\frac{3}{4(2-\delta+d\log 2)}}, \,
\lambda_{2,n} := n^{-\frac{2\log 2d+1}{2-\delta+d\log 2}}, \,  
T_n K_n := n^{\frac{1}{4(2-\delta+d\log 2)}}, \, 
p_n \asymp \frac{d\log n}{2-\delta+d\log 2 },
\end{align*}
where $\delta :=1/(d\cdot 2^p)$. Then, for sufficiently large $n$, there holds
\begin{align} \label{UpperBoundEnsemblelarge}
\mathbb{E}_{\mathrm{P}_{R,Z}}\mathcal{R}_{L,\mathrm{P}}(\bar{f}_{\mathrm{D},B}) - \mathcal{R}_{L,\mathrm{P}}^* 
\lesssim n^{-\frac{1}{2-\delta+d\log 2}}
\end{align}
with probability $\mathrm{P}^n$ equal to one.
\end{theorem}

\begin{remark}[Convergence rate]
When $f_{L, \mathrm{P}}^* \in C^{1,0}$, GBBHE attains the asymptotically convergence rate $n^{-1/(2+d\log 2)}$, which is the same as that of GBBH.
\end{remark}

\begin{remark}[Effect of ensemble]
Recall that for GBBH in Theorem \ref{sec::c1}, to achieve the convergence rate, we require the number of iterations $T_n$ to be of the order $n^{1/(8+4d\log2)}$.
On the other hand, for GBBHE, to achieve the same convergence rate, we require that $T_nK_n$ is of the order $n^{1/(8+4d\log2)}$, which indicates that we can reduce the number of iterations $T$ by enlarging the number of base learners for ensemble $K$.
Note that boosting algorithms with a large number of iterations can be quite time-consuming, while acceleration techniques such as parallel computing is not directly available.
On the contrary, by combining ensemble methods, which enjoy high computational efficiency with the help of parallel computing, we reduce the running time of GBBHE by reducing the number of iterations $T$.
\end{remark}

\begin{remark}[Inclusive framework]
Theorem \ref{thm::optimalForestlarge} applies to both pure gradient boosting algorithm and pure ensemble algorithm.
To be specific, when $K=1$, the Theorem \ref{thm::optimalForestlarge} corresponds to Theorem \ref{thm::tree}.
When $T=1$, Theorem \ref{thm::optimalForestlarge} shows the convergence rate for the algorithm of ensembling the base learners.
\end{remark}

\section{Numerical Experiments} \label{sec::numerical}

In this section, we conduct numerical experiments including parameter analysis and comparisons with other state-of-the-art large-scale regression algorithms. Aiming at empirically evaluating the large-scale application of our algorithm, we conduct the experiments following Algorithm \ref{alg::GBBHE}. Yet in this part, the split points in the binary histogram partition are selected as the mean point of data from a randomly selected dimension, since the support of real data is usually unknown in the high-dimensional space. We mention that in this way our proposed GBBHE actually enjoys more adaptivity to various datasets. This section is organized as follows. In Section \ref{sec::parameter}, we conduct parameter analysis of the four important parameters in Algorithm \ref{alg::GBBHE}. Then in Section \ref{sec::numcompare}, we compare our GBBHE with other state-of-the-art methods for large-scale regression on moderate-sized and large-scale real datasets.

\subsection{Parameter Analysis}\label{sec::parameter}

In this subsection, we apply parameter analysis to explore the effects of hyper-parameters in the proposed gradient boosted binary histogram transform ensemble (GBBHE) algorithm. In this part, we conduct experiments with GBBHE without random rotation. The reasons are bifold. Firstly, here we are mainly interested in the behavior of the binary histogram partition, the boosting procedure, and the ensemble. Secondly and more importantly, rotations introduce extra randomness that may affect the performance of the algorithm. Therefore, to conduct parameter analysis, we control this factor and let the transformation $H$ be the identity matrix. The data set we used for parameter analysis is the \textit{Physicochemical Properties of Protein Tertiary Structure Data Set} (${\tt PTS}$) from the UCI machine learning repository \citep{Dua:2019}, which contains $45,730$ samples of dimension $9$.

There are four hyper-parameters to be discussed, including the number of iteration $T$, the learning rate $\rho$, the number of binary histograms in each iteration $K$, and the depth of binary histograms $p$. We randomly split the $ {\tt PTS}$ data set into $70\%$ training set and $30\%$ testing set and repeat the experiments for $50$ times in each hyper-parameter setting.

\subsubsection{Regularization under Different Numbers of Binary Histograms}

\begin{figure}[htbp]
\centering
\vskip 0.0in
\includegraphics[width=0.45\linewidth]{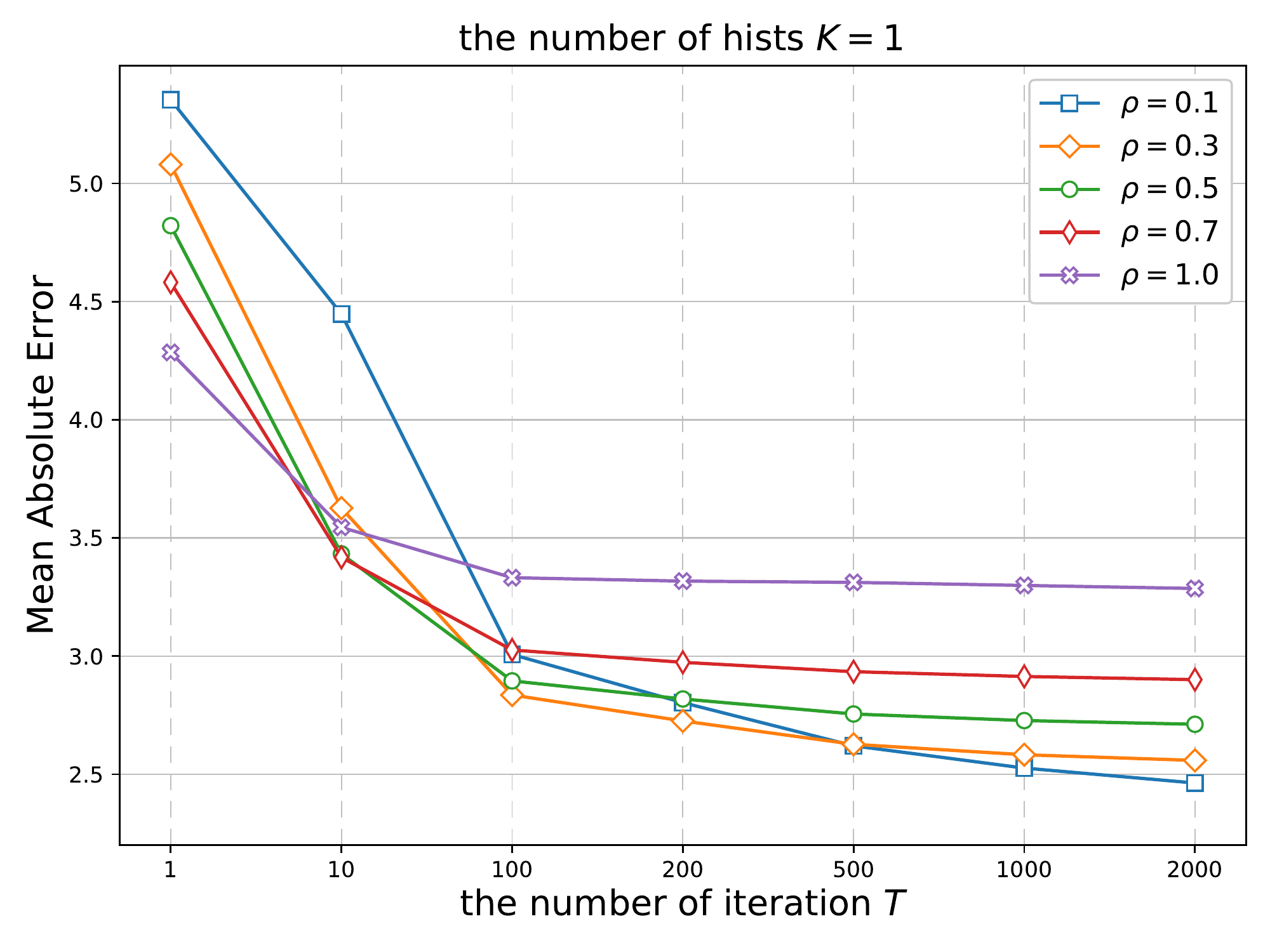}
\includegraphics[width=0.45\linewidth]{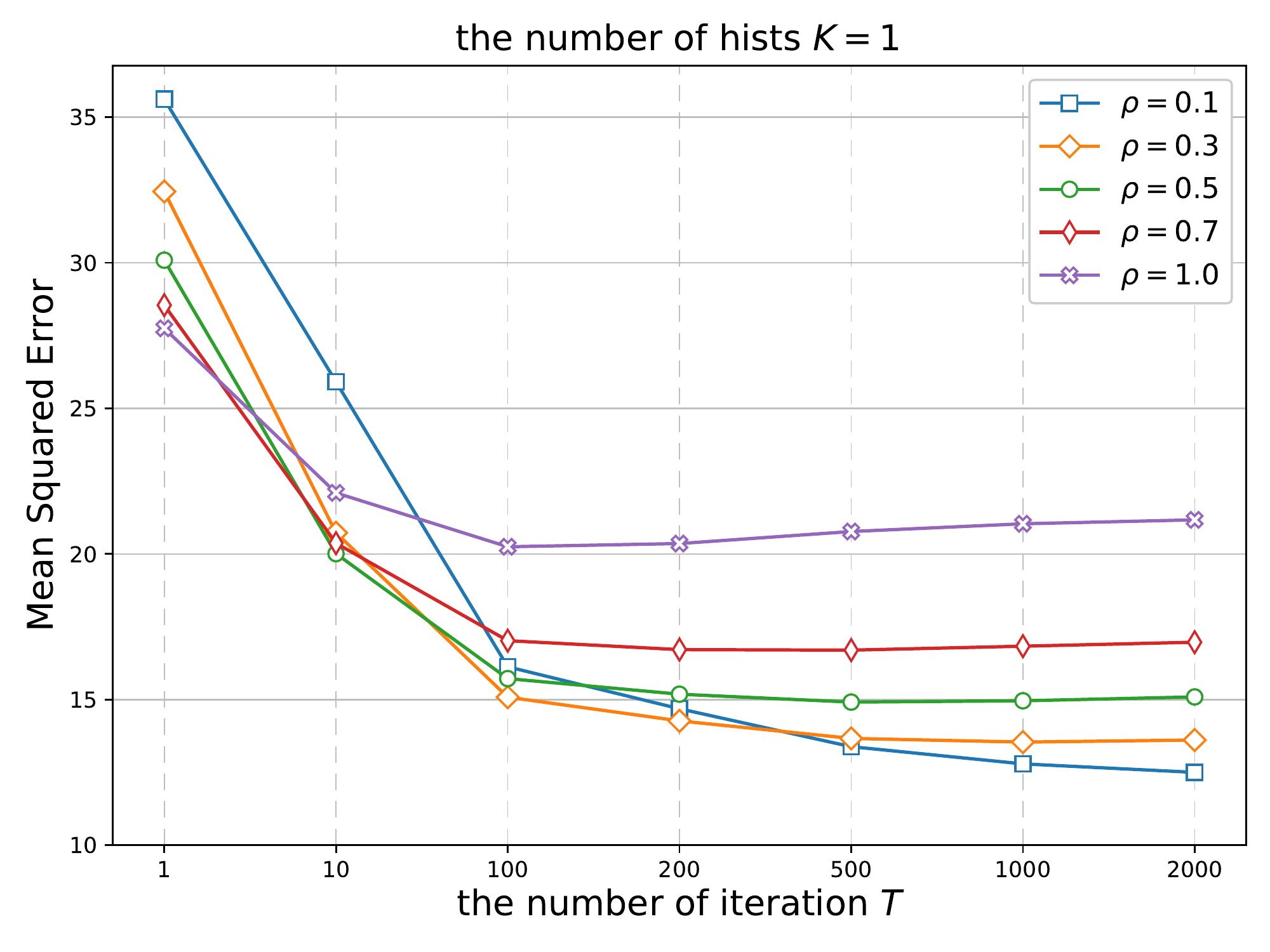}
\vskip 0.0in
\captionsetup{justification=centering}
\caption{Test error curves when the number of binary histograms $K = 1$.}
\label{fig::large-param1_t1}
\vskip 0.0in
\end{figure}

\begin{figure}[htbp]
\centering
\vskip 0.0in
\includegraphics[width=0.45\linewidth]{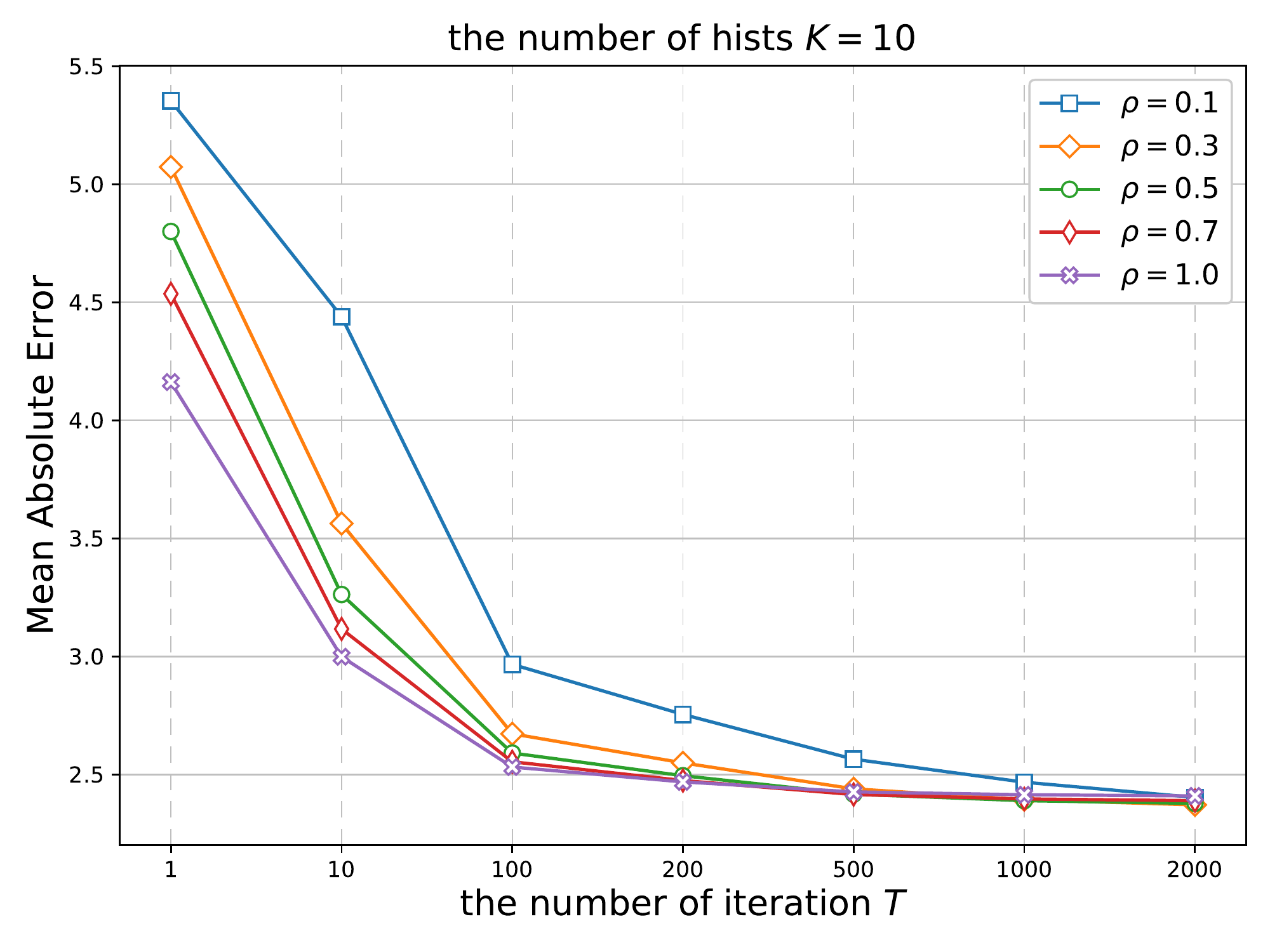}
\includegraphics[width=0.45\linewidth]{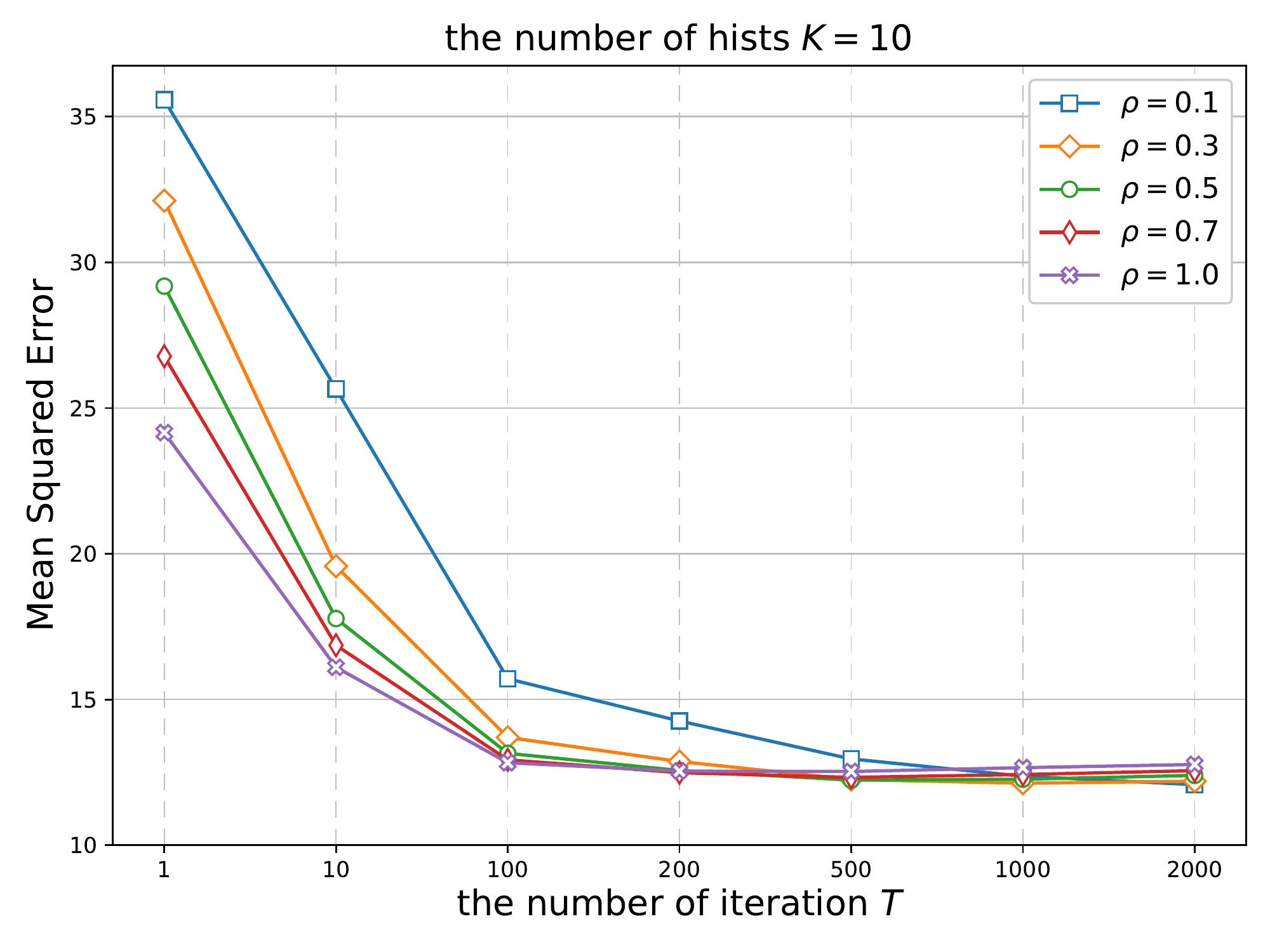}
\vskip 0.0in
\captionsetup{justification=centering}
\caption{Test error curves when the number of binary histograms $K = 10$.}
\label{fig::large-param1_t10}
\vskip 0.0in
\end{figure}

Firstly, we would like to show our surprising finding that ensemble of binary histograms in each iteration of the proposed GBBHE can greatly boost the numerical performance with easier hyper-parameter selection in the empirical aspect. To illustrate this, we vary the number of iteration $T \in \{1, 10, 100, 200, 500, 1000, 2000 \}$ and the learning rate $\rho \in \{0.1, 0.3, 0.5, 0.7, 1.0 \}$, fix the depth of binary histograms $p= 8$, and plot the test error curves under different numbers of binary histograms $K = 1$ and $K = 10$. Results are shown in Figure \ref{fig::large-param1_t1} and Figure \ref{fig::large-param1_t10} respectively.

In Figure \ref{fig::large-param1_t1}, only when the number of iteration $T$ is large enough and the learning rate $\rho$ is small enough does the boosted rotated binary histograms without ensemble in each iteration have good performance. This verifies the common phenomenon that there is a trade-off between the number of iterations $T$ and the learning rate $\rho$ in gradient boosting. However, it is very difficult to determine the optimal $T$ and $\rho$ because the trade-off is quite sensitive when the number of hists $K = 1$.

By contrast, when ensemble is introduced in each iteration of boosted rotated binary histograms, e.g., $K = 10$, the trade-off between the number of iteration $T$ and the learning rate $\rho$ is much less sensitive, see Figure \ref{fig::large-param1_t10} in detail. Although the performance of GBBHE will slightly worse when $T$ and $\rho$ become too large, i.e, the trade-off between the number of iteration $T$ and the learning rate $\rho$ illustrated in Figure \ref{fig::large-param1_t1} also exists, we find that the performance is satisfactory among a wide range of $T$ and $\rho$. Moreover, it is well worth mentioning that the boosting performance when $K = 10$ converges faster than that of $K = 1$ compared with Figure \ref{fig::large-param1_t1}, resulting in better performance. This exactly corresponds to the theoretical result in Theorem \ref{thm::optimalForestlarge} that ensemble can reduce the number of iterations $T$ to achieve the fast convergence rate. To conclude, the introduced ensemble with $K = 10$ not only stabilizes the hyper-parameter selection of $T$ and $\rho$, but also boosts the convergence of the boosting algorithm.


\subsubsection{Parameter Analysis about the Learning Rate}\label{sec::param_rho}

Secondly, we would like to discuss the tendency of the learning rate $\rho$ among different number of iterations $T$. 
We fix the number of binary histograms $K = 10$ and the depth of binary histograms $p = 8$, and then vary the choice of the learning rate $\rho \in \{0.1, 0.3, 0.5, 0.7, 1.0\}$ among four different $T \in \{200, 500, 1000, 2000\}$.

\begin{figure}[htbp]
\centering
\vskip 0.0in
\includegraphics[width=0.45\linewidth]{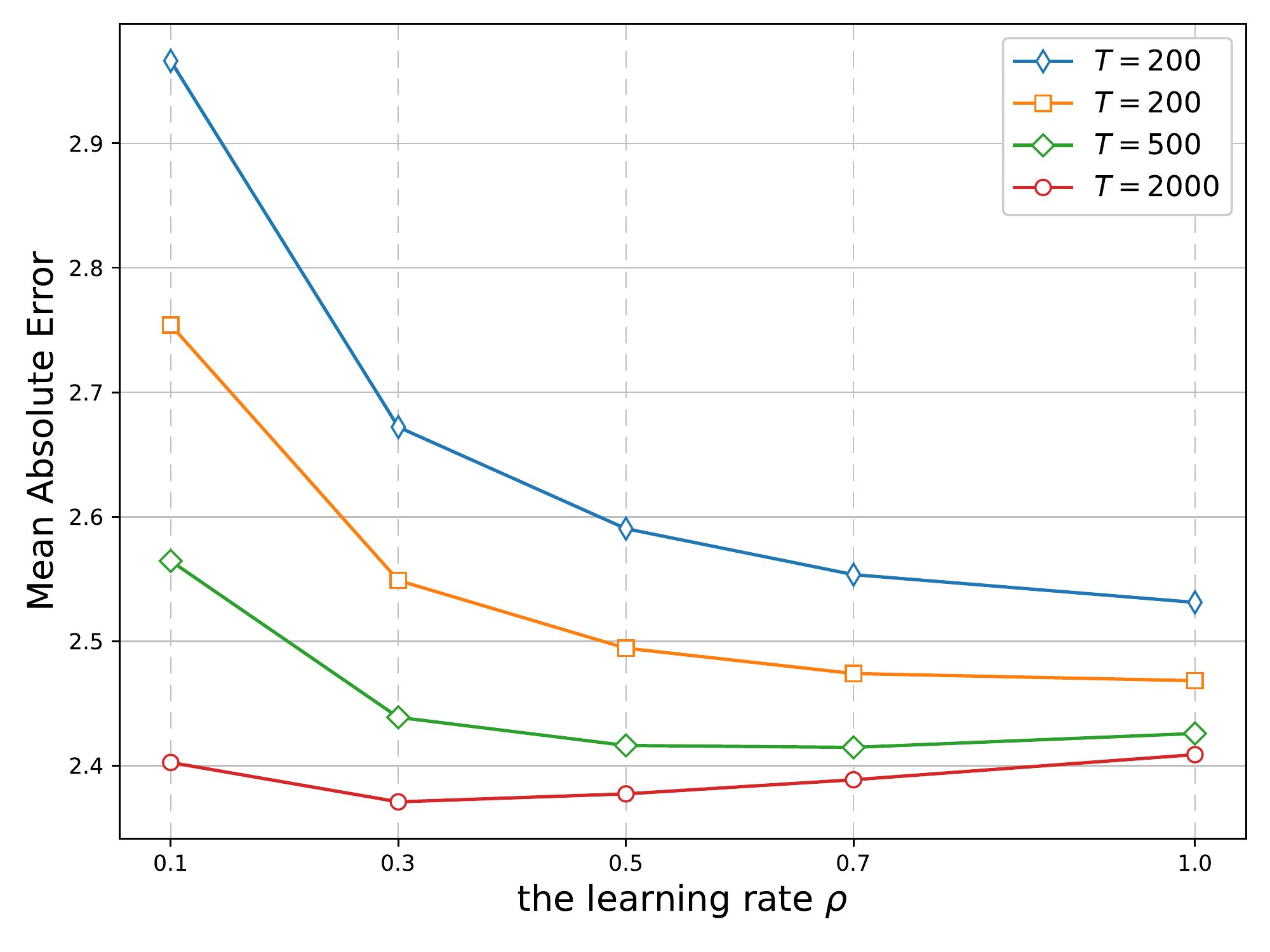}
\includegraphics[width=0.45\linewidth]{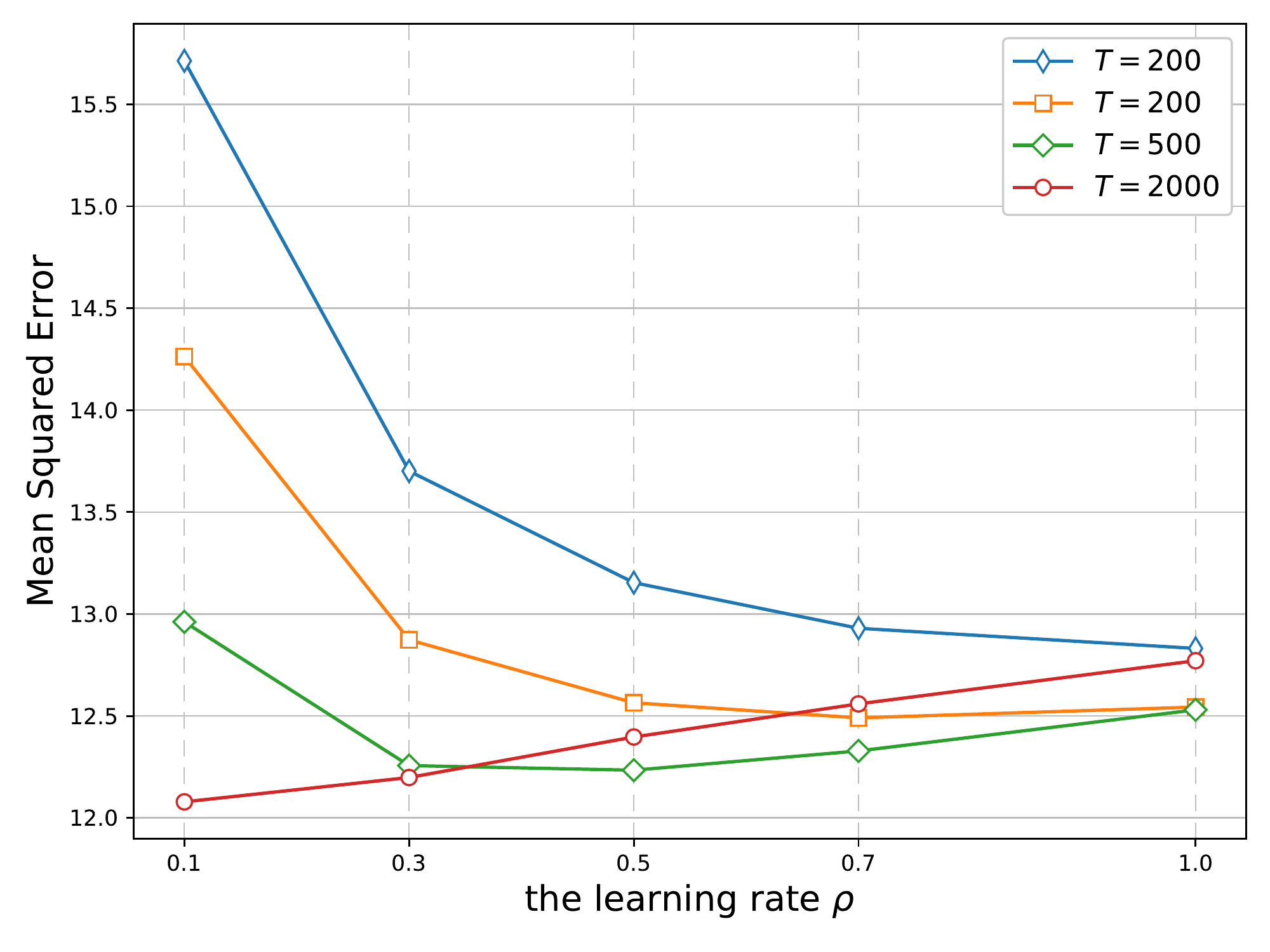}
\vskip 0.0in
\captionsetup{justification=centering}
\caption{Test error curves among various $\rho$ under fixed $K=10$ and $p = 8$.}
\label{fig::large-param_lr}
\vskip 0.0in
\end{figure}

From Figure \ref{fig::large-param_lr} we see that the test error decreases to an optimal value and then increases as the learning rate $\rho$ increases from $0.1$ to $1.0$. Besides, we observe that the optimal $\rho$ varies by different numbers of iterations. 
To be specific, smaller $T$ leads to larger optimal $\rho$, while the optimal $\rho$ goes down with larger $T$. 
For example, the optimal $\rho$ when $T=200$ is $1.0$, while the optimal $\rho$ when $T = 2000$ is $0.3$.
Moreover, as the number of iteration $T$ goes up, the test error under optimal $\rho$ consistently decreases, which shows empirically that larger $T$ with smaller $\rho$ yields better numerical performance.

\subsubsection{Parameter Analysis about the Number of Binary Histograms}

Secondly, we discuss the choice of the number of binary histograms $K$, which controls how ensemble is used in the algorithm GBBHE: larger $K$ means more binary histograms built in each iteration of boosting. To be specific, we fix the number of iterations $T = 100$ and the depth of binary histograms $p = 8$, and then vary $K \in \{1, 2, 5, 10, 20, 50, 100, 200\}$ among different learning rates $\rho \in \{0.1, 0.3, 0.5, 0.7, 1.0\}$.

\begin{figure}[!h]
\centering
\vskip 0.0in
\includegraphics[width=0.45\linewidth]{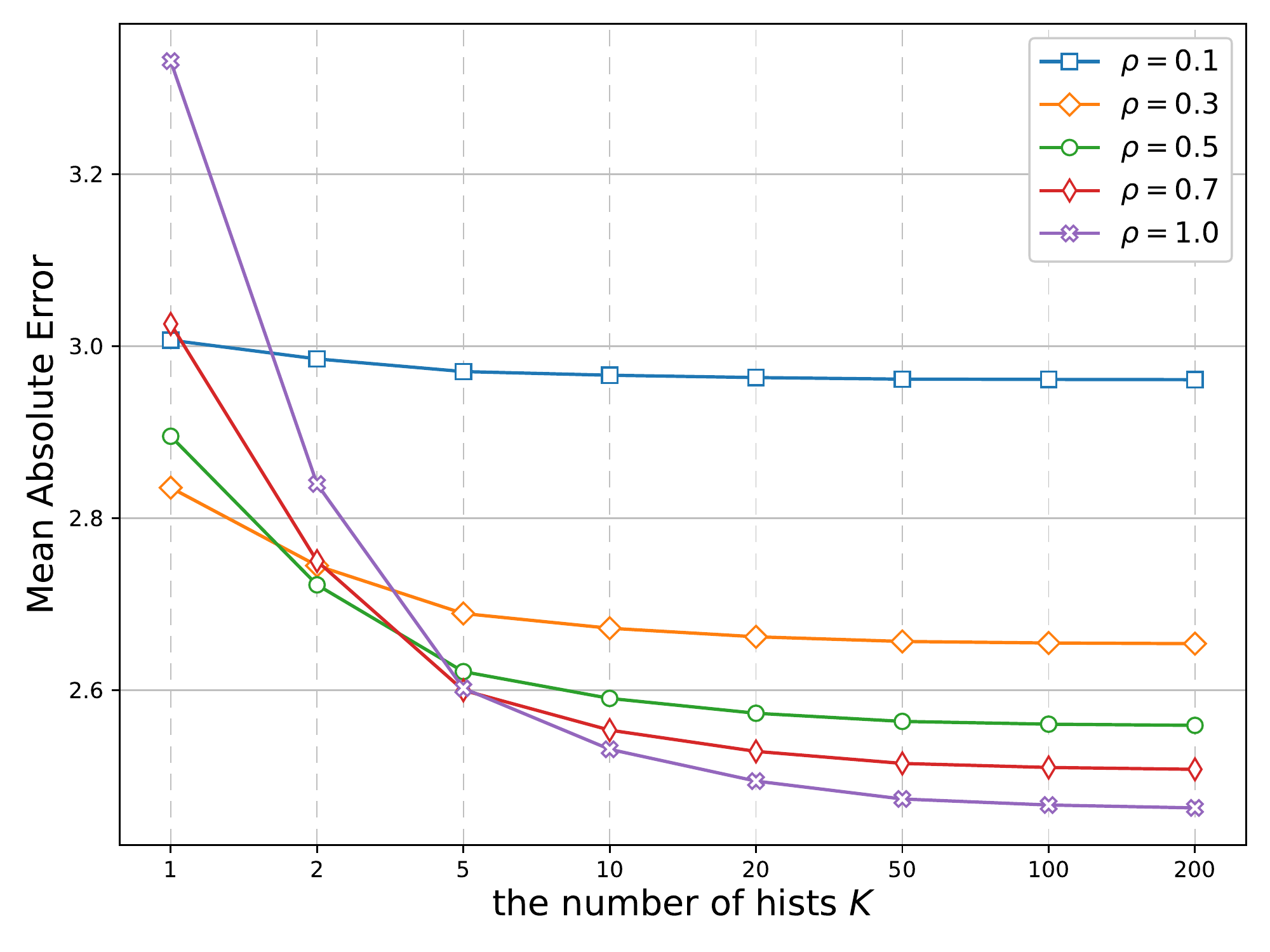}
\includegraphics[width=0.45\linewidth]{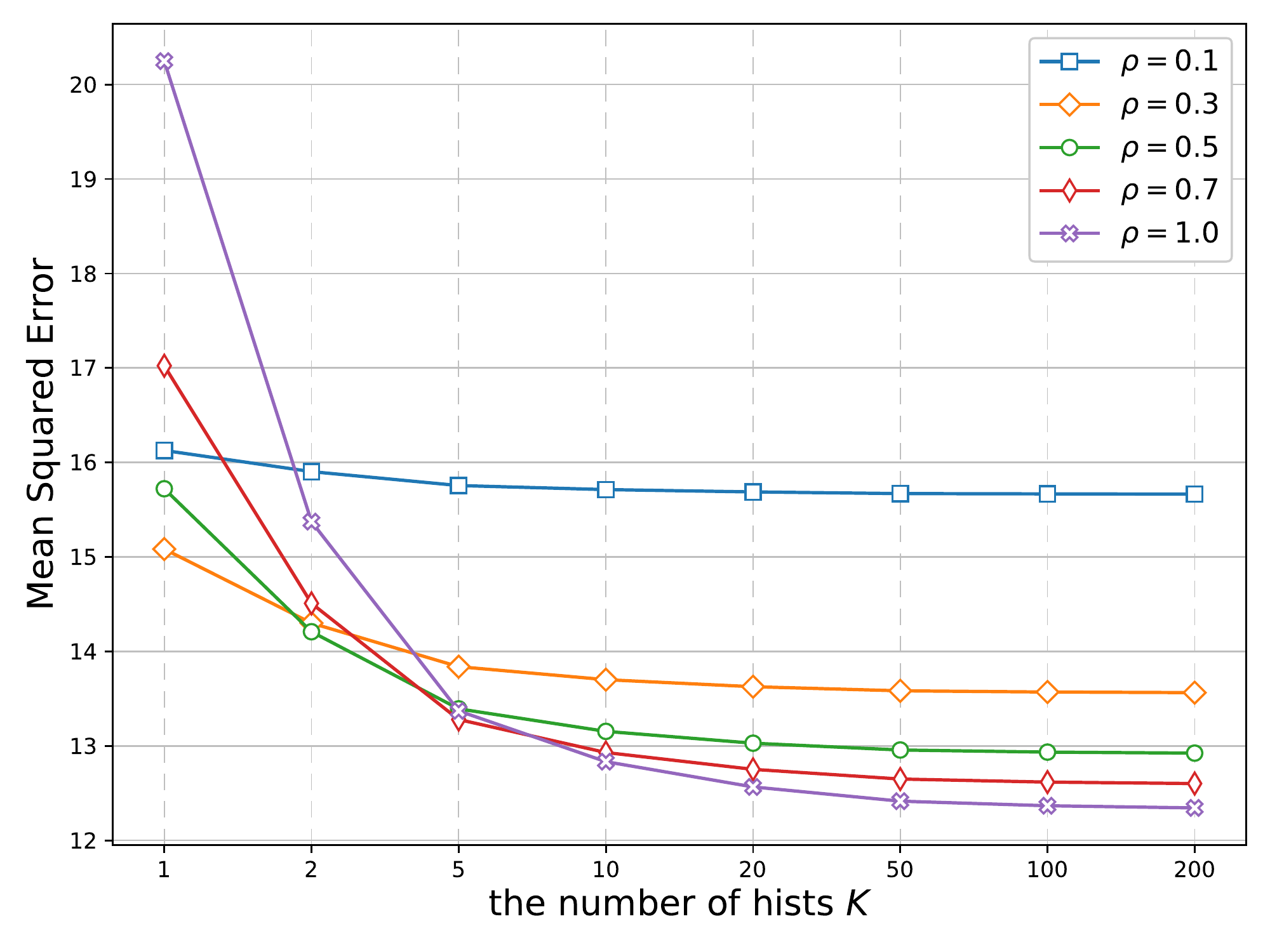}
\vskip 0.0in
\captionsetup{justification=centering}
\caption{Test error curves among various $K$ under fixed $T=100$ and $p = 8$.}
\label{fig::large-param_K_1}
\vskip 0.0in
\end{figure}

\begin{figure}[!h]
\centering
\vskip 0.0in
\centerline{\includegraphics[width=0.45\linewidth]{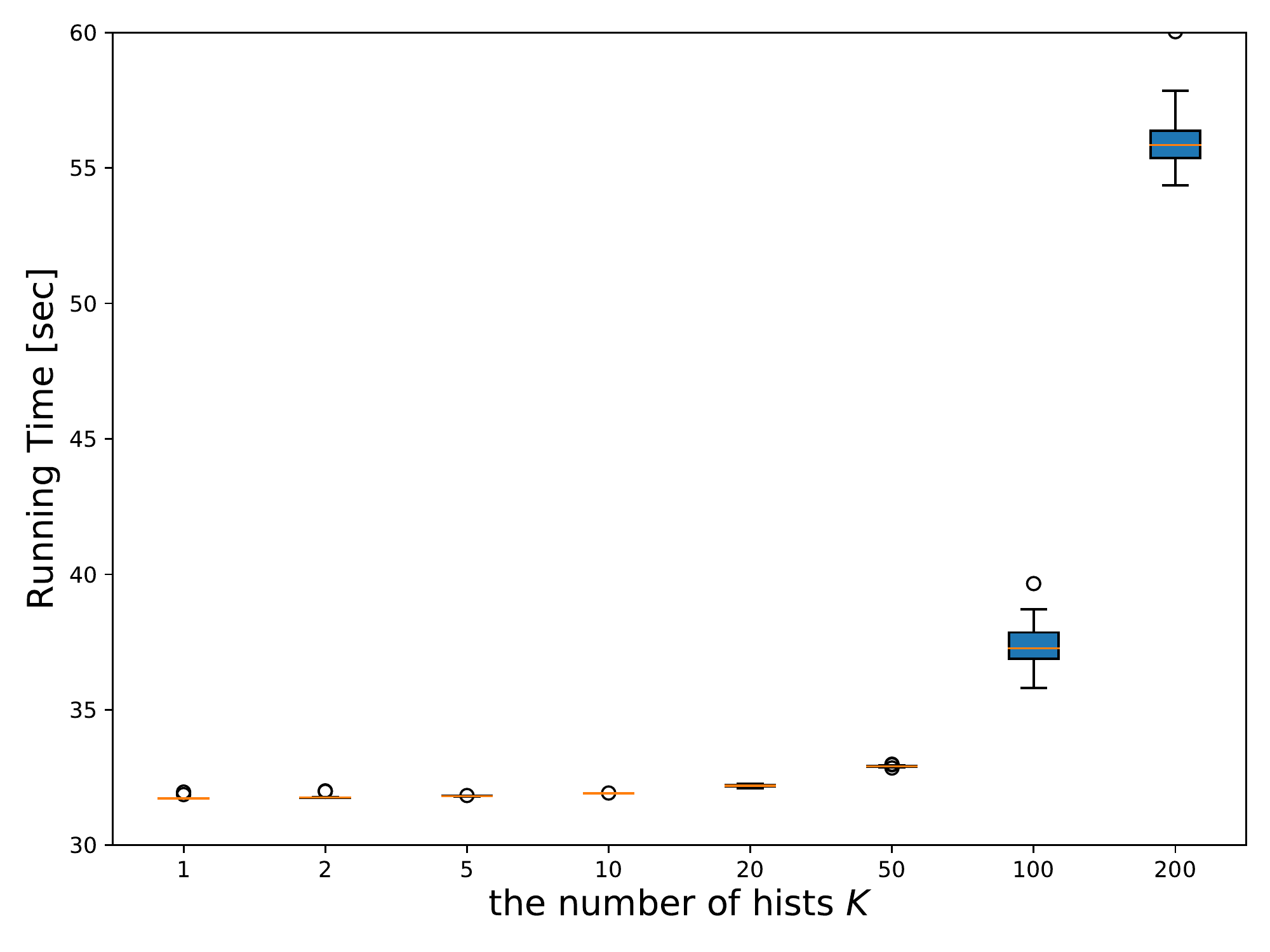}}
\vskip 0.0in
\captionsetup{justification=centering}
\caption{Running time among various $K$ under fixed $T=100$ and $p = 8$.}
\label{fig::large-param_K_2}
\vskip 0.0in
\end{figure}

Figure \ref{fig::large-param_K_1} and Figure \ref{fig::large-param_K_2} show the accuracy performances and the running times under different numbers of binary histograms $K$ respectively. As is shown, on the one hand, as $K$ goes up, the performances of regression continue to rise, only with the cost of higher computation times. On the other hand, the marginal increase of performance drops as $K$ becomes larger. This phenomenon is consistent regardless of the learning rate $\rho$ varies.
Therefore, considering the trade-off between accuracy and running time, $K=100$ shows a satisfactory performance with tolerable running time. Moreover, Figure \ref{fig::large-param_K_1} suggests that ensemble is especially beneficial for large learning rate $\rho$. This to some extent corresponds to the numerical result shown in Section \ref{sec::param_rho} that ensemble helps improve performance and the stability of $\rho$.


\subsubsection{Parameter Analysis about the Depth of Binary Histograms}

One more hyper-parameter we need to discuss is the depth of binary histograms $p$. The depth $p$ controls the denseness of binary histogram partition: the total cells of each binary histogram is $2^p$, and  larger $p$ means more partitions for each binary histogram. To numerically analyze the effect of different depths of binary histograms $p$, we fix the number of iterations $T = 100$ and the number of binary histograms in each iteration $K = 100$, and vary the depth $p \in \{2, 4, 6, 8, 10, 12, 14\}$ among different learning rate $\rho \in \{0.1, 0.3, 0.5, 0.7, 1.0\}$.

\begin{figure}[!h]
\centering
\vskip 0.0in
\includegraphics[width=0.45\linewidth]{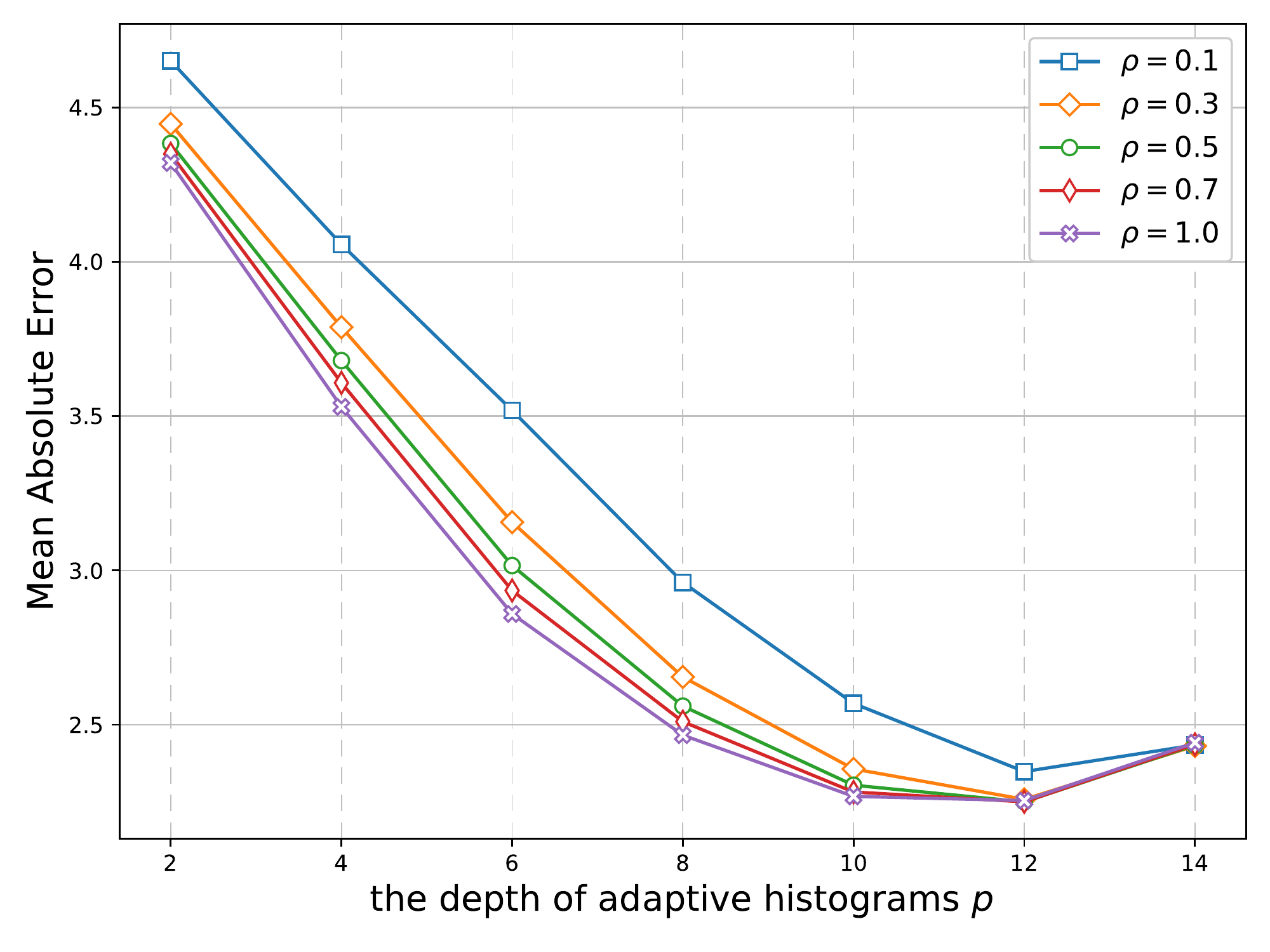}
\includegraphics[width=0.45\linewidth]{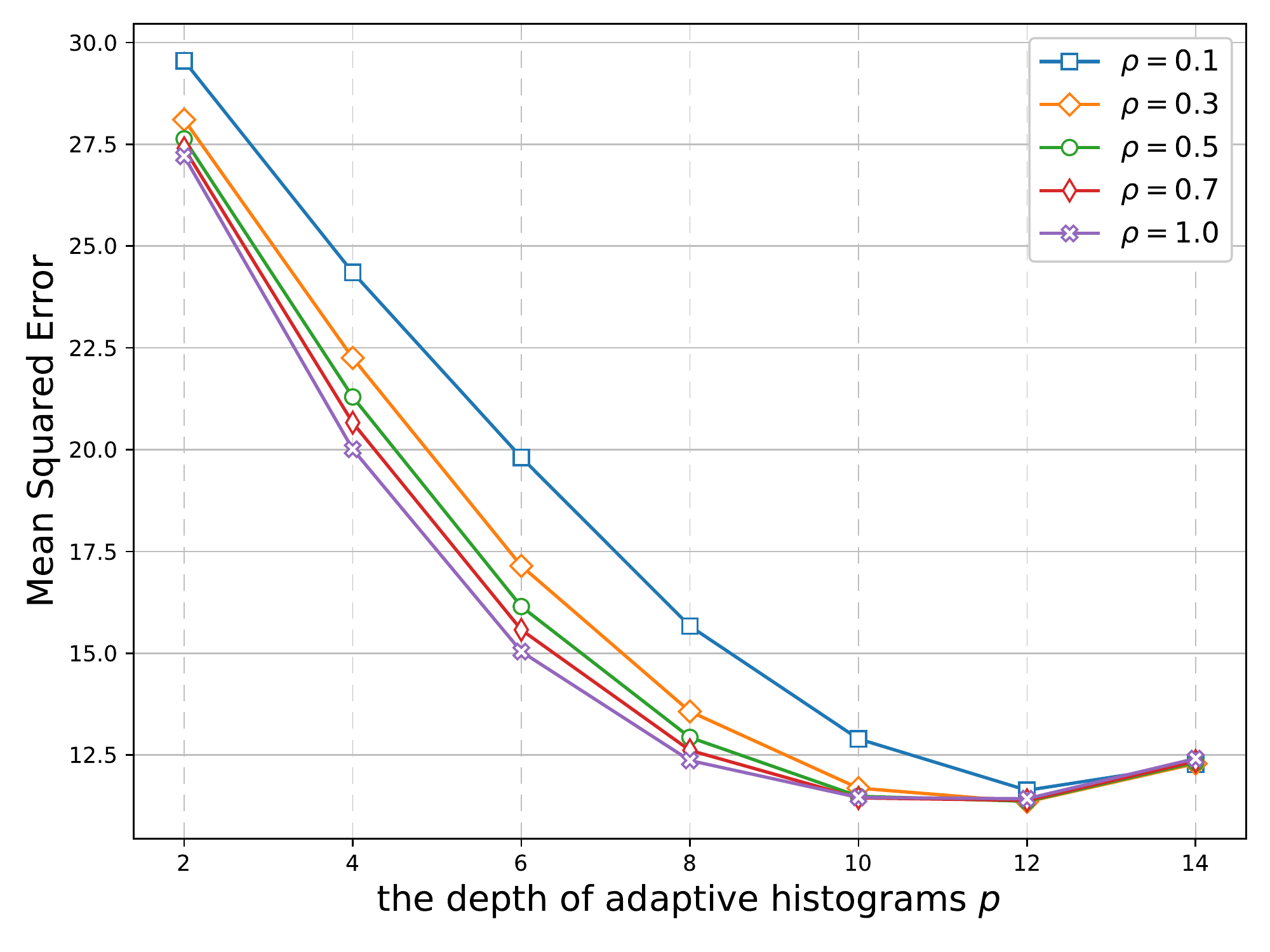}
\vskip 0.0in
\captionsetup{justification=centering}
\caption{Test error curves among various $p$ under fixed $T=100$ and $K=100$.}
\label{fig::large-param_p_1}
\vskip 0.0in
\end{figure}

\begin{figure}[!h]
\centering
\vskip 0.0in
\centerline{\includegraphics[width=0.45\linewidth]{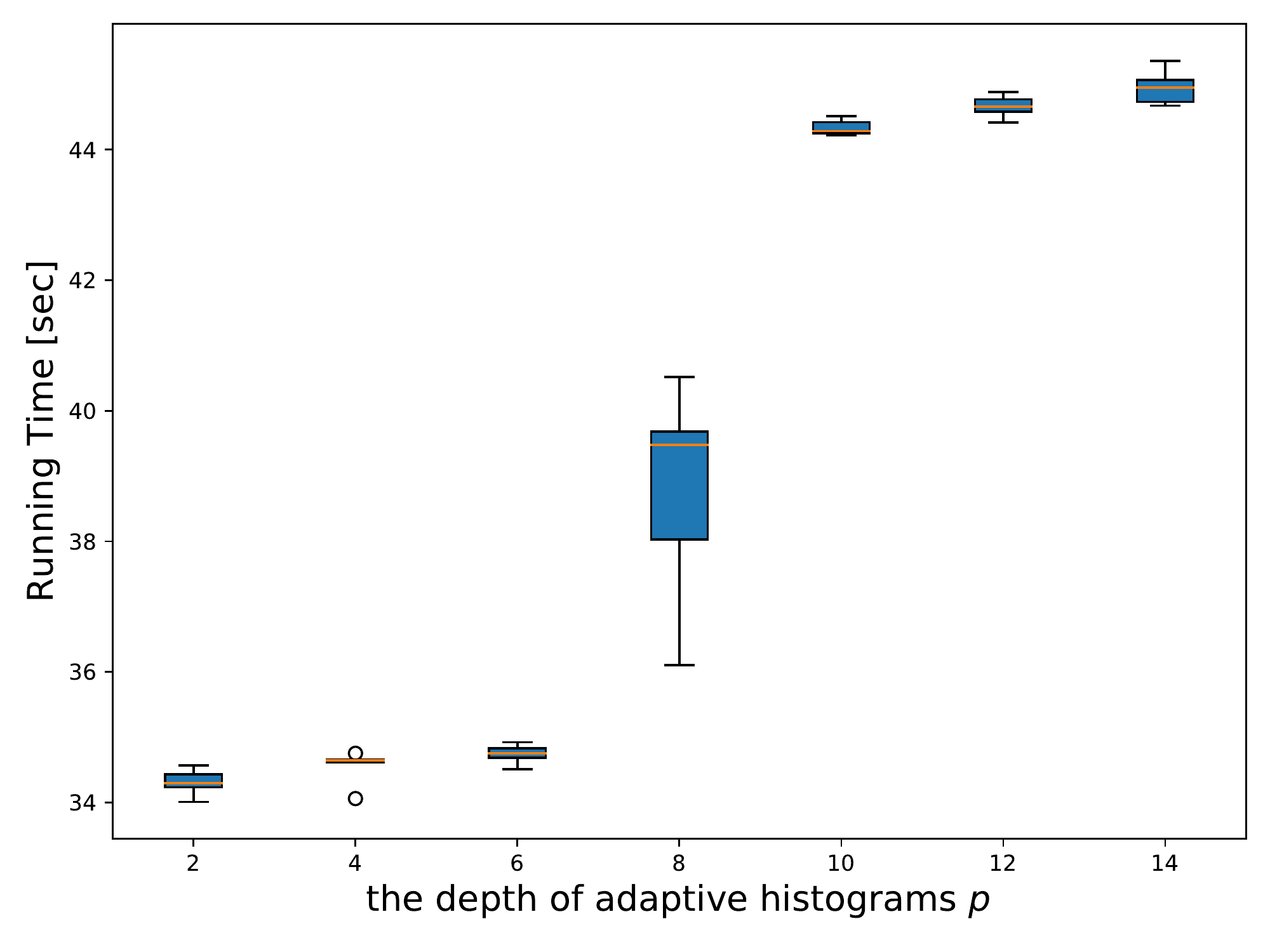}}
\vskip 0.0in
\captionsetup{justification=centering}
\caption{Running time among various $p$ under fixed $T=100$ and $K=100$.}
\label{fig::large-param_p_2}
\vskip 0.0in
\end{figure}

Figure \ref{fig::large-param_p_1} and Figure \ref{fig::large-param_p_2} show the accuracy performances and the running times under different depths of binary histograms $p$ respectively. As the depth increases, the accuracy performance first goes up and reaches its optimum, and then deteriorates when the depth of binary histograms becomes too large, with the running time monotonically increasing. This phenomenon is consistent regardless of the learning rate $\rho$ varies. It may attribute to the fact that splitting more in each binary histogram helps to build up binary histograms with local adaptivity, but if $p$ is too large, the partitions of binary histograms will be too dense and it is at the risk of over-fitting. Besides, it would be too time-consuming to adopt such large $p$. This also verifies the theoretical results in Theorem \ref{thm::c01gbbhte} and \ref{thm::optimalForestlarge} that there exists an optimal order of depth $p$.

\subsection{Numerical Comparisons}\label{sec::numcompare}

In this subsection, we conduct numerical studies to evaluate the effectiveness of our proposed Gradient Boosting Binary Histogram Transforms Ensemble (GBBHE) algorithm on moderate and large-scale real data sets.
In this part, we consider the rotation transform as an option, i.e., we compare both GBBHE without random rotation (GBBHE w.o.~Rotation) and standard GBBHE with random rotation (GBBHE w. Rotation) as is shown in Algorithm \ref{alg::GBBHE} with other state-of-the-art large-scale regression algorithms.

\subsubsection{Experimental Setups}

Comparisons are conducted among 
\begin{itemize}
\item 
Gradient Boosting Regression Tree (GBRT): Gradient Boosting Regression Trees is a well-known ensemble method for regression proposed by \citet{friedman2001greedy}. However, different from random forest, trees in GBRT are fitted in a sequential manner. We use the {\tt scikit-learn} implementation in {\tt Python}. Three hyper-parameters are considered, including the number of iteration times $T$, learning rate shrinking the contribution of each tree $\rho$, and the minimum number of samples required to split an internal node {\tt min\_samples\_split}. We use a validation set containing $10\%$ of the training samples to find the best $T \in \{100, 200\}$, $\rho \in \{0.1, 0.2, \ldots, 0.9, 1.0 \}$, and ${\tt min\_samples\_split} \in \{2, 5, 10, 20, 50, 100, 200, 500\}$.
\item 
Random Forest: Random forest is a well-known ensemble algorithm for regression proposed by \citet{breiman2001random}. As each tree is fitted in parallel, random forest is naturally suitable in large-scale applications. We use the {\tt scikit-learn} implementation in {\tt Python}. Two hyper-parameters are the number of estimators in the forest $T$ and the minimum number of samples required to split an internal node {\tt min\_samples\_split}. We use a validation set containing $10\%$ of the training samples to find the best $T \in \{100, 200, 500\}$ and ${\tt min\_samples\_split} \in \{2, 5, 10, 20, 50, 100, 200, 500\}$.
\item 
LiquidSVM: Support vector machines for regression being a global algorithm is impeded by super-linear computational requirements in terms of the number of training samples in large-scale applications. To address this, \citet{meister16a} employs a spatially oriented method to generate the chunks in feature space, and fit LS-SVMs for each local region using training data belonging to the region. This is called the Voronoi partition support vector machine (VP-SVM). We use the implementation provided by the authors in {\tt Python}. In each local region, we conduct five-fold cross-validation (CV) to find the best hyper-parameters $C$ and $\epsilon$.
\end{itemize}

As for the hyper-parameter tuning of our method, we use a validation set to find the number of iterations $T \in \{100, 200, 500\}$, the number of binary histograms $K \in \{10, 100, 200\}$, best learning rate $\rho \in \{ 0.1, 0.3, 0.5, 0.7, 1.0\}$ and the best depth parameter $p \in \{4, 6, 8, 10, 12\}$.

For moderate-sized datasets, we also randomly split $70\%$ of the data set for training and the other $30\%$ for testing. 
We scale each feature individually to the range $[0,1]$ on the training set. Hyper-parameters are selected by validation and experiments are repeated $50$ times. For the large-scale data sets, the partition of the training set and the testing set is specified in their detailed descriptions in Section \ref{sec::datadescript}, and we repeat the experiments $20$ times. We record and summarize the average and the standard deviation of the mean squared error, the mean absolute error, and the running time under the best hyper-parameter setting over all experiments. All experiments are conducted on a 64-bit machine with 40-cores Intel Xeon 2.0GHz CPU (E5-4620) and 256GB main memory.

\subsubsection{Description of Datasets}\label{sec::datadescript}

We use for the evaluation seven moderate-sized datasets and five well-known large-scale datasets from the UCI machine learning repository \citep{Dua:2019}, LIBSVM Data\footnote{LIBSVM Data: Classification, Regression, and Multi-label. \url{https://www.csie.ntu.edu.tw/~cjlin/libsvmtools/datasets/}}, and Delve Datasets\footnote{Delve Datasets: Collections of data for developing, evaluating, and comparing learning methods. \url{https://www.cs.toronto.edu/~delve/data/datasets.html}}. Details of these data sets, including size and dimension, are summarized in Table \ref{tab::Realdata_discription}.

\begin{table}[H] 
\setlength{\tabcolsep}{9pt}
\centering
\captionsetup{justification=centering}
\caption{\footnotesize{Description over Real Data Sets}}
\label{tab::Realdata_discription} 
\begin{tabular}{c|c|c|c}
\toprule
& \text{datasets} & size & dimension \\
\midrule
\midrule
\multirow{7}*{Moderate-sized} & {\tt  EGS } & $10,000$ & $12$ \\
& {\tt  AEP } & $19,735$ & $27$ \\
& {\tt  CAD } & $20,640$ & $8$ \\
& {\tt  SCD } & $21,263$ & $81$ \\
& {\tt  HPP } & $22,784$ & $8$ \\
& {\tt  ONP }  & $39,644$ & $58$ \\
& {\tt  PTS } & $45,730$ & $9$ \\
\hline
\multirow{5}*{Large-scale} 
& {\tt  MSD } & $515,345$ & $90$ \\
& {\tt  BUZ } & $583,250$ & $77$ \\
& {\tt  GHG } & $955,167$ & $15$ \\
& {\tt  GTM } & $3,843,160$ & $18$ \\
& {\tt  DGM } & $4,208,261$ & $16$ \\
\bottomrule
\end{tabular}
\end{table}

\begin{itemize}
\item {\tt EGS}: 
The \textit{Electrical Grid Stability Simulated Data Set} ({\tt EGS}) \citep{arzamasov2018towards} is available on the UCI Machine Learning Repository. It contains $10,000$ samples in total. $12$ attributes are used to predict the maximal real part of the characteristic equation root.
\item {\tt AEP}: 
The \textit{Appliances Energy Prediction Data Set} ({\tt AEP}) \citep{candanedo2017data}, available on UCI Machine Learning Repository, contains $19,735$ samples of dimension $27$ with attribute ``date'' removed from the original data set. The data is used to predict the appliances energy use in a low-energy building.
\item {\tt CAD}: The \textit{California Housing Prices Data Set} ({\tt CAD}) is avaliable on the LIBSVM Data. This spacial data can be traced back to \cite{Pace1997Sparse}. It consists $20,640$ observations on housing prices with $8$ economic covariates.
Note that for the sake of clarity, all house prices in the original data set have been modified to be counted in thousands.
\item {\tt SCD}: The \textit{Superconductivity Data Set} ({\tt SCD}) \citep{hamidieh2018data}, available on the UCI Machine Learning Repository, is supported by the NIMS, a public institution based in Japan. This database has $21,263$ samples with $81$ features. The goal is to predict the critical temperature based on the features extracted.
\item {\tt HPP}: The \textit{House Price Prototask Data Set} ({\tt HPP}) is originally taken from the census-house dataset in the DELVE Datasets. We use the house-price-8H prototask, which contains $22,784$ observations. We use $8$ features to predict the median house prices from $1990$ US census data. Similar to the data preprocessing for {\tt CAD}, all house prices in the original data set have been modified to be counted in thousands.
\item {\tt ONP}: The \textit{Online News Popularity Data Set} ({\tt ONP}) \citep{fernandes2015proactive}, available on the UCI Machine Learning Repository, is a database summarizing a heterogeneous set of features about articles published by Mashable in a period of two years. It contains $39,644$ observations with $58$ predictive attributes. This data set is used to predict the number of shares of the online news.
\item {\tt PTS}: \textit{Physicochemical Properties of Protein Tertiary Structure Data Set} (${\tt PTS}$) is available on the UCI Machine Learning Repository. It contains $45,730$ samples of dimension $9$. The regression task is to predict the size of the residue.
\item {\tt MSD}: The \textit{Year Prediction MSD Data Set} ({\tt MSD}) \citep{Bertin-Mahieux2011} is available on the UCI Machine Learning Repository. It contains $463,715$ training samples and $51,630$ testing samples with $90$ attributes, depicting the timbre average and timbre covariance of songs released between the years 1922 and 2011. The main task is to learn the audio features of a song and to predict its release year.
\item {\tt BUZ}: \textit{Buzz in Social Media Data Set} ({\tt BUZ}) \citep{kawala2013predictions} is available on the UCI Machine Learning Repository. It contains examples of buzz events from two different social networks: Twitter, and Tom's Hardware. We select the Twitter part, which contains $583,250$ samples of dimension $77$ and the last column is for prediction. We use the first $450,000$ samples for training, and use the remaining $133,250$ samples for testing.
\item {\tt GHG}: The \textit{Greenhouse Gas Observing Network Data Set} ({\tt GHG}) \citep{lucas2015designing} is available on the UCI Machine Learning Repository. It contains $955,167$ samples of dimension $15$, which are time series of GHG tracers released from $14$ distinct spatial regions in California and one outside of California. The main task is to predict emissions of greenhouse gas. We use the first $800,000$ samples for training, and use the remaining $155,167$ samples for testing.
\item {\tt GTM}: The \textit{Gas Sensor Array Temperature Modulation Data Set} ({\tt GTM}) \citep{burgues2018estimation, burgues2018multivariate} is available on the UCI Machine Learning Repository. We use the readings of $14$ temperature-modulated metal oxide semiconductor gas sensors, a temperature sensor, and a humidity sensor, as well as the values of mass flow rate and the heater voltage to predict the concentration level of CO. The total number of samples is $3,843,160$ and the feature dimension is $18$. We use the first $3,500,000$ samples for training and the remaining $343,160$ samples for evaluation.
\item {\tt DGM}: The \textit{Gas Sensor Array Under Dynamic Gas Mixtures Data Set} ({\tt DGM}) \citep{fonollosa2015reservoir} is available on the UCI Machine Learning Repository. Two gas mixtures are generated in this data set: Ethylene and Methane in air, and Ethylene and CO in air. We select the ethylene-CO mixture in the experimental section, with $4,208,261$ samples in total. We use readings of $16$ chemical sensors to predict the concentration level of CO. In the experiments, we use the first $3,500,000$ samples for training and the remaining $708,261$ samples for testing.
\end{itemize}

\subsubsection{Results}

The comparing results of the average \textit{MSE} and \textit{MAE} on seven moderate datasets {\tt EGS}, {\tt AEP}, {\tt CAD}, {\tt SCD}, {\tt HPP}, {\tt ONP}, {\tt PTS} and five large-scale datasets {\tt MSD}, {\tt BUZ}, {\tt GHG}, {\tt GTM}, and {\tt DGM} are presented in Figures \ref{tab::largescale_mse} and \ref{tab::largescale_mae}. Moreover, we point out that the Wilcoxon test for paired samples with significance level $\alpha = 0.05$ are applied.

\begin{table}[ht] 
\vskip 0.0in
\setlength{\tabcolsep}{8pt}
\centering
\captionsetup{justification=centering}
\caption{\footnotesize{Average \textit{MSE} over Moderate and Large-scale datasets}}
\label{tab::largescale_mse} 
\scriptsize
\resizebox{\textwidth}{!}{
\begin{tabular}{cccccc}
\toprule 
&{\text{Ours (w.o.~Rotation)}} &{\text{Ours (w.  Rotation)}} &{\text{Random Forest}} &{\text{LiquidSVM}} &{\text{GBRT}} \\
\midrule
{\tt EGS} &  $1.40\text{e-}4(8.04\text{e-}6) $ & $1.01\text{e-}4(4.40\text{e-}6) $ & $1.42\text{e-}4(4.85\text{e-}6) $ & $\mathbf{7.38\text{e-}5(3.73\text{e-}6)} $ & $9.12\text{e-}5(3.23\text{e-}6)$ \\
{\tt AEP} &  $\mathbf{5078.88(319.57)} $ & $ 6596.43(364.30)  $ & $5184.99(312.02) $ & $6728.02(398.056) $ & $6006.91(335.37)$ \\
{\tt CAD} &  $2582.74(77.07) $ & $2980.76(101.08) $ & $2434.30(73.60) $ & $ 2996.90(91.70)$ & $ \mathbf{2308.72(65.28)}$ \\
{\tt SCD} &  $\mathbf{84.21(3.21)} $ & $89.89(3.22)$ & $ 90.67(3.50)  $ & $110.02(5.89) $ & $ 103.93(3.17) $ \\
{\tt HPP} &  $\mathbf{1100.14(68.62)} $ & $ 1218.92(70.13)  $ & $1125.47(73.79) $ & $1262.78(81.57) $ & $1211.36(69.17)$ \\
{\tt ONP} &  $\mathbf{123.68(47.63)} $ & $125.04(47.80)$ & $ 125.75(47.26)$ & $125.65(47.64) $ & $ 127.60(46.60)$ \\
{\tt PTS} &  $\mathbf{11.38(0.18)} $ & $12.40(0.19) $ & $12.60(0.17) $ & $13.73(0.23) $ & $ 16.69(0.22)$ \\
\hline
{\tt MSD} & $80.45(0.04) $ & $ \mathbf{77.42(0.07)}$ & $85.72(0.04)$ & $85.33(0.73)$ & $81.84(0.01)$ \\ 
{\tt BUZ} &  $1.38\text{e}4(40.15)$  & $\mathbf{1.28e4(65.12)}$ & $1.44\text{e}4(41.49)$ & $2.99\text{e}4(124.23)$ & $1.43\text{e}4(56.64)$ \\
{\tt GNG} & $389.00(0.08)$ & $\mathbf{261.16(0.03)}$ & $267.70(0.04)$ &  $272.98(5.28)$ & $261.56(0.00)$ \\
{\tt GTM} & $\mathbf{3.44(0.00)}$ & $12.73(0.08)$ & $4.79(0.02)$ & $13.45(0.16)$ & $4.80(0.00)$ \\ 
{\tt DGM} & $8.08e3(38.57)$ & $\mathbf{6.23e3(45.00)}$ & $1.39e4(68.16)$ & $1.00e4(102.17)$ & $8.19e3(0.00)$ \\
\hline
\bottomrule
\end{tabular}
}
\begin{minipage}{\textwidth}
\begin{tablenotes}
\item{*} The best results are marked in \textbf{bold}, and the standard deviation is reported in the parenthesis beside each value.
\end{tablenotes}
\end{minipage}
\end{table}

\begin{table}[ht] 
\vskip 0.0in
\setlength{\tabcolsep}{8pt}
\centering
\captionsetup{justification=centering}
\caption{\footnotesize{Average \textit{MAE} over Moderate and Large-scale datasets}}
\label{tab::largescale_mae} 
\scriptsize
\begin{tabular}{cccccc}
\toprule 
&{\text{Ours (w.o.~Rotation)}}  &{\text{Ours (w. Rotation)}} &{\text{Random Forest}} &{\text{LiquidSVM}} &{\text{GBRT}} \\
\midrule
{\tt EGS} &  $8.92\text{e-}3(2.88\text{e-}4) $ & $7.35\text{e-}3(1.46\text{e-}4) $ & $9.17\text{e-}3(1.43\text{e-}4) $ & $\mathbf{5.98\text{e-}3(1.67\text{e-}4)} $ & $7.09\text{e-}3(1.18\text{e-}4) $ \\
{\tt AEP} &  $\mathbf{34.06(0.94)} $ & $40.92(0.83)$ & $34.40(0.84) $ & $42.69(0.85) $ & $ 41.04(0.84)$ \\
{\tt CAD} &  $34.63(0.40) $ & $ 36.82(0.52)   $ & $\mathbf{32.10(0.41)} $ & $37.33(0.47) $ & $32.43(0.42) $ \\
{\tt SCD} &  $\mathbf{4.97(0.07)} $ & $ 5.24(0.08)  $ & $ 5.32(0.08)$ & $6.09(0.18) $ & $6.30(0.09) $ \\
{\tt HPP} &  $17.23(0.27) $ & $18.65(0.31)  $ & $\mathbf{17.10(0.28)} $ & $18.90(0.36) $ & $18.31(0.27) $ \\
{\tt ONP} &  $\mathbf{3.01(0.08)} $ & $ 3.07(0.07) $ & $3.06(0.07)$ & $3.18(0.06)$ & $3.05(0.11) $ \\
{\tt PTS} &  $\mathbf{2.27(0.03)} $ & $2.37(0.02) $ & $2.41(0.02) $ & $2.59(0.02) $ & $3.05(0.02) $ \\
\hline
{\tt MSD} & $6.29(0.00) $ & $\mathbf{6.08(0.00)} $ & $6.52(0.01)$ & $6.47(0.02)$ & $6.37(0.00)$ \\ 
{\tt BUZ} & $38.70(0.06)$ & $\mathbf{38.68(0.05)}$ & $39.70(0.03)$ & $47.33(0.49)$ & $40.37(0.02)$ \\
{\tt GNG} & $13.90(0.03)$ & $\mathbf{12.54(0.00)}$ & $12.69(0.00)$ &  $12.66(0.02)$ & $12.55(0.00)$ \\
{\tt GTM} & $\mathbf{1.09(0.00)}$ & $2.16(0.01)$ & $1.28(0.00)$ & $2.24(0.02)$ & $1.32(0.00)$ \\ 
{\tt DGM} & $55.89(0.47)$ & $\mathbf{44.24(0.41)}$ & $69.63(0.23)$ & $47.93(1.84)$ & $60.15(0.00)$ \\
\hline
\bottomrule
\end{tabular}
\begin{minipage}{\textwidth}
\begin{tablenotes}
\item{*} The best results are marked in \textbf{bold}, and the standard deviation is reported in the parenthesis under each value.
\end{tablenotes}
\end{minipage}
\end{table}

\begin{table}[ht] 
\vskip 0.0in
\setlength{\tabcolsep}{8pt}
\centering
\captionsetup{justification=centering}
\caption{\footnotesize{Performance Comparisons w.r.t. Accuracy and Running Time on {\tt MSD} Dataset}}
\label{tab::largescale_runningtime_msd} 
\scriptsize
\begin{tabular}{cccc}
\toprule 
{\text{Algorithms}} & {\textit{MSE}} & {\textit{MAE}} & {\text{Running Time}} \\
\midrule
{\text{Ours (w.o.~Rotation)}}, $T=100$, $K=100$ & $81.34(0.06)$ &$6.341(0.00)$ & $377.33(6.41)$ \\
{\text{Ours (w.o.~Rotation)}}, $T=200$, $K=100$ & $80.66(0.08)$ &$6.301(0.00)$  & $737.06(8.74)$ \\
{\text{Ours (w. Rotation)}}, $T=100$, $K=100$ & $78.07(0.06)$ &$6.121(0.00)$  & $539.08(4.89)$ \\
{\text{Ours (w. Rotation)}}, $T=200$, $K=100$ & $77.42(0.07)$ &$6.083(0.00)$  &  $1070.29(7.02)$ \\	
\hline
{\text{Random Forest}}, $T=100$ & $85.98(0.06)$ &$6.533(0.00)$  &  $223.79(4.37)$ \\
{\text{Random Forest}}, $T=200$ & $85.82(0.05)$ &$6.526(0.00)$  &  $417.46(4.32)$ \\
{\text{Random Forest}}, $T=500$ & $85.72(0.04)$ &$6.525(0.01)$  &  $1009.95(21.03)$ \\
{\text{LiquidSVM}} & $85.33(0.73)$ &$6.475(0.02)$  &  $380.99(6.99)$ \\
{\text{GBRT}}, $T=100$ & $84.21(0.00)$ &$6.490(0.00)$  &  $1072.73(21.03)$ \\
{\text{GBRT}}, $T=200$ & $83.01(0.00)$ &$6.435(0.00)$  &  $2430.94(28.41)$ \\
{\text{GBRT}}, $T=500$ & $81.84(0.00)$ &$6.373(0.00)$  &  $6286.47(89.26)$ \\
\hline
\bottomrule
\end{tabular}
\end{table}


\begin{table}[ht] 
	\vskip 0.0in
	\setlength{\tabcolsep}{8pt}
	\centering
	\captionsetup{justification=centering}
	\caption{\footnotesize{Performance Comparisons w.r.t. Accuracy and Running Time on {\tt GTM} Dataset }}
	\label{tab::largescale_runningtime_gtm} 
	\scriptsize
	\begin{tabular}{cccc}
		\toprule 
		{\text{Algorithms}} & {\textit{MSE}} & {\textit{MAE}} & {\text{Running Time}} \\
		\midrule
		{\text{Ours (w.o. Rotation)}}, $T=100$, $K=100$ & $3.46(0.01)$ &$1.08(0.00)$  & $1375.34(3.39)$ \\
		{\text{Ours (w.o. Rotation)}}, $T=200$, $K=100$ & $3.44(0.00)$ &$1.08(0.00)$  &  $2575.51(63.14)$ \\	
		{\text{Ours (w. Rotation)}}, $T=100$, $K=100$ & $12.78(0.08)$ &$2.16(0.01)$  & $1557.91(45.96)$ \\
		{\text{Ours (w. Rotation)}}, $T=200$, $K=100$ & $12.73(0.08)$ &$2.17(0.01)$  &  $3054.23(74.22)$ \\	
		\hline
		{\text{Random Forest}}, $T=100$ & $4.82(0.03)$ &$1.28(0.00)$  &  $255.23(0.12)$ \\
		{\text{Random Forest}}, $T=200$ & $4.81(0.02)$ &$1.28(0.00)$  &  $452.59(8.89)$ \\
		{\text{Random Forest}}, $T=500$ & $4.79(0.02)$ &$1.28(0.00)$  &  $1106.80(10.45)$ \\
		{\text{LiquidSVM}} & $13.45(0.16)$ &$2.24(0.02)$  &  $3734.02(53.02)$ \\
		{\text{GBRT}}, $T=100$ & $5.46(0.00)$ &$1.45(0.00)$  &  $1850.05(68.79)$ \\
		{\text{GBRT}}, $T=200$ & $4.80(0.00)$ &$1.32(0.00)$  &  $3442.05(19.90)$ \\
		\hline
		\bottomrule
	\end{tabular}
\end{table}

\begin{table}[ht] 
	\vskip 0.0in
	\setlength{\tabcolsep}{8pt}
	\centering
	\captionsetup{justification=centering}
	\caption{\footnotesize{Performance Comparisons w.r.t. Accuracy and Running Time on {\tt DGM} Dataset}}
	\label{tab::largescale_runningtime_dgm} 
	\scriptsize
	\begin{tabular}{cccc}
		\toprule 
		{\text{Algorithms}} & {\textit{MSE}} & {\textit{MAE}} & {\text{Running Time}} \\
		\midrule
		{\text{Ours (w.o. Rotation)}}, $T=100$, $K=100$ & $8077.41(38.57)$ &$57.36(0.79)$  & $745.06(10.50)$ \\
		{\text{Ours (w.o. Rotation)}}, $T=200$, $K=100$ & $8116.47(109.24)$ &$55.89(0.47)$  &  $1795.54(5.95)$ \\	
		{\text{Ours (w. Rotation)}}, $T=100$, $K=100$ & $6234.73(45.00)$ &$44.24(0.41)$  & $1232.47(14.21)$ \\
		{\text{Ours (w. Rotation)}}, $T=200$, $K=100$ & $6242.31(42.92)$ &$44.45(0.40)$  &  $2486.76(24.15)$ \\	
		\hline
		{\text{Random Forest}}, $T=100$ & $13930.05(105.56)$ &$69.50(0.31)$  &  $202.14(3.84)$ \\
		{\text{Random Forest}}, $T=200$ & $13907.25(64.86)$ &$69.64(0.24)$  &  $365.03(0.09)$ \\
		{\text{Random Forest}}, $T=500$ & $13902.79(68.16)$ &$69.63(0.23)$  &  $905.94(0.23)$ \\
		{\text{LiquidSVM}} & $10040.21(102.17)$ &$47.93(1.84)$  &  $2201.74(5.89)$ \\
		{\text{GBRT}}, $T=100$ & $8214.59(0.00)$ &$60.18(0.00)$  &  $820.83(15.58)$ \\
		{\text{GBRT}}, $T=200$ & $8192.71(0.00)$ &$60.15(0.000)$  &  $1622.61(19.46)$ \\
		\hline
		\bottomrule
	\end{tabular}
\end{table}

In Tables \ref{tab::largescale_mse} and \ref{tab::largescale_mae}, we observe that our method with the optimal rotation strategy can reach the comparable or even the best performance in many moderate-sized and large-scale datasets. 
Here we take the rotation transform as an option, where we can select the best strategy through validation in practice.
At a significance level of $0.05$, our method is significantly different from other methods in most cases. 
This means that with the combination of boosting and ensemble, we come to a conclusion that our method shows promising performance compared to the efficient algorithms such as boosting-based algorithm GBRT, forest-based algorithm Random Forest, and kernel-based algorithm LiquidSVM.

Furthermore, Table \ref{tab::largescale_runningtime_msd}, \ref{tab::largescale_runningtime_gtm}, and \ref{tab::largescale_runningtime_dgm} record the \textit{MSE}, \textit{MAE}, and running time performance of each method under different parameter settings on three large-scale data sets {\tt MSD}, {\tt GTM} and {\tt DGM}. 
It can be seen in the tables that our GBBHE algorithm significantly outperforms other compared methods w.r.t. accuracy. 
Yet the running time of our method, with rotation or not, is comparable among other algorithms in large-scale circumstances. 
For example, on {\tt MSD}, the running time our GBBHE algorithm (w.o. Rotation) with $T=100$ and $K=100$ is comparable with Random Forest ($T=500$) and GBRT ($T=100$).
Nonetheless, the \textit{MSE} of our GBBHE (w.o. Rotation) is $1.37$ less than Random Forest ($T=500$) and is $2.00$ less than GBRT ($T=100$).
On the other hand, when compared with GBRT, our method turns out to converge faster. 
For example, on the {\tt GTM} dataset, the \textit{MSE} of our GBBHE (w.o. Rotation) drops by only $0.02$ when $T$ changes from $100$ to $200$, while the \textit{MAE} is nearly the same, which indicates convergence of our algorithm.
By contrast, the \textit{MSE} of GBRT decreases by a large margin from $5.46$ to $4.80$ when $T$ changes from $100$ to $200$. 
Besides, the \textit{MAE} decreases from $1.45$ to $1.32$.
Moreover, despite with $T=200$, where more running time is consumed, GBRT still shows worse performance than our GBBHE (w.o. Rotation).
This exactly demonstrates the faster convergence to a better performance of our algorithm.


\section{Comments and Discussions}\label{sec::comments}

\subsection{Comparisons with the Prior Work}

In the prior work of this paper \citep{cai2020boosted}, we take ordinary histogram transforms as the base learner in the gradient boosting algorithm and proposed the \textit{Boosted Histogram Transform for Regression} (BHTR). The convergence rate is proved to be $n^{-2\alpha/(4\alpha+d)}$ in the space $C^{0,\alpha}$ and $n^{-2(1+\alpha)/(2(1+\alpha)+d)}$ in the space $C^{1,\alpha}$. Moreover, a lower bound of convergence rates for the base learner is proved to be $n^{-2/(2+d)}$ which demonstrates the advantage of boosting. In this paper, we take a step further to analyze the behavior of an algorithm that applies better to the regression problem with large-scale and especially high-dimensional data. To be specific, we introduce additional randomness to GBBH by utilizing the rotated binary histogram as the base learner and manage to derive fast convergence in the H\"{o}lder function space. Unfortunately, the GBBH converges slightly slower than BHTR. However, it is worth pointing out that the major shortcoming of ordinary histogram partition is that the number of splits grows exponentially with dimension $d$. Thus, it is difficult for BHTR to apply to high-dimensional data. Therefore, in this paper, we adopt the binary histogram partition to deal with this problem, and further use the ensemble method to improve the computational efficiency for large-scale regression.

\subsection{Comments on Theoretical Results}

Previous theoretical works about boosting algorithms for regression include \citet{buhlmann2003boosting} and \citet{lin2019boosted}, where linear regressors and kernel ridge regressors are used as the base learners. These works analyze the learning performance by using the integral operator approach and prove the optimal convergence rate. However, this analysis turns out to be inapplicable to our method. In this paper, we conduct analysis under the framework of \textit{regularized empirical risk minimization} (RERM).

Throughout the ensemble learning algorithms, perhaps the most related work to ours is \citet{biau2012analysis}, where they investigate a random forest model with a midpoint splitting rule which coincides with the construction procedure of our binary histogram. The convergence rate of their proposed algorithm is proved to be $n^{-0.75/(0.75 + d\log 2)}$ in the space $C^{0,1}$.
As for this paper, the convergence rate of our GBBH when the target function lies in the space $C^{0,1}$ turns out to be $n^{-0.75/(1.5+d\log 2)}$, slower than that of \citet{biau2012analysis}'s. However, for smoother functions in the subspace $C^{1,0}$, the convergence rate of GBBH $n^{-1/(2+d\log 2)}$ is actually faster than that of \citet{biau2012analysis}'s when $d \geq 5$, which indicates that our GBBH can deal better with smoother target functions.

\subsection{Comments on Large-scale Regression}

In the literature, there have been many efforts on solving the large-scale regression problem. For example, the mainstream solutions fall into two categories, the \emph{horizontal methods} and the \emph{vertical methods}. The former partitions the data set into several disjoint subsets, implements a certain learning algorithm to each data subset to obtain a local predictor, and finally synthesizes a global output. However, this approach suffers from its own inherent disadvantages that the local predictor may be quite different from the global optimal predictor. On the other hand, vertical methods divide the feature space into multiple non-overlapping cells through different partition methods, e.g. \citet{suykens2002least, EsSuMo06a, biau2012analysis}. Then a predictor is embedded on each partitioning cell, such as Gaussian process regression \citep{park11a, park16a, park2018patchwork}, support vector machines \citep{meister16a, thomann2017spatial}, etc.

In this paper, our algorithm is inspired by the vertical methods. However, we notice that previous vertical methods for large-scale regression usually adopt kernel-based approaches to ensure sound theoretical properties and enhance the performance, especially under high-dimensional scenarios. By contrast, instead of resorting to kernel methods, we achieve comparable or even better performance by combining two ensemble learning methods. Moreover, our algorithm adopts the binary histogram partition, which enjoys high computational efficiency compared with kernel methods even on high-dimensional data. As is shown in Table \ref{tab::largescale_mse} and \ref{tab::largescale_mae}, GBBHE enjoys the lowest testing error measured by both \textit{MSE} and \textit{MAE}. Moreover, our GBBHE turns out to be more computationally efficient than these kernel-based methods. For example, in Table \ref{tab::largescale_runningtime_msd}, \ref{tab::largescale_runningtime_gtm}, and \ref{tab::largescale_runningtime_dgm}, our method runs faster than the kernel-based LiquidSVM.

Previous works on boosting algorithms for solving large-scale regression often adopt acceleration techniques mainly from the perspective of optimization. For example, in \citet{biau2019accelerated}, the computational efficiency is enhanced by incorporating an accelerated gradient descent technique. By contrast, our GBBHE for the first time reduces the computational cost by accelerating the convergence with respect to $T$, which is guaranteed by statistical learning theory. To be specific, we show that in Theorem \ref{thm::optimalForestlarge} to achieve the same convergence rate, we require $T_nK_n$ (instead of $T_n$) to be of the order $n^{1/(8+4d\log 2)}$. That is, we can enhance the computational efficiency of our proposed gradient boosting algorithm through theoretically proved reduction of the number of iterations $T$.

\section{Error Analysis}\label{sec::ErrorAnalysis}

This section provides more comprehensive error analysis for the theoretical results in Section \ref{sec::Thorecialresults}. In Subsection \ref{sec::fundmental}, we present some fundamental lemmas and propositions for the properties of the binary histogram transform and the sample error analysis. 
Then, in Subsections \ref{subsec::analysisc0alpha} and \ref{subsec::analysisc1alpha},
we conduct approximation error analysis for the boosted regressor $f_{\mathrm{D},B}$ under the assumption that the Bayes decision function $f^*_{L,\mathrm{P}}$ lies in the H\"{o}lder spaces $C^{0,\alpha}$ and $C^{1,0}$, respectively.

\subsection{Fundamental Lemmas and Propositions}\label{sec::fundmental}

\subsubsection{Properties of Binary Histogram Transform}\label{sec::adapht}

Throughout the proof of this paper, we will make repeated use of the following two facts proposed by \cite{biau2012analysis}.

\begin{fact} \label{fact1}
For $x \in H^{-1}(B_r)$, let $A_p(x)$ defined by \eqref{equ::InputBin} be the rectangular cell of the rotated binary histogram containing $x$ and $S_p^j(x)$ be the number of times that $A_p(x)$ is split on the $j$-th coordinate ($j=1,\ldots,d$) in the transformed space. Then conditionally on the rotation transformation $H$, $S_p^j(x)$ has binomial distribution with parameters $p$ and probability $1/d$ and satisfies
\begin{align*}
\sum_{j=1}^d S^j_p(x) = p.
\end{align*}
Moreover, let $A_p^j(x)$ be the size of the $j$-th dimension of $H(A_p(x))$ in the transformed space. Then we have
\begin{align}\label{equ::apjr}
A_p^j(x) | R \overset{\mathcal{D}}{=} 2 r \cdot 2^{- S_p^j(x)},
\end{align}
where $\cdot|R$ denotes the probability distribution conditionally on the rotation transformation $H(x) = R \cdot x$ and $\overset{\mathcal{D}}{=}$ indicates that variables in the two sides of the equation have the same distribution.
\end{fact}

\begin{fact} \label{fact2}
Let $\mu$ be the Lebesgue measure.  For $x \in H^{-1}(B_r)$, let $N_p(x)$ be the number of samples falling in the same cell as $x$, that is,
\begin{align*}
N_p(x) = \sum_{i=1}^n \eins_{\{ X_i \in A_p(x) \}}.
\end{align*}
By construction, we have
\begin{align}\label{equ::muapx}
\mu(A_p(x)) = (2r)^d \cdot 2^{-p}.
\end{align}
\end{fact}

Before we proceed, we present the following lemma, which helps to bound the diameter of the rectangular cell $A_p(x)$.

\begin{lemma}\label{equ::basicinequality}
Suppose that $x_i > 0$, $1 \leq i \leq d$ and $0 < \alpha \leq 1$. Then we have
\begin{align}\label{equ::sumnxialpha}
\biggl(\sum^d_{i=1}x_i\biggr)^{\alpha}\leq \sum^d_{i=1} x_i^\alpha.
\end{align}
\end{lemma}

Combining Lemma \ref{equ::basicinequality} with Fact \ref{fact2}, it is easy to derive the following lemma which plays an important role to bound the approximation error of the estimator.

\begin{lemma}\label{lem::diamapx}
Let the diameter of the set $A\subset \mathbb{R}^d$ be defined by
\begin{align*}
\mathrm{diam}(A) := \sup_{x, x' \in A} \|x - x'\|_2.
\end{align*}
Then for any $x \in \mathcal{X}$ and $0 < \beta \leq 2$, there  holds
\begin{align*}
\mathbb{E}_{\mathrm{P}_{R,Z}} \bigl( \mathrm{diam}(A_p(x))^{\beta} \bigr)
\leq (2r)^{\beta} d \exp \biggl( \frac{(2^{-\beta}-1)p}{d} \biggr).
\end{align*}
\end{lemma}

For any $x \in B_r$, let $\underline{a}_p^j(x)$ and $\overline{a}_p^j(x)$ be the minimum and maximum values of the $j$-th entries of points in  $H(A_p(x))$. Then, by the construction of rotated binary histogram, there holds
\begin{align*}
H(A_p(x)) = [\underline{a}_p^1(x), \overline{a}_p^1(x)] \times \cdots \times [\underline{a}_p^d(x), \overline{a}_p^d(x)].
\end{align*}

The next theorem gives an explicit form of the distance between $x_i$ and the center of the interval $[\underline{a}_p^j(x), \overline{a}_p^j(x)]$, which is used to derive the lower bound for the error of single binary histogram regressor.

\begin{lemma}\label{lem::basic2}
Let the rotated binary histogram $A_p$ be defined as in Algorithm \ref{alg::AdaptivePartitioning} with identity map $H(x) := x$. Moreover, let $A_p(x)$ defined by \eqref{equ::InputBin} be the rectangular cell containing $x$ and $S^j_p(x)$ be the number of times that $A_p(x)$ is split on the $j$-th coordinate ($j=1,\ldots,d$) in the transformed space. For any $x \in B_r$, let $x_j$ be the $j$-th entry of $x$. If $S_p^j(x) = k$, $0 \leq k \leq q$, then we have
\begin{align*}
\biggl| x_j - \frac{\underline{a}_p^j(x) + \overline{a}_p^j(x)}{2} \biggr|
= \min_{q \in Q_k} |x_j - q|,
\end{align*}
where
\begin{align*}
Q_k := \biggl\{ \frac{r(2i-1)}{2^k} \, \bigg| \, - 2^{k-1}+1 \leq i \leq 2^{k-1} \biggr\}.
\end{align*}
\end{lemma}

\subsubsection{Bounding the Sample Error Term} \label{sec::boundsamplerror}

To derive bounds on the sample error of regularized empirical risk minimizers, let us briefly recall the definition of VC dimension measuring the complexity of the underlying function class.

\begin{definition}[VC dimension] \label{def::VCdimension}
Let $\mathcal{B}$ be a class of subsets of $\mathcal{X}$ and $A \subset \mathcal{X}$ be a finite set. The trace of $\mathcal{B}$ on $A$ is defined by $\{ B \cap A : B \subset \mathcal{B}\}$. Its cardinality is denoted by $\Delta^{\mathcal{B}}(A)$. We say that $\mathcal{B}$ shatters $A$ if $\Delta^{\mathcal{B}}(A) = 2^{\#(A)}$, that is, if for every $\tilde{A} \subset A$, there exists a $B \subset \mathcal{B}$ such that $\tilde{A} = B \cap A$. For $k \in \mathrm{N}$, let
\begin{align}\label{equ::VC dimension}
m^{\mathcal{B}}(k) := \sup_{A \subset \mathcal{X}, \, \#(A) = k} \Delta^{\mathcal{B}}(A).
\end{align}
Then, the set $\mathcal{B}$ is a Vapnik-Chervonenkis class if there exists $k<\infty$ such that $m^{\mathcal{B}}(k) < 2^k$ and the minimal of such $k$ is called the VC dimension of $\mathcal{B}$, and abbreviate as $\mathrm{VC}(\mathcal{B})$.
\end{definition}

To prove Lemma \ref{VCIndex}, we need the following fundamental lemma concerning with the VC dimension of the tree-based partitions of $\mathbb{R}^d$ with $s$ internal nodes, that is, we use $s$ hyper-planes without intersections between each other to split $\mathbb{R}^d$ into $s+1$ sub-regions. In fact, the histogram transform partition $A_p$ proposed in Section \ref{sub::histogram} is a binary tree-based partition with $s=2^{p}-1$ internal nodes. The key of the lemma \ref{VCIndex} follows the idea put forward by \citet{breiman2000some} of the construction of purely random forest. To this end, let $s \in \mathbb{N}$ be fixed and $\tilde{\pi}_s$ be a tree-based partition of $\mathbb{R}^d$ with $s$ internal nodes.

\begin{lemma}\label{VCIndex}
Let $\mathcal{B}_s$ be defined by
\begin{align} \label{Bp}
\mathcal{B}_s := \biggl\{ B : B = \bigcup_{j \in J} A_j, J \subset \{ 0, 1, \ldots, s \}, A_j \in \tilde{\pi}_{s} \biggr\}.
\end{align}
Then the VC dimension of $\mathcal{B}_s$ can be upper bounded by $ds + 2$. 
\end{lemma}

To investigate the capacity property of continuous-valued functions, we need to introduce the concept 
\textit{VC-subgraph class}. To this end, the \emph{subgraph} of a function $f : \mathcal{X} \to \mathbb{R}$ is defined by 
\begin{align*}
\textit{sg}(f) := \{ (x, t) : t < f(x) \}.
\end{align*}
A class $\mathcal{F}$ of functions on $\mathcal{X}$ is said to be a VC-subgraph class, if the collection of all subgraphs of functions in $\mathcal{F}$, which is denoted by $\textit{sg}(\mathcal{F}) := \{ \textit{sg}(f) : f \in \mathcal{F} \}$ is a VC class of sets in $\mathcal{X} \times \mathbb{R}$. Then the VC dimension of $\mathcal{F}$ is defined by the VC dimension of the collection of the subgraphs, that is, 
$\mathrm{VC}(\mathcal{F}) = \mathrm{VC}(\textit{sg}(\mathcal{F}))$.

Before we proceed, we also need to recall the definitions of the convex hull and VC-hull class. 
The symmetric \textit{convex hull} $\mathrm{Co}(\mathcal{F})$ of a class of functions $\mathcal{F}$ is defined as the set of functions $\sum_{i=1}^m \alpha_i f_i$ with $\sum_{i=1}^m |\alpha_i| \leq 1$ and each $f_i$ contained in $\mathcal{F}$. 
A set of measurable functions is called a \textit{VC-hull class}, if it is in the pointwise sequential closure of the symmetric convex hull of a VC-class of functions.

We denote the function set $\mathcal{F}$ as
\begin{align}\label{equ::functionFH}
\mathcal{F} := \bigcup_{p\in \mathbb{N}_+,R\sim \mathrm{P}_R} \mathcal{F}_p,
\end{align}
which contains all the functions of $\mathcal{F}_p$ induced by all possible rotation transformations $R$ with the size parameter $p$.
The following lemma presents the upper bound for the VC dimension of the function set $\mathcal{F}$.

\begin{lemma}\label{lem::VCFn}
Let $\mathcal{F}$ be the function set defined as in \eqref{equ::functionFH}. Then $\mathcal{F}$ is a $\mathrm{VC}$-subgraph class with 
\begin{align*}
\mathrm{VC}(\mathcal{F}) 
\leq d\cdot 2^{p+1}.
\end{align*}
\end{lemma}

To further bound the capacity of the function sets, we need to introduce the following fundamental descriptions which enables an approximation of an infinite set by finite subsets.

\begin{definition}[Covering Numbers]\label{def::Covering Numbers}
Let $(\mathcal{X}, d)$ be a metric space, $A \subset \mathcal{X}$ and $\varepsilon > 0$. We call $A' \subset A$ an $\varepsilon$-net of $A$ if for all $x \in A$ there exists an $x' \in A'$ such that $d(x, x') \leq \varepsilon$. Moreover, the $\varepsilon$-covering number of $A$ is defined as
\begin{align*}
\mathcal{N}(A, d, \varepsilon)
& = \inf \biggl\{ n \geq 1 : \exists x_1, \ldots, x_n \in \mathcal{X}
\text{ such that } A \subset \bigcup_{i=1}^n B_d(x_i, \varepsilon) \biggr\},
\end{align*}
where $B_d(x, \varepsilon)$ denotes the closed ball in $\mathcal{X}$ centered at $x$ with radius $\varepsilon$.
\end{definition}

The following lemma follows directly from Theorem 2.6.9 in \citet{van1996weak}. For the sake of completeness, we present the proof in Section \ref{sec::proofrelatsample}.

\begin{lemma}\label{thm::vart}
Let $\mathrm{Q}$ be a probability measure on $\mathcal{X}$ and 
\begin{align*}
\mathcal{F} := \bigl\{ f : \mathcal{X} \to \mathbb{R} : f \in [-M, M] \bigr\}.
\end{align*}
Assume that for some fixed $\varepsilon > 0$ and $v > 0$, the covering number of $\mathcal{F}$ satisfies
\begin{align} \label{CoverAssumption}
\mathcal{N}(\mathcal{F}, L_2(\mathrm{Q}), M \varepsilon)
\leq c \, (1/\varepsilon)^v.
\end{align}
Then there exists a universal constant $c'$ such that
\begin{align*}
\log \mathcal{N}(\mathrm{Co}(\mathcal{F}), L_2(\mathrm{Q}), M \varepsilon)
\leq c' c^{2/(v+2)} \varepsilon^{-2v/(v+2)}.
\end{align*}
\end{lemma}

The next theorem shows that covering numbers of $\mathcal{F}$ grow at a polynomial rate.

\begin{proposition}\label{the::Fncovering}
Let $\mathcal{F}$ be a function set defined as in \eqref{equ::functionFH}. Then there exists a universal constant $c_0 < \infty$ such that for any $\varepsilon \in (0, 1)$ and any probability measure $\mathrm{Q}$,
we have
\begin{align*}
\mathcal{N}(\mathcal{F},L_2(\mathrm{Q}),M\varepsilon)
\leq c_0 d \cdot 2^{p+1}  (16e)^{d\cdot 2^{p+1}}        
\varepsilon^{2-d\cdot 2^{p+2}}.
\end{align*}
\end{proposition}

The following theorem gives an upper bound on the covering number of the $\mathrm{VC}$-hull class  $\mathrm{Co}(\mathcal{F})$.

\begin{proposition}\label{the::convexFn}
Let $\mathcal{F}$ be the function set defined as in \eqref{equ::functionFH}. Then there exists a constant $c_1$ such that for any $\varepsilon \in (0, 1)$ and any probability measure $\mathrm{Q}$, there holds
\begin{align}\label{equ::convexFn}
\log \mathcal{N}(\mathrm{Co}(\mathcal{F}),L_2(\mathrm{Q}),M\varepsilon)
\leq c_1 2^{p/2}\cdot \varepsilon^{1/(d\cdot 2^p)-2}.
\end{align}
\end{proposition}

Next, let us recall the definition of entropy numbers.

\begin{definition}[Entropy Numbers] \label{def::entropy numbers}
Let $(\mathcal{X}, d)$ be a metric space, $A \subset \mathcal{X}$ and $m \geq 1$ be an integer. The $m$-th entropy number of $(A, d)$ is defined as
\begin{align*}
e_m(A, d) 
= \inf \biggl\{ \varepsilon > 0 : \exists x_1, \ldots, x_{2^{m-1}} \in \mathcal{X} 
\text{ such that } A \subset \bigcup_{i=1}^{2^{m-1}} B_d(x_i, \varepsilon) \biggr\}.
\end{align*}
Moreover, if $(A, d)$ is a subspace of a normed space $(E, \|\cdot\|)$ and the metric $d$ is given by $d(x, x') = \|x - x'\|$, $x, x' \in A$, we write $e_m(A, \|\cdot\|) := e_m(A, E) := e_m(A, d)$.
Finally, if $S : E \to F$ is a bounded, linear operator between the normed space $E$ and $F$, we denote
$e_m(S) := e_m(S B_E, \|\cdot\|_F)$.
\end{definition}

It is well-known that entropy numbers are closely related to the covering numbers. To be specific, entropy and covering numbers are in some sense inverse to each other. 
More precisely, for all constants $a > 0$ and $q > 0$, the implication
\begin{align}
e_i (T, d) \leq a i^{-1/q}, \quad \forall \, i \geq 1
\quad 
\Longrightarrow 
\quad 
\ln \mathcal{N}(T, d, \varepsilon) \leq \ln(4) (a/\varepsilon)^q, \quad \forall \, \varepsilon > 0
\label{EntropyCover}
\end{align}
holds by Lemma 6.21 in \citet{StCh08}. 
Additionally, Exercise 6.8 in \citet{StCh08} yields the opposite implication, namely
\begin{align}
\ln \mathcal{N}(T, d, \varepsilon) < (a/\varepsilon)^q, \quad \forall \, \varepsilon > 0 
\quad 
\Longrightarrow 
\quad e_i(T, d) \leq 3^{1/q} a i^{-1/q}, \quad \forall \, i \geq 1.
\label{CoverEntropy}
\end{align}

For a finite set $D \in \mathcal{X}^n$, we define the norm of an empirical $L_2$-space by
\begin{align*}
\|f\|^2_{L_2(\mathrm{D})}
= \mathbb{E}_{\mathrm{D}} |f|^2
:= \frac{1}{n} \sum_{i=1}^n |f(x_i)^2|.
\end{align*}
If $E$ is the function space \eqref{equ::En} and $\mathrm{D} \in \mathcal{X}^n$, then the entropy number $e_m(\mathrm{id} : E \to L_2(\mathrm{D}_X))$ equals the $m$-th entropy number of the symmetric convex hull of the family $\{ (f_i), f_i \in \mathcal{F}_i \}$, where $\mathrm{id} : E \to L_2(\mathrm{D}_X)$ denotes the identity map that assigns to every $f \in E$ the corresponding equivalence class in $L_2(\mathrm{D}_X)$.

\subsection{Error Analysis for $f^*_{L,\mathrm{P}} \in C^{0,\alpha}$} \label{subsec::analysisc0alpha}

First of all, we introduce some definitions and notations. For a given rotated binary histogram $H$ and split coordinates $Z$, we write
\begin{align} \label{def::fPH}
f_{\mathrm{P}}^p := \argmin_{f \in \mathcal{F}^p} \mathcal{R}_{L,\mathrm{P}}(f).
\end{align}
In other words, $f_{\mathrm{P}}^p$ is the function that minimizes the excess risk $\mathcal{R}_{L,\mathrm{P}}(f)$ over the function set $\mathcal{F}_p$. Then, elementary calculation yields
\begin{align*}
f_{\mathrm{P}}^p
= \mathbb{E}_{\mathrm{P}}(f_{L,\mathrm{P}}^*(X) | A_p(x))
= \sum_{j \in \mathcal{I}_p} \frac{\int_{A_j} f_{L,\mathrm{P}}^* \, d\mathrm{P}_X}{\mathrm{P}_X(x\in A_j)} \cdot \eins_{A_j}
= \sum_{j \in\mathcal{I}_p} \frac{\int_{A_j} \mathbb{E}(Y | X) \, d\mathrm{P}_X}{\mathrm{P}_X(x\in A_j)} \cdot \eins_{A_j}.
\end{align*}
Moreover, we write
\begin{align} \label{def::fDH}
f_{\mathrm{D}}^p:= \argmin_{f \in \mathcal{F}^p} \mathcal{R}_{L,\mathrm{D}}(f) 
\end{align}
for the empirical version, which can be further presented as
\begin{align*}
f_{\mathrm{D}}^p = \sum_{j \in \mathcal{I}_p} \frac{\sum_{i=1}^n Y_i \eins_{A_j}(X_i)}{\sum_{i=1}^n \eins_{A_j}(X_i)} \cdot \eins_{A_j}.
\end{align*}

The following proposition shows that the $L_2$ distance between $f_{\mathrm{P}}^p$ and $f^*_{L,\mathrm{P}}$ behaves polynomial in the regularization parameter $\lambda$ if we choose the size of binary histogram partition $p$ appropriately.

\begin{proposition}\label{pro::c0alphasingle}
Let the rotated binary histogram $A_p$ be defined in Algorithm \ref{alg::AdaptivePartitioning} with depth $p$. Moreover, suppose that the Bayes decision function $f^*_{L,\mathrm{P}}\in C^{0,\alpha}$. Then, for any fixed $\lambda > 0$, there holds
\begin{align*}
\mathbb{E}_{\mathrm{P}_{R,Z}} \bigl(
\lambda \cdot 4^p + \mathcal{R}_{L,\mathrm{P}}(f_{\mathrm{P}}^p) - \mathcal{R}^*_{L,\mathrm{P}} \bigr)
\leq c \cdot \lambda^{\frac{1-4^{-\alpha}}{2d\log 2+(1-4^{-\alpha})}},
\end{align*}
where $c$ is some constant depending on $\alpha$, $d$ and $r$.
\end{proposition}

\subsubsection{Upper Bound of Convergence Rate of GBBH}\label{sec::upperc0}

\begin{theorem}\label{thm::OracalBoost}
Let the rotated binary histogram $A_p$ be defined as in Algorithm \ref{alg::AdaptivePartitioning} with depth $p$. Furthermore, let $f_{\mathrm{D},B}$ be the GBBH regressor defined by \eqref{equ::fdlambda} and $A(\lambda)$ be the corresponding approximation error defined by \eqref{equ::approximationerror}. Then for all $\tau>0$, with probability $\mathrm{P}^n \otimes \mathrm{P}_{R,Z}$ not less than $1-3e^{-\tau}$, we have
\begin{align*}
\Omega_{\lambda}(f) + \mathcal{R}_{L,\mathrm{P}}(f_{\mathrm{D},B}) - \mathcal{R}_{L,\mathrm{P}}^*
\leq 12 A(\lambda) + 3456 M^2 \tau / n+ 3 c_0' T^{2\delta'} \lambda_1^{-2\delta'} \lambda_2^{-1} n^{-2},
\end{align*}
where $c_0'$ is a constant, $\delta=1/(d\cdot 2^p)$ and $\delta'=1-\delta$.
\end{theorem}

\subsubsection{Upper Bound of Convergence Rate of GBBHE}\label{sec::upperc0large}

\begin{theorem}\label{equ::OracalBoostlarge}
Let the rotated binary histogram $A_p$ be defined as in Algorithm \ref{alg::AdaptivePartitioning} with depth $p$. Furthermore, let $\bar{f}_{\mathrm{D},B}$ be the GBBHE regressor defined by \eqref{equ::barfdlambda} and $\bar{A}(\lambda)$ be the corresponding approximation error defined by \eqref{equ::barapproximationerror}. Then for all $\tau>0$, with probability $\mathrm{P}^n \otimes \mathrm{P}_{R,Z}$ not less than $1-3e^{-\tau}$, we have
\begin{align*}
\bar{\Omega}_{\lambda}(f) + 
\mathcal{R}_{L,\mathrm{P}}(\bar{f}_{\mathrm{D},B}) - \mathcal{R}_{L,\mathrm{P}}^*
\leq 12 \bar{A}(\lambda) + 3456 M^2 \tau / n+ 3 c_0' T^{2\delta'} \lambda_1^{-2\delta'} \lambda_2^{-1} n^{-2},
\end{align*}
where $c_0'$ is a constant, $\delta=1/(d\cdot 2^p)$ and $\delta'=1-\delta$.
\end{theorem}

\subsection{Error Analysis for $f^*_{L,\mathrm{P}}\in  C^{1,0}$} \label{subsec::analysisc1alpha}

A drawback to the analysis in $C^{0,\alpha}$ is that the usual Taylor expansion involved techniques for error estimation may not apply directly. As a result, we fail to prove the exact benefits of the boosting procedure. Therefore, in this subsection, we turn to the function space $C^{1,0}$ consisting of smoother functions. To be specific, we study the convergence rates of $f_{\mathrm{D},B}$ to the Bayes decision function $f_{L, \mathrm{P}}^* \in C^{1,0}$. To this end, there is a point in introducing some notations.

For fixed $p\in \mathbb{N}_{+}$, let $\{ A_{p,t}\}_{t=1}^T$ be rotated binary histograms with depth $p$ and split coordinates $Z_t$, $t = 1, \ldots, T$. 
Moreover, let $\{ f_{\mathrm{P}}^{p,t} \}_{t=1}^T$  and $\{ f_{\mathrm{D}}^{p,t} \}_{t=1}^T$ be defined as in \eqref{def::fPH} and \eqref{def::fDH}, respectively. 
For $x \in \mathcal{X}$, we define
\begin{align}\label{fpeensemble}
f_{\mathrm{P},\mathrm{E}}(x)
:= \frac{1}{T} \sum_{t=1}^T f_{\mathrm{P}}^{p,t}(x)
\end{align}
and 
\begin{align}\label{fempensemble}
f_{\mathrm{D},\mathrm{E}}(x)
:= \frac{1}{T} \sum_{t=1}^T f_{\mathrm{D}}^{p,t}(x).
\end{align}

Moreover, in the scenario of gradient boosted binary histogram ensemble for large-scale regression, we write
\begin{align}\label{equ::fptk}
f_{\mathrm{P}}^{p,t,k}:=\argmin_{f\in \mathcal{F}_t^k} \mathrm{R}_{L,\mathrm{P}}(f).
\end{align}
Furthermore, we define
\begin{align}\label{equ::barfppt}
\bar{f}_{\mathrm{P}}^{p,t}:=\frac{1}{K}\sum_{k=1}^K 	f_{\mathrm{P}}^{p,t,k},
\end{align}
and
\begin{align}\label{equ::barfpe}
\bar{f}_{\mathrm{P},\mathrm{E}}:=\frac{1}{T}\sum^T_{t=1} 	\bar{f}_{\mathrm{P}}^{p,t}.
\end{align}

In particular, for the binary histogram regressor, we are concerned with the lower bound for $f_{\mathrm{D}}^p$. In this case, we suppose that $\mathrm{P}_X$ is the uniform distribution on $B_r:=[-r,r]^d $. As a result, it is sufficient for us to consider the fixed transformation $H(x)=x$ to apply binary histogram partition on $B_r$.  Consequently, we make the error decomposition
\begin{align}     
\mathbb{E}_{\mathrm{P}^n\otimes \mathrm{P}_Z} \bigl( \mathcal{R}_{L, \mathrm{P}}(f_{\mathrm{D}}^p) - \mathcal{R}_{L, \mathrm{P}}^* \bigr)
& = \mathbb{E}_{\mathrm{P}^n\otimes\mathrm{P}_Z} \mathbb{E}_{\mathrm{P}_X} \bigl( f_{\mathrm{D}}^p(X) - f^*_{L, \mathrm{P}}(X) \bigr)^2
\nonumber\\
& = \mathbb{E}_{\mathrm{P}^n\otimes \mathrm{P}_Z} \mathbb{E}_{\mathrm{P}_X} \bigl( f_{\mathrm{D}}^p(X) - f_{\mathrm{P}}^p(X) \bigr)^2
\nonumber\\
&\phantom{=}
+ \mathbb{E}_{\mathrm{P}^n\otimes \mathrm{P}_Z} \mathbb{E}_{\mathrm{P}_X} \bigl( f_{\mathrm{P}}^p(X) - f_{L, \mathrm{P}}^*(X) \bigr)^2.
\label{equ::L2Decomposition}
\end{align}
It is important to note that both of the two terms on the right-hand side of  \eqref{equ::L2Decomposition} are data- and partition-independent due to the expectation with respect to $\mathrm{P}^n$ and $\mathrm{P}_Z$ respectively. Loosely speaking, the first error term corresponds to the expected estimation error of the estimator $f_{\mathrm{D}}^p$, while the second one demonstrates the expected approximation error.

\subsubsection{Upper Bound of Convergence Rate of GBBH} \label{sec::upperc1}

The next proposition presents the upper bound of the $L_2$ distance between the boosted regressor $f_{\mathrm{P},\mathrm{E}}$ and the Bayes decision function $f_{L,\mathrm{P}}^*$ in the H\"{o}lder space $C^{1,0}$.

\begin{proposition}\label{prop::biasterm}
Let $f_{\mathrm{P},\mathrm{E}}$ be defined by \eqref{fpeensemble}. Moreover, let the Bayes decision function $f_{L, \mathrm{P}}^* \in C^{1,0}$ and $\mathrm{P}_X$ be the uniform distribution on $B_{r,d}= [-r/\sqrt{d}, r/\sqrt{d}]^d$.
Then we have
\begin{align}\label{eq::barPE}
\mathbb{E}_{\mathrm{P}_{R,Z}}  \bigl( \mathcal{R}_{L,\mathrm{P}}( f_{\mathrm{P},\mathrm{E}}) - \mathcal{R}^*_{L,\mathrm{P}} \bigr) 
\leq \frac{c_L^2 (2r)^4 d}{T} \exp \biggl( - \frac{3p}{4d} \biggr) + 4 c_L^2 (2r)^{2\alpha+2} d^2 \exp \biggl( - \frac{p}{d} \biggr).
\end{align}
\end{proposition}

\subsubsection{Upper Bound of Convergence Rate of GBBHE}\label{sec::upperc1large}
The next proposition presents the upper bound of the $L_2$ distance between the boosted rotated binary histogram ensemble  $\bar{f}_{\mathrm{P},\mathrm{E}}$ and the Bayes decision function $f_{L,\mathrm{P}}^*$ in the H\"{o}lder space $C^{1,0}$. 
\begin{proposition}\label{prop::barbiasterm}
Let $\bar{f}_{\mathrm{P},\mathrm{E}}$ be defined by \eqref{equ::barfpe}. Moreover, let the Bayes decision function $f_{L, \mathrm{P}}^* \in C^{1,0}$ and $\mathrm{P}_X$ be the uniform distribution on $B_{r,d}= [-r/\sqrt{d},r/\sqrt{d}]^d$.
Then we have
\begin{align}\label{eq::PE}
\mathbb{E}_{\mathrm{P}_{R,Z}}  \bigl( \mathcal{R}_{L,\mathrm{P}}( \bar{f}_{\mathrm{P},\mathrm{E}}) - \mathcal{R}^*_{L,\mathrm{P}} \bigr) 
\leq \frac{c_L^2(2r)^4 d}{KT} \exp \biggl(-\frac{3p}{4d}\biggr)+4c_L^2(2r)^{2\alpha+2}d^2\exp\biggl(-\frac{p}{d}\biggr).
\end{align}
\end{proposition}

\subsubsection{Lower Bound of Convergence Rate of Binary Histogram Regression} \label{sec::lowerboundconve}

The following two propositions present the lower bound of approximation error and sample error of GBBH Regression, respectively.

\begin{proposition}\label{counterapprox}
Let the rotated binary histogram $A_p$ be defined as in Algorithm \ref{alg::AdaptivePartitioning} with the identity map $H(x) = x$ and the regression model be defined by 
\begin{align} \label{regressionmodel}
Y := f(X) + \varepsilon
\end{align}
with $f \in C^{1,0}$. Furthermore, suppose that $\mathrm{P}_X$ is the uniform distribution on $B_r=[-r,r]^d$ and $\varepsilon$ is independent of $X$ such that $\mathbb{E}(\varepsilon) = 0$. Moreover, suppose that there exists a fixed constant $\underline{c}_f \in (0, \infty)$ such that
\begin{align}\label{DegenerateSetF}
\|\nabla f\| \geq \underline{c}_f.
\end{align} 
Then we have
\begin{align*}
\mathbb{E}_{\mathrm{P}_Z} \bigl( 
\mathcal{R}_{L,\mathrm{P}}(f_{\mathrm{P}}^p) - \mathcal{R}_{L,\mathrm{P}}^* \bigr)
\geq \frac{3\underline{c}_f^2r^2d}{4}\biggl(1-\frac{3}{4d}\biggr)^p.
\end{align*}
\end{proposition}

\begin{proposition}\label{counterapprox2}
Let the rotated binary histogram $A_p$ be defined as in Algorithm \ref{alg::AdaptivePartitioning} with the identity map $H(x)=x$. Furthermore, let the regression model be defined as in \eqref{regressionmodel} such that $f \in C^{1,0}$ and $\|f(x)\|_{\infty}\geq \underline{c}_f>0$. Moreover, suppose that $\mathrm{P}_X$ is the uniform distribution on $B_r=[-r,r]^d$ and $\varepsilon$ is independent of $X$ such that $\mathbb{E}(\varepsilon) = 0$ and $\mathrm{Var}(\varepsilon) =: \sigma^2 \leq 4 M^2$. Then we have
\begin{align*}
\mathbb{E}_{\mathrm{P}^n\otimes \mathrm{P}_Z} \bigl( 
\mathcal{R}_{L,\mathrm{P}}(f_{\mathrm{D}}^p) - \mathcal{R}_{L,\mathrm{P}}(f_{\mathrm{P}}^p) \bigr)
\geq \frac{\sigma^2\cdot 2^p}{n}\biggl(1-2\exp\biggl(-\frac{n}{2^p}\biggr)\biggr)+(2r)^{2d} \underline{c}_f^2\biggl(1-\frac{1}{2^p}\biggr)^n.
\end{align*}
\end{proposition}

\section{Proofs}\label{sec::proofs}

\subsection{Proofs Related to Section \ref{sec::fundmental}}

\subsubsection{Proofs Related to Section \ref{sec::adapht}}

\begin{proof}[of Lemma \ref{equ::basicinequality}]
For any $1 \leq i \leq n$, it is easy to see that
\begin{align*}
0< \frac{x_i}{\sum^d_{i=1}x_i} < 1.
\end{align*}
Since $0 < \alpha \leq 1$, we have
\begin{align*}
\frac{\sum^d_{i=1}x_i^\alpha}{(\sum^d_{i=1}x_i)^{\alpha}}
= \sum^d_{i=1} \biggl(\frac{x_i}{\sum^d_{i=1}x_i}\biggr)^{\alpha} 
\geq \sum^d_{i=1} \frac{x_i}{\sum^d_{i=1}x_i}
= \frac{\sum^d_{i=1}x_i}{\sum^d_{i=1}x_i}=1.
\end{align*}
Consequently, we get
\begin{align*}
\biggl( \sum_{i=1}^d x_i \biggr)^{\alpha}
\leq \sum_{i=1}^d x_i^\alpha.
\end{align*}
Thus we complete the proof.
\end{proof}

\begin{proof}[of Lemma \ref{lem::diamapx}]
By definition, we have
\begin{align*}
\mathrm{diam}(A_p(x))
:= \biggl( \sum_{j=1}^d A_p^j(x)^2 \biggr)^{1/2}.
\end{align*} 
Consequently, \eqref{equ::apjr} in Fact \ref{fact1} implies
\begin{align*}
\mathrm{diam}(A_p(x))^{\beta}
= (2r)^{\beta} \biggl( \sum^d_{j=1} 2^{-2S_p^j(x)} \biggr)^{\beta/2}.
\end{align*}
Applying Lemma \ref{equ::basicinequality}, we get
\begin{align*}
\mathrm{diam}(A_p(x))^{\beta}\leq (2r)^{\beta}\sum^d_{j=1} 2^{-\beta S_p^j(x)}.
\end{align*}
Therefore, we obtain
\begin{align*}
\mathbb{E}_{ \mathrm{P}_Z} \bigl( \mathrm{diam}(A_p(x))^{\beta} \big| R \bigr)
& \leq \mathbb{E}_{ \mathrm{P}_Z} \biggl( (2r)^{\beta} \sum^d_{j=1} 2^{-\beta S_p^j(x)} \bigg| R \biggr)
   = (2r)^{\beta} \sum^d_{j=1} \mathbb{E}_{ \mathrm{P}_Z} \biggl( 2^{-\beta S_p^j(x)} \bigg| R \biggr)
\\
& = (2r)^{\beta} d \biggl( 1 - \frac{1-2^{-\beta}}{d} \biggr)^p
    \leq (2r)^{2\alpha} d \exp \biggl( \frac{(2^{-\beta}-1)p}{d} \biggr).
\end{align*}
Taking expectation with respect to $\mathrm{P}_R$, we prove the desired assertion.
\end{proof}

\begin{proof}[of Lemma \ref{lem::basic2}]
If $S_p^j(x)=k$, by the construction of rotated binary histogram partition, we have
\begin{align}\label{equ::12qk}
\frac{\underline{a}_p^j(x) + \overline{a}_p^j(x)}{2} \in Q_k.
\end{align}
By the definition of $Q_k$, for any $q^* \in Q_k$, there holds
\begin{align*}
\biggl| q^* - \frac{\underline{a}_p^j(x) + \overline{a}_p^j(x)}{2} \biggr|
\geq \frac{r}{2^{k-1}}.
\end{align*}
Since $x \in A_p(x)$, we have
\begin{align}\label{equ::xjapjl}
\biggl| x_j - \frac{\underline{a}_p^j(x) + \overline{a}_p^j(x)}{2} \biggr|
\leq \frac{r}{2^k}.
\end{align}
Therefore, using the triangular inequality, we obtain
\begin{align*}
|x_j - q^*| 
\geq \Biggl| \biggl| x_j - \frac{\underline{a}_p^j(x) + \overline{a}_p^j(x)}{2} \biggr|
                  - \biggl| q^* - \frac{\underline{a}_p^j(x) + \overline{a}_p^j(x)}{2} \biggr| \Biggr|
\geq \frac{r}{2^k}.
\end{align*}
This together with \eqref{equ::xjapjl} implies that
\begin{align*}
|x_j - q^*|
\geq \biggl| x_j - \frac{\underline{a}_p^j(x) + \overline{a}_p^j(x)}{2} \biggr| 
\end{align*}
holds for any $q^* \in Q_k$. Combining this with \eqref{equ::12qk}, we get
\begin{align*}				
\biggl| x_j - \frac{\underline{a}_p^j(x) + \overline{a}_p^j(x)}{2} \biggr|
= \min_{q \in Q_k} |x_j - q|,
\end{align*}
which leads to the desired assertion.
\end{proof}

\subsubsection{Proofs Related to Section \ref{sec::boundsamplerror}} \label{sec::proofrelatsample}

\begin{proof}[of Lemma \ref{VCIndex}]
This proof is conducted from the perspective of geometric constructions. 

\begin{figure}[htbp]
\centering
\begin{minipage}[b]{0.15\textwidth}
\centering
\includegraphics[width=\textwidth]{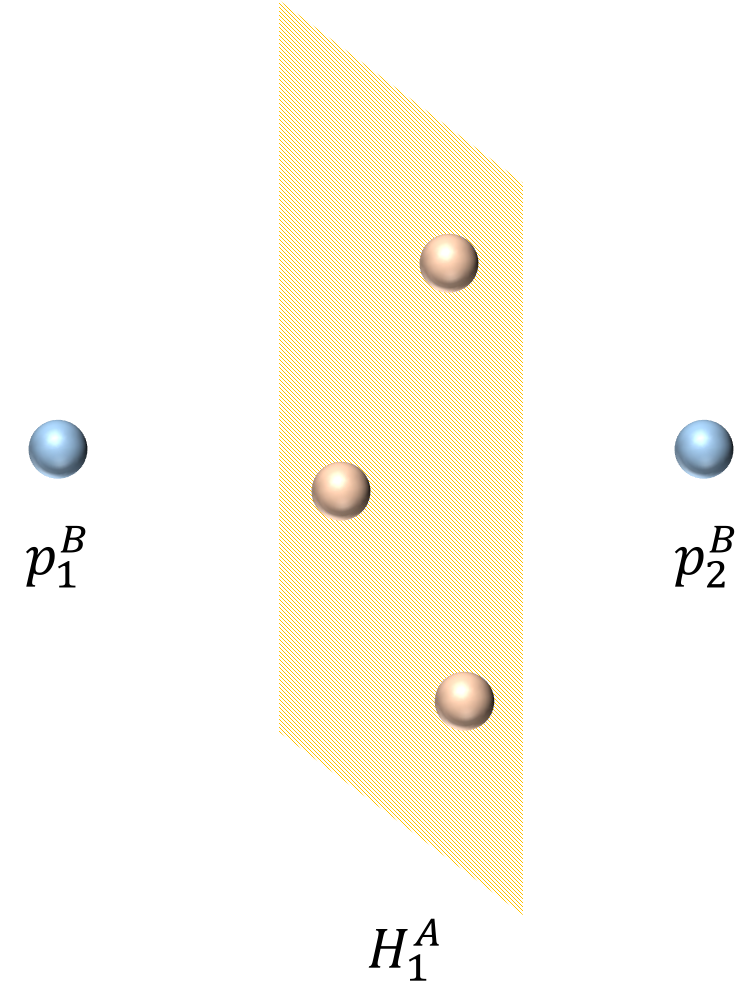}
$p=1$
\centering
\label{fig::p=1}
\end{minipage}
\qquad
\begin{minipage}[b]{0.21\textwidth}
\centering
\includegraphics[width=\textwidth]{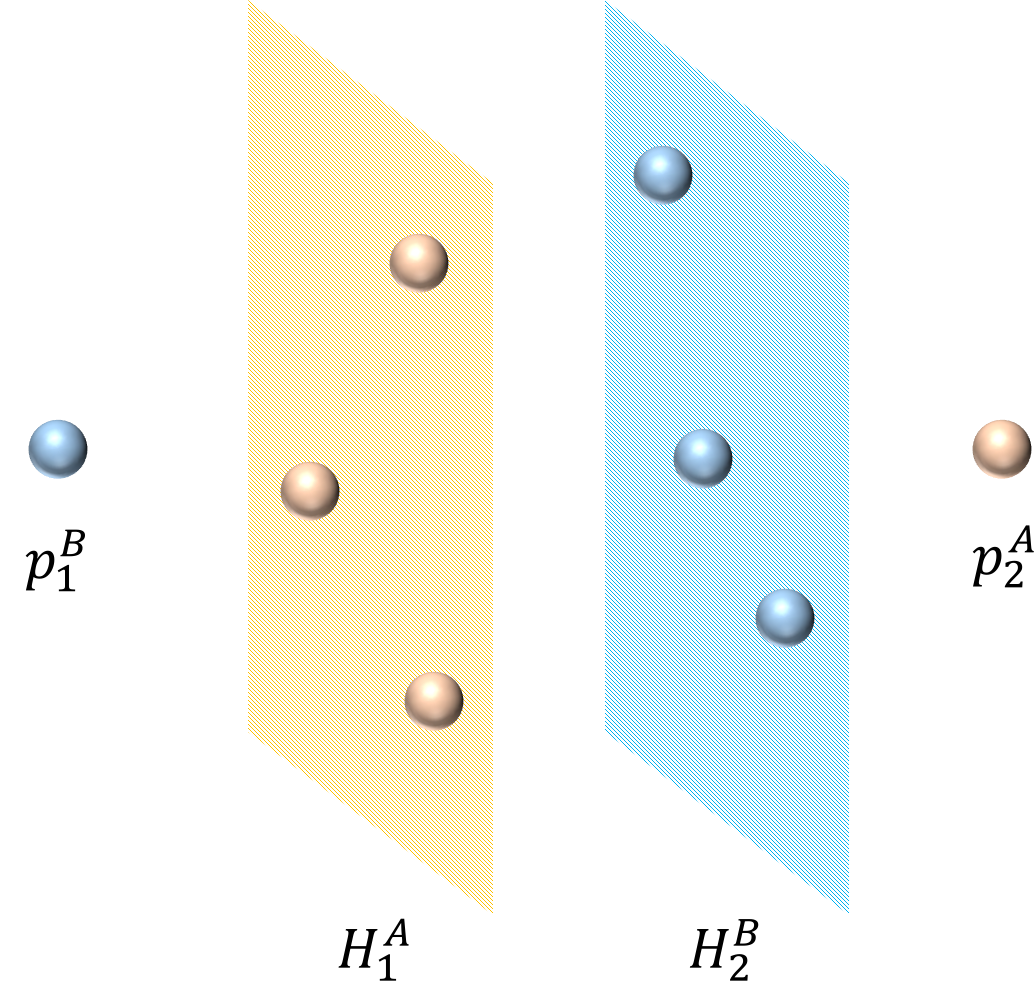}
$p=2$
\label{fig::p=2}
\end{minipage}
\qquad
\begin{minipage}[b]{0.32\textwidth}
\centering
\includegraphics[width=\textwidth]{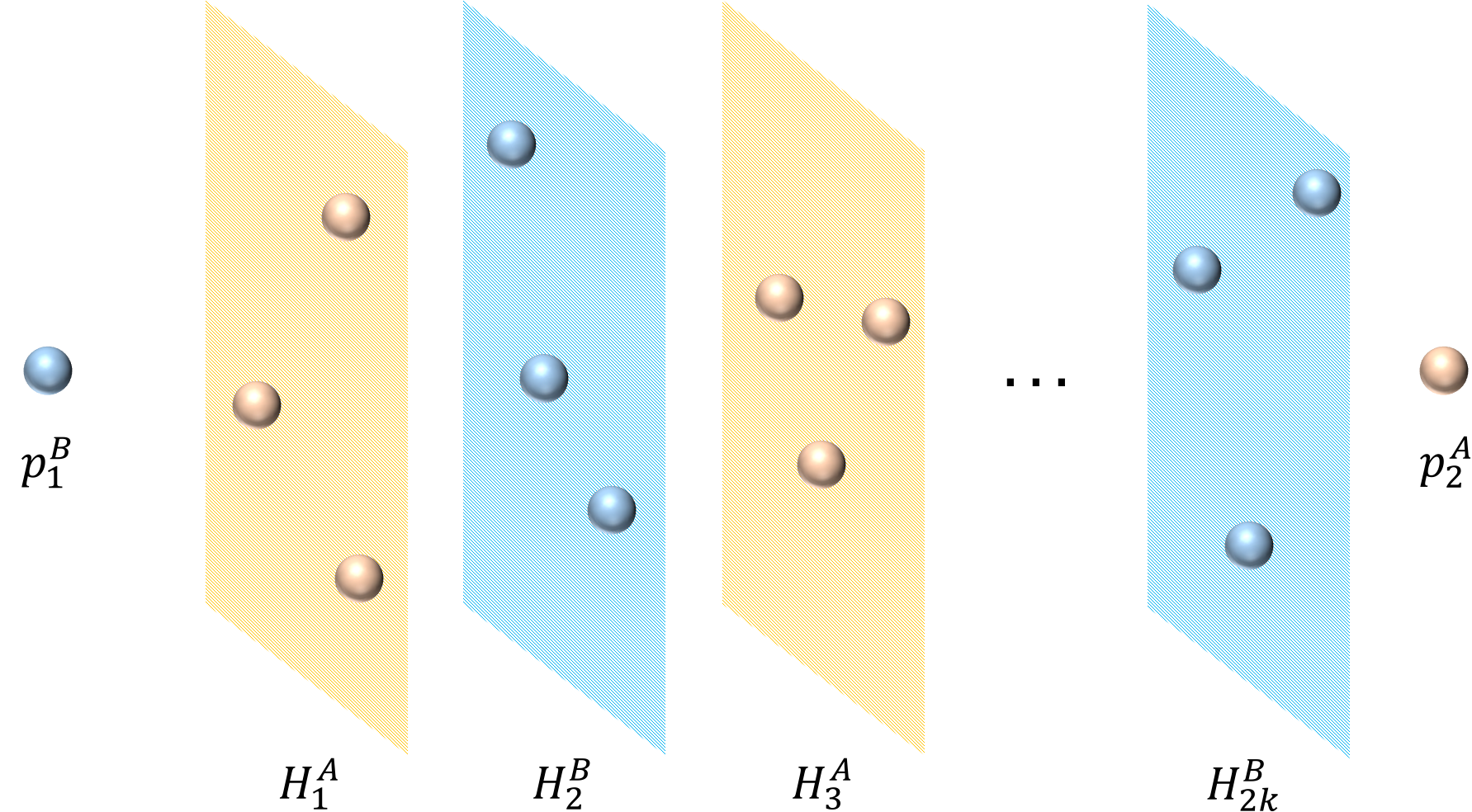}
$p=2k$
\label{fig::p=2k}
\end{minipage}
\caption{We take one case with $d=3$ as an example to illustrate the geometric interpretation of the VC dimension. The yellow balls represent samples from class $A$, blue ones are from class $B$ and slices denote the hyper-planes formed by samples. }
\label{fig::VC}
\end{figure}

We proceed by induction.
Firstly, we concentrate on partition with the number of splits $s=1$. Because of the dimension of the feature space is $d$,  the smallest number of sample points that cannot be divided by $s=1$ split is $d+2$. Concretely, owing to the fact that $d$ points can be used to form $d-1$ independent vectors and hence a hyperplane in a $d$-dimensional space, we might take the following case into consideration: There is a hyperplane consisting of $d$ points all from one class, say class $A$, and two points $s_1^B$, $s_2^B$ from the opposite class $B$ located on the opposite sides of this hyperplane, respectively. We denote this hyperplane by $H_1^A$. In this case, points from two classes cannot be separated by one split (since the positions are $s_1^B, H_1^A, s_2^B$), so that we have $\mathrm{VC}(\mathcal{B}_1) \leq d + 2$.

Next, when the partition is with the number of splits $s=2$, we analyze in the similar way only by extending the above case a little bit. Now, we pick either of the two single sample points located on opposite side of the $H_1^A$, and add $d-1$ more points from class $B$ to it. Then, they together can form a hyperplane $H_2^B$ parallel to $H_1^A$. After that, we place one more sample point from class $A$ to the side of this newly constructed hyperplane $H_2^B$. In this case, the location of these two single points and two hyperplanes are $s_1^B, H_1^A, H_2^B, s_2^A$. Apparently, $s=2$ splits cannot separate these $2d+2$ points. As a result, we have $\mathrm{VC}(\mathcal{B}_2) \leq 2d + 2$.

Inductively, the above analysis can be extended to the general case of number of splits $s \in \mathbb{N}$. In this manner, we need to add points continuously to form $s$ mutually parallel hyperplanes where any two adjacent hyperplanes should be constructed from different classes. Without loss of generality, we consider the case for $s=2k+1$, $k \in \mathbb{N}$, where two points (denoted as $s_1^B$, $s_2^B$) from class $B$ and $2k+1$ alternately appearing hyperplanes form the space locations: $s_1^B, H_1^A, H_2^B, H_3^A, H_4^B, \ldots, H_{(2k+1)}^A, s_2^B$. 
Accordingly, the smallest number of points that cannot be divided by $s$ splits is $ds+2$, leading to $\mathrm{VC}(\mathcal{B}_s) \leq d s + 2$. This completes the proof.
\end{proof}

\begin{proof}[of Lemma \ref{lem::VCFn}]
Recall that for an histogram transform splitting rule, we construct the  binary tree-based partition on the hypercube $[-r,r]^d$ with depth $p$, which leads to $s=2^{p}-1$ internal nodes of the partition. Consequently, applying lemma \ref{VCIndex}, we get
\begin{align*}
\mathrm{VC}(\mathcal{F})\leq d(2^p-1)+2\leq d\cdot 2^{p+1}.
\end{align*}
Thus we complete the proof.
\end{proof}

\begin{proof}[of Lemma \ref{thm::vart}]
Let $\mathcal{F}_{\varepsilon}$ be an $\varepsilon$-net over $\mathcal{F}$. 
Then, for any $f \in \mathrm{Co}(\mathcal{F})$, there exists an $f_{\varepsilon} \in \mathrm{Co}(\mathcal{F}_{\varepsilon})$
such that $\|f - f_{\varepsilon}\|_{L_2(\mathrm{Q})} \leq \varepsilon$.
Therefore, we can assume without loss of generality that $\mathcal{F}$ is finite.

Obviously, \eqref{CoverAssumption} holds for $1 \leq \varepsilon \leq c^{1/v}$.
Let $v' := 1/2 + 1/v$ and $M' := c^{1/v}M$. Then \eqref{CoverAssumption} implies that
for any $n \in \mathbb{N}$, there exists $f_1, \ldots, f_n \in \mathcal{F}$ such that
for any $f \in \mathcal{F}$, there exists an $f_i$ such that
\begin{align*}
\|f - f_i\|_{L_2(\mathrm{Q})} \leq M' n^{-1/v}.
\end{align*}
Therefore, for each $n \in \mathbb{N}$, we can find
sets $\mathcal{F}_1 \subset \mathcal{F}_2 \subset \cdots \subset \mathcal{F}$ such that 
the set $\mathcal{F}_n$ is a $M' n^{-1/v}$-net over $\mathcal{F}$ and $\#(\mathcal{F}_n) \leq n$.

In the following, we show by induction that for $q =7v/2$ and $n, k \geq 1$, there holds
\begin{align}\label{eq::main}
\log \mathcal{N} \bigl( \mathrm{Co}(\mathcal{F}_{nk^q}), L_2(\mathrm{Q}), c_k M' n^{-v'} \bigr)
\leq c'_k n,
\end{align}
where $c_k$ and $c'_k$ are constants depending only on $c$ and $v$ such that $\sup_k \max \{ c_k, c'_k \} < \infty$.
The proof of \eqref{eq::main} will be conducted by a nested induction argument.

Let us first consider the case $k = 1$. For a fixed $n_0$, let $n \leq n_0$. Then for $c_1$ satisfying $c_1 M' n_0^{-v'} \geq M$, there holds 
\begin{align*}
\log \mathcal{N} \bigl( \mathrm{Co}(\mathcal{F}_{nk^q}), L_2(\mathrm{Q}), c_k M' n^{-v'} \bigr) = 0,
\end{align*}
which immediately implies \eqref{eq::main}. For a general $n \in \mathbb{N}$, let $m := n/\ell$ for large enough $\ell$ to be chosen later. Then for any $f \in \mathcal{F}_n \setminus \mathcal{F}_m$, there exists an $f^{(m)} \in \mathcal{F}_m$ such that
\begin{align*}
\|f - f^{(m)}\|_{L_2(\mathrm{Q})} \leq M' m^{-1/v}.
\end{align*}
Let $\pi_m : \mathcal{F}_n \setminus \mathcal{F}_m \to \mathcal{F}_m$ be the projection operator. Then for any $f \in \mathcal{F}_n \setminus \mathcal{F}_m$, there holds
\begin{align*}
\|f - \pi_m f\|_{L_2(\mathrm{Q})} \leq M' m^{-1/v}
\end{align*}
Therefore, for $\lambda_i, \mu_j \geq 0$ and $\sum_{i=1}^n \lambda_i = \sum_{j=1}^m \mu_j = 1$, we have
\begin{align*}
\sum_{i=1}^n \lambda_i f^{(n)}_i 
= \sum_{j=1}^m \mu_j f^{(m)}_j 
+ \sum_{k=m+1}^n \lambda_k \bigl( f^{(n)}_k - \pi_m f^{(n)}_k \bigr).
\end{align*}
Let $\mathcal{G}_n$ be the set 
\begin{align*}
\mathcal{G}_n := \{ 0 \} \cup \{ f - \pi_m f : f \in \mathcal{F}_n \setminus \mathcal{F}_m \}.
\end{align*}
Then we have $\#(\mathcal{G}_n) \leq n$ and for any $g \in \mathcal{G}_n$, there holds
\begin{align*}
\|g\|_{L_2(\mathrm{Q})} 
\leq M'm^{-1/v}.
\end{align*}
Moreover, we have
\begin{align} \label{SpaceSplit}
\mathrm{Co}(\mathcal{F}_n) \subset \mathrm{Co}(\mathcal{F}_m) + \mathrm{Co}(\mathcal{G}_n).
\end{align}

Applying Lemma 2.6.11 in \citet{van1996weak} with $\varepsilon := \frac{1}{2} c_1 m^{1/v} n^{-v'}$ 
to $\mathcal{G}_n$, we can find a $\frac{1}{2}c_1 M'n^{-v'}$-net over $\mathrm{Co}(\mathcal{G}_n)$ consisting of at most 
\begin{align} \label{CapacityCoGn}
(e + e n \varepsilon^2)^{2/\varepsilon^2} 
\leq \biggl( e + \frac{e c_1^2}{\ell^{2/v}} \biggr)^{8 \ell^{2/v} c_1^{-2} n}
\end{align}
elements.

Suppose that \eqref{eq::main} holds for $k = 1$ and $n = m$.
In other words, there exists a $c_1 M' m^{-v'}$-net over $\mathrm{Co}(\mathcal{F}_m)$ consisting of at most $e^m$ elements, which partitions $\mathrm{Co}(\mathcal{F}_m)$ into $m$-dimensional cells of diameter at most $2 c_1 M' m^{-v'}$. 
Each of these cells can be isometrically identified with a subset of a ball of radius $c_1 M' m^{-v'}$ in $\mathbb{R}^m$
and can be therefore further partitioned into
\begin{align*}
\bigg(\frac{3c_1 M' m^{-v'}}{\frac{1}{2} c_1 M' n^{-v'}} \bigg)^m = (6\ell^{v'})^{n/\ell}
\end{align*}
cells of diameter $\frac{1}{2}c_1 M' n^{-v'}$. As a result, we get a $\frac{1}{2}c_1 M' n^{-v'}$-net of $\mathrm{Co}(\mathcal{F}_m)$ containing at most 
\begin{align} \label{CapacityCoFm}
e^m \cdot (6\ell^{v'})^{n/\ell}
\end{align}
elements.

Now, \eqref{SpaceSplit} together with \eqref{CapacityCoGn} and \eqref{CapacityCoFm}
yields that there exists a $c_1 M' n^{-v'}$-net of $\mathrm{Co}(\mathcal{F}_n)$ 
whose cardinality can be bounded by 
\begin{align*}
e^{n/\ell} \bigl( 6 \ell^{v'} \bigr)^{n/\ell} 
\biggl( e + \frac{e c_1^2}{\ell^{2/v}} \biggr)^{8 \ell^{2/v} c_1^{-2} n}
\leq e^n,
\end{align*}
for suitable choices of $c_1$ and $\ell$ depending only on $v$. 
This concludes the proof of \eqref{eq::main} for $k=1$ and every $n \in \mathbb{N}$.

Let us consider a general $k \in \mathbb{N}$. 
Similarly as above, there holds
\begin{align} \label{SpaceSplitGeneral}
\mathrm{Co}(\mathcal{F}_{nk^q}) \subset \mathrm{Co}(\mathcal{F}_{n(k-1)^q}) + \mathrm{Co}(\mathcal{G}_{n,k}),
\end{align}
where the set $\mathcal{G}_{n,k}$ contains at most $n k^q$ elements with norm smaller than $M'(n(k-1)^q)^{-1/v}$.
Applying Lemma 2.6.11 in \citet{van1996weak} to $\mathcal{G}_{n,k}$, we can find an $M'k^{-2}n^{-v'}$-net over $\mathrm{Co}(\mathcal{G}_{n,k})$ consisting of at most 
\begin{align} \label{CapacityCoGnk}
\bigl( e + e k^{2q/v-4+q} \bigr)^{2^{2q/v+1}k^{4-2q/v}n}
\end{align}
elements. Moreover, by the induction hypothesis, we have a $c_{k-1}M'n^{-v'}$-net over $\mathrm{Co}(\mathcal{F}_{n(k-1)^q})$ consisting of at most 
\begin{align} \label{CapacityCoFnk-1q}
e^{c'_{k-1}n}
\end{align}
elements. Using \eqref{SpaceSplitGeneral}, \eqref{CapacityCoGnk}, and \eqref{CapacityCoFnk-1q},
we obtain a $c_k M' n^{-v'}$-net over $\mathrm{Co}(\mathcal{F}_{nk^q})$ consisting of at most $e^{c'_k n}$ elements, where
\begin{align*}
c_k &= c_{k-1} + \frac{1}{k^2},
\\
c'_k &= c'_{k-1} + 2^{2q/v+1}\frac{1+\log(1+k^{2q/v-4+q})}{k^{2q/v-4}}.
\end{align*}
Form the elementary analysis we know that if $2q/v - 5 = 2$, 
then there exist constants $c''_1$, $c''_2$, and $c''_3$ such that
\begin{align*}
\lim_{k \to \infty} c_k 
& = c^{-1/v} n_0^{(v+2)/2v} + \sum_{i=2}^{\infty} 1/i^2 
\leq c''_1 c^{-1/v} + c''_2,
\\
\lim_{k \to \infty} c'_k 
& = 1 + c \sum_{i=1}^{\infty} 2 (2/i)^{2q/v}i^5 
\leq c''_3.
\end{align*}
Thus \eqref{eq::main} is proved. 
Taking $\varepsilon := c_k M' n^{-v'} / M$ in \eqref{eq::main}, we get 
\begin{align*}
\log \mathcal{N} ( \mathrm{Co}(\mathcal{F}_{nk^q}), L_2(\mathrm{Q}), M \varepsilon)
\leq c'_k c_k^{1/v'} (M')^{1/v'}M^{-1/v'}\varepsilon^{-1/v'}.
\end{align*}
This together with 
\begin{align*}(M')^{1/v'} = (c^{1/v}M)^{1/v'}=c^{2/(v+2)}M^{1/v'}
\end{align*}
yields
\begin{align*}
\log \mathcal{N} ( \mathrm{Co}(\mathcal{F}), L_2(\mathrm{Q}), M \varepsilon)
\leq c' c^{2/(v+2)}\varepsilon^{-2v/(v+2)},
\end{align*}
where the constant $c'$ depends on the constants $c''_1$, $c''_2$ and $c''_3$.
This finishes the proof.
\end{proof}

\begin{proof}[of Proposition \ref{the::Fncovering}]
A direct application of Theorem 2.6.7 in \citet{van1996weak} yields the assertion.
\end{proof}

\begin{proof}[of Proposition \ref{the::convexFn}]
The assertion follows directly from Lemma \ref{thm::vart} with 
\begin{align*}
c := c_0 d\cdot 2^{p+1} (16e)^{d\cdot 2^{p+1}}
\qquad
\text{ and }
\qquad
v := 2(d\cdot 2^{p+1}-1).
\end{align*}
Then we have
\begin{align*}
c^{2/(v+2)} = (c_0\delta^{-1}(16e)^{1/\delta})^{\delta}
= 16e (c_0 d\cdot 2^{p+1})^{1/(d\cdot 2^{p+1})}.
\end{align*}
Let $g(t):=(c_0t)^{1/t}$, $t=d\cdot 2^{p+1}\geq 4d$, then we have $g'(t)=(c_0t)^{1/t}(1-\log(c_0t))/t^2$. Therefore, there exists $c_1'$ only depending on $d$ and $c_0$ such that $g(t)\leq c_1$ for all $t\geq d\cdot 2^{p+1}$.
Consequently, we have
\begin{align*}
\log \mathcal{N} ( \mathrm{Co}(\mathcal{F}), L_2(\mathrm{Q}), M \varepsilon) \leq 16 e c_1' \varepsilon^{1/(d\cdot 2^p)-2}\leq 16ec_1' \cdot 2^{p/2} \varepsilon^{1/(d\cdot 2^p)-2}.
\end{align*}
Hence we obtain the assertion with $c_1:=16 ec_1'$.
\end{proof}

\subsection{Proofs for $f^*_{L,\mathrm{P}} \in C^{0,\alpha}$}

\begin{proof}[of Proposition \ref{pro::c0alphasingle}]
The assumption $f_{L, \mathrm{P}}^* \in C^{0,\alpha}$ implies that for any $x\in \mathcal{X}$, there holds
\begin{align*}
|f_{\mathrm{P}}^p(X) - f_{L, \mathrm{P}}^*|
& = \biggl| \frac{1}{\mathrm{P}_X(A_p(x'))} \int_{A_p(x')}f_{L, \mathrm{P}}^*(x') 
                       - f_{L, \mathrm{P}}^*(x)  \, d \mathrm{P}_X(x') \biggr|
\\
& \leq \frac{1}{\mathrm{P}_X(A_p(x'))} \int_{A_p(x')}|f_{L, \mathrm{P}}^*(x') 
                       - f_{L, \mathrm{P}}^*(x)|  \, d \mathrm{P}_X(x')
\\
& \leq \frac{1}{\mathrm{P}_X(A_p(x'))} \int_{A_p(x')} \mathrm{diam}(A_p(x))^{\alpha} \, d \mathrm{P}_X(x')
\\
& = \mathrm{diam}(A_p(x))^{\alpha}.
\end{align*}
Therefore, we get
\begin{align*}
\mathcal{R}_{L,\mathrm{P}}(f_{\mathrm{P}}^p) - \mathcal{R}_{L,\mathrm{P}}^*
= \mathbb{E}_{\mathrm{P}_X}(f_{\mathrm{P}}^p(X) - f_{L, \mathrm{P}}^*)^2
\leq \mathbb{E}_{\mathrm{P}_X}\mathrm{diam}(A_p(x))^{2\alpha}.
\end{align*}
Consequently we obtain
\begin{align*}
\mathbb{E}_{\mathrm{P}_{R,Z}} \bigl( \lambda h^{-2d} + \mathcal{R}_{L,\mathrm{P}}(f_{\mathrm{P}}^p) - \mathcal{R}_{L,\mathrm{P}}^* \bigr)
& \leq \lambda \cdot 4^p + \mathbb{E}_{\mathrm{P}_{R,Z}}(\mathbb{E}_{\mathrm{P}_X}\mathrm{diam}(A_p(x))^{2\alpha})
\\
& = \lambda \cdot 4^p+ \mathbb{E}_{\mathrm{P}_X}(\mathbb{E}_{\mathrm{P}_{R,Z}}\mathrm{diam}(A_p(x))^{2\alpha}).
\end{align*}
Applying Lemma \ref{lem::diamapx}, we get
\begin{align*}
\mathbb{E}_{\mathrm{P}_{R,Z}} \bigl( \lambda \cdot 4^p+ \mathcal{R}_{L,\mathrm{P}}(f_{\mathrm{P}}^p) - \mathcal{R}_{L,\mathrm{P}}^* \bigr)
& \leq \lambda \cdot 4^p+(2r)^{2\alpha}d\exp\biggl(\frac{(4^{-\alpha}-1)p}{d}\biggr)
\\
& = \lambda\cdot 4^p+(2r)^{2\alpha}d(2^{-p})^{\frac{1-4^{-\alpha}}{d\log 2}}
\\
& \leq c\lambda^{\frac{1-4^{-\alpha}}{2d\log 2+(1-4^{-\alpha})}},
\end{align*}
by choosing
\begin{align*}
p=\frac{d}{2d\log 2+1-4^{-\alpha}}\log \biggl(\frac{(2r)^{2\alpha}d}{\lambda}\biggr),
\end{align*}
where the constant $c := ((2r)^{2\alpha}d)^{\frac{d\log2}{1-4^{-\alpha}+2d\log 2}}$.
\end{proof}

\subsubsection{Proofs Related to Section \ref{sec::upperc0}}

\begin{proof}[of Theorem \ref{thm::OracalBoost}]
Denote
\begin{align*}
r^*:=\Omega_{\lambda}(f)+\mathcal{R}_{L,\mathrm{P}}(f)-R^*_{L,\mathrm{P}},
\end{align*}
and for $r > r^*$, we write
\begin{align*}
\mathcal{F}_r & := \{ f \in E : \Omega_{\lambda}(f) + \mathcal{R}_{L,\mathrm{P}}(f) - \mathcal{R}^*_{L,\mathrm{P}} \leq r \},
\\
\mathcal{H}_r & := \{ L \circ f - L \circ f^*_{L,\mathrm{P}} : f \in \mathcal{F}_r \}.
\end{align*}
Note that for $f \in \mathcal{F}_r$, we have $\lambda_1 \|f\|_E \leq r$, that is,
\begin{align*}
\sum_{i=1}^T |w_i|^2 
\leq r/\lambda_1.
\end{align*}
Then, by the Cauchy-Schwarz inequality, we get
\begin{align*}
\sum_{i=1}^T |w_i|
\leq \biggl( T \sum_{i=1}^T |w_i|^2 \biggr)^{1/2}
\leq (rT/\lambda_1)^{1/2}.
\end{align*}
Consequently, we have $\mathcal{F}_r \subset (r T/\lambda_1)^{1/2} B_E$. 
Since $L$ is Lipschitz continuous with $|L|_1 \leq 4M$, we find
\begin{align*}
\mathbb{E}_{D \sim \mathrm{P}^n} e_i(\mathcal{H}_r, L_2(\mathrm{D}))
& \leq 4 M \mathbb{E}_{D \sim \mathrm{P}_X^n} e_i(\mathcal{F}_r, L_2(\mathrm{D}))
\\
& \leq 8 M (r T/\lambda_1)^{1/2} \mathbb{E}_{D \sim \mathrm{P}_X^n} e_i(B_E, L_2(\mathrm{D}))
\\
& \leq 8 M (r T/\lambda_1)^{1/2} \mathbb{E}_{D \sim \mathrm{P}_X^n} e_i(\mathrm{Co}(\mathcal{F}), L_2(\mathrm{D})).
\end{align*}
Let $\delta := 1/(d\cdot 2^p)$, $\delta' := 1 - \delta$, and $a := c_1^{1/(2\delta')} M$. 
Then \eqref{equ::convexFn} together with \eqref{CoverEntropy}  implies that
\begin{align*}
e_i(\mathrm{Co}(\mathcal{F}), L_2(\mathrm{D}))
\leq (3c_1)^{1/(2\delta')}(2^p)^{1/(2\delta')} i^{-1/(2\delta')}
\end{align*}
Taking expectation with respect to $\mathrm{P}^n$, we get
\begin{align}\label{equ::emcofH}
\mathbb{E}_{D \sim \mathrm{P}_X^n} e_i(\mathrm{Co}(\mathcal{F}),L_2(\mathrm{D})) \leq c_2' (2^p)^{1/(2\delta')} i^{-1/(2\delta')},
\end{align}
where $c_2' := (3c_1)^{1/(2\delta')} $. Moreover, we easily find
\begin{align*}
\lambda_2\cdot 4^p
= \Omega(h)
\leq \Omega_{\lambda}(f) + \mathcal{R}_{L,\mathrm{P}}(f)-\mathcal{R}^*_{L,\mathrm{P}}
\leq r.
\end{align*}
Therefore, \eqref{equ::emcofH} can be further estimated by
\begin{align*}
\mathbb{E}_{D \sim \mathrm{P}_X^n} e_i(\mathrm{Co}(\mathcal{F}), L_2(\mathrm{D}))
\leq c_2' (r/\lambda_2)^{1/(4\delta')} i^{-1/(2\delta')},
\end{align*}
which leads to 
\begin{align*}
\mathbb{E}_{D \sim \mathrm{P}_X^n} e_m(\mathcal{H}_r, L_2(\mathrm{D}))
\leq 8 c_2 M (r T/\lambda_1)^{1/2} (r/\lambda_2)^{1/(4\delta')}i^{-1/(2\delta')}.
\end{align*}
For the least square loss, the superemum bound
\begin{align*}
L(x,y,t) \leq 4 M^2,
\qquad \forall \, (x,y) \in \mathcal{X} \times \mathcal{Y}, \, t \in [-M, M],
\end{align*}
and the variance bound
\begin{align*}
\mathbb{E}(L \circ g - L \circ f_{L,\mathrm{P}}^*)^2
\leq V (\mathbb{E}(L \circ g - L \circ f^*_{L,\mathrm{P}}))^{\vartheta}
\end{align*}
holds for $V = 16 M^2$ and $\vartheta = 1$. Therefore, for $h \in \mathcal{H}_r$, we have
\begin{align*}
\|h\|_{\infty} \leq 8 M^2,
\qquad 
\mathbb{E}_{\mathrm{P}} h^2 \leq 16 M^2 r.
\end{align*}
Then Theorem 7.16 in \citet{StCh08} with $a := 8 c_2 M (r T/\lambda_1)^{1/2} (r/\lambda_2)^{1/(4\delta')}$ yields that there exist a constant $c_0' > 0$ such that
\begin{align*}
& \mathbb{E}_{D \sim \mathrm{P}^n} \mathrm{Rad}_D(\mathcal{H}_r,n)
\\
& \leq c_0' \max \Bigl\{ r^{3/4} T^{\delta'/2} \lambda_1^{-\delta'/2} \lambda_2^{-1/4} n^{-1/2},
r^{(2\delta'+1)/(2\delta'+2)} (T/\lambda_1)^{\delta'/(1+\delta')}
\lambda_2^{1/(2+2\delta')}n^{-1/(1+\delta')} \Bigr\}
\\
& =: \varphi_n(r).
\end{align*}
Simple algebra shows that the condition $\varphi_n(4r) \leq 2\sqrt{2} \varphi_n(r)$ is satisfied. Since $2\sqrt{2} < 4$, similar arguments show that there still hold the statements of the Peeling Theorem 7.7 in \citet{StCh08}. Consequently, Theorem 7.20 in \citet{StCh08} can also be applied, if the assumptions on $\varphi_n$ and $r$ are modified to $\varphi_n(4r) \leq 2\sqrt{2} \varphi_n(r)$ and
$r \geq \max\{75\varphi_n(r), 1152M^2\tau/n, r^*\}$, respectively. It is easy to verify that the condition $r\geq 75\varphi_n(r)$ is satisfied if
\begin{align*}
r \geq c_0' T^{2\delta'} \lambda_1^{-2\delta'} \lambda_2^{-1} n^{-2},
\end{align*}
where $c_0'$ is a constant, which yields the assertion.
\end{proof}

\subsubsection{Proofs Related to Section \ref{sec::upperc0large}}

\begin{proof}[of Theorem \ref{equ::OracalBoostlarge}]
Denote
\begin{align*}
\bar{r}^*:=\bar{\Omega}_{\lambda}(f)+\mathcal{R}_{L,\mathrm{P}}(f)-R^*_{L,\mathrm{P}},
\end{align*}
and for $r > \bar{r}^*$, we write
\begin{align*}
\bar{\mathcal{F}}_r & := \{ f \in \bar{E} : \bar{\Omega}_{\lambda}(f) + \mathcal{R}_{L,\mathrm{P}}(f) - \mathcal{R}^*_{L,\mathrm{P}} \leq r \},
\\
\bar{\mathcal{H}}_r & := \{ L \circ f - L \circ f^*_{L,\mathrm{P}} : f \in \mathcal{F}_r \}.
\end{align*}
Note that for $f \in \bar{\mathcal{F}}_r$, we have $\lambda_1 \|f\|_{\bar{E}} \leq r$, that is,
\begin{align*}
\sum_{i=1}^T |w_i|^2 
\leq r/\lambda_1.
\end{align*}
Then, by the Cauchy-Schwarz inequality, we get
\begin{align*}
\sum_{i=1}^T |w_i|
\leq \biggl( T \sum_{i=1}^T |w_i|^2 \biggr)^{1/2}
\leq (rT/\lambda_1)^{1/2}.
\end{align*}
Consequently, we have $\bar{\mathcal{F}}_r \subset (r T/\lambda_1)^{1/2} B_{\bar{E}}$. 
Since $L$ is Lipschitz continuous with $|L|_1 \leq 4M$, we find
\begin{align*}
\mathbb{E}_{D \sim \mathrm{P}^n} e_i(\bar{\mathcal{H}}_r, L_2(\mathrm{D}))
& \leq 4 M \mathbb{E}_{D \sim \mathrm{P}_X^n} e_i(\bar{\mathcal{F}}_r, L_2(\mathrm{D}))
\\
& \leq 8 M (r T/\lambda_1)^{1/2} \mathbb{E}_{D \sim \mathrm{P}_X^n} e_i(B_{\bar{E}}, L_2(\mathrm{D}))
\end{align*}
By the definition of $\bar{E}$ in \eqref{equ::barEn}, we have $B_{\bar{E}}\subset \mathrm{Co}(\mathcal{F})$. Consequently, we find
\begin{align*}
\mathbb{E}_{D \sim \mathrm{P}^n} e_i(\bar{\mathcal{H}}_r, L_2(\mathrm{D}))\leq 8 M (r T/\lambda_1)^{1/2} \mathbb{E}_{D \sim \mathrm{P}_X^n} e_i(\mathrm{Co}(\mathcal{F}), L_2(\mathrm{D})).
\end{align*}
Through the similar arguments in the proof of Theorem \ref{thm::OracalBoost}, we are able to obtain the desired assertion. Thus, we omit the details here.
\end{proof}

\subsubsection{Proofs Related to Section \ref{sec::c0}}

\begin{proof}[of Theorem \ref{thm::tree}]
It is easy to see that $f_{\mathrm{P},\mathrm{E}}$ defined by \eqref{fpeensemble} satisfies $f_{\mathrm{P},\mathrm{E}}\in E$ and $\lambda_1 \|f_{\mathrm{P},\mathrm{E}}\|_E\leq \lambda_1/T$. Moreover, by Jensen's inequality, we have
\begin{align*}
\mathcal{R}_{L,\mathrm{P}}(f_{\mathrm{P},\mathrm{E}})-\mathcal{R}^*_{L,\mathrm{P}}
& = \int_{\mathcal{X}} \biggl( \frac{1}{T} \sum_{t=1}^T f_{\mathrm{P}}^{p,t} - f_{L, \mathrm{P}}^* \biggr)^2 \, d\mathrm{P}_X
\\
& \leq \frac{1}{T} \sum_{t=1}^T \int_{\mathcal{X}} (f_{\mathrm{P}}^{p,t} - f_{L, \mathrm{P}}^*)^2 \, d\mathrm{P}_X
\\
& = \frac{1}{T} \sum_{t=1}^T\biggl( \mathcal{R}_{L, \mathrm{P}}(f_{\mathrm{P}}^{p,t}) - \mathcal{R}_{L,\mathrm{P}}^*\biggr).
\end{align*}
Combining this with Proposition \ref{pro::c0alphasingle}, we obtain
\begin{align*}
\mathbb{E}_{\mathrm{P}_{R,Z}} \mathcal{R}_{L,\mathrm{P}}(f_{\mathrm{P},\mathrm{E}})-\mathcal{R}^*_{L,\mathrm{P}}
& \leq (2r)^{2\alpha}d(2^{-p})^{\frac{1-4^{-\alpha}}{d\log 2}}.
\end{align*}
Consequently we get
\begin{align*}
\mathbb{E}_{\mathrm{P}_{R,Z}} A(\lambda)
& = \mathbb{E}_{\mathrm{P}_{R,Z}}\biggl(\inf_{f \in E} \lambda_1 \|f\|_E + \lambda_2 \Omega(h) + \mathcal{R}_{L,\mathrm{P}}(f) - \mathcal{R}^*_{L,\mathrm{P}}\biggr)
\\
& \leq \lambda_1 \|f_{\mathrm{P},\mathrm{E}}\|_E + \lambda_2 \Omega(h)           
+\mathbb{E}_{\mathrm{P}_{R,Z}}\bigl( \mathcal{R}_{L,\mathrm{P}}(f_{\mathrm{P},\mathrm{E}}) - \mathcal{R}^*_{L,\mathrm{P}}\bigr)
\\
& \leq \lambda_1/T + c \cdot \lambda_2^{\frac{1-4^{-\alpha}}{2d\log 2+(1-4^{-\alpha})}}.
\end{align*}
Then Theorem \ref{thm::OracalBoost} implies that with probability $\mathrm{P}^n$ not less than $1-3/n^2$, there holds
\begin{align}
& \mathbb{E}_{\mathrm{P}_{R,Z}} \bigl( 
\lambda_1 \|f\|_E + \lambda_2 \Omega(p) + \mathcal{R}_{L,\mathrm{D}}(f_{\mathrm{D},B}) - \mathcal{R}^*_{L,\mathrm{P}}
\bigr)
\nonumber\\
& \leq 6 \lambda_1/T + 6 c \lambda_2^{\frac{1-4^{-\alpha}}{2d\log 2+(1-4^{-\alpha})}}+ 3 c_0' T^{2\delta'} \lambda_1^{-2\delta'} \lambda_2^{-1} n^{-2} 
+ 3456 M^2 (2\log n)/ n,
\label{equ::c0lambda1}
\end{align}
where $\delta=1/(d\cdot 2^p)$, $c$ and $c_0'$ are constants defined as in Proposition \ref{pro::c0alphasingle} and Theorem \ref{thm::OracalBoost}. Minimizing the right hand side of \eqref{equ::c0lambda1}, we get 
\begin{align}\label{equ::c0rate}
\mathbb{E}_{\mathrm{P}_{R,Z}}\bigl(\mathcal{R}_{L,\mathrm{P}}(f_{\mathrm{D},B}) - \mathcal{R}^*_{L,\mathrm{P}}\bigr)
\leq c'' n^{-\frac{2(1-4^{-\alpha})}{(4-2\delta)(1-4^{-\alpha})+2d\log 2}},
\end{align}
if we choose 
\begin{align*}
\lambda_{1,n} = n^{-\frac{2(1-4^{-\alpha})}{(4-2\delta)(1-4^{\alpha})+2d\log 2}}, \,
\lambda_{2,n} = n^{-\frac{2(1-4^{-\alpha}+2d\log 2)}{(4-2\delta)(1-4^{-\alpha})+2d\log 2}}, \,
p_n \asymp \frac{2d\log 2\log n}{(4-2\delta)(1-4^{-\alpha})+2d\log 2},
\end{align*}
where $c''$ is a constant depending on $c$, $c_0'$, $d$, $M$, $R$ and $T$. Consequently, by the Borel-Cantelli Lemma, \eqref{equ::c0rate} holds with probability $\mathrm{P}^n$ equal to one for sufficiently large $n$, which leads to the desired assertion.
\end{proof}

\subsubsection{Proofs Related to Section \ref{sec::c0large}}

\begin{proof}[of Theorem \ref{thm::c01gbbhte}]
It is easy to see that $\bar{f}_{\mathrm{P},\mathrm{E}}$ defined by \eqref{equ::barfpe} satisfies $\bar{f}_{\mathrm{P},\mathrm{E}}\in E$ and $\lambda_1 \|\bar{f}_{\mathrm{P},\mathrm{E}}\|_E\leq \lambda_1/T$. Moreover, by Jensen's inequality, we have
\begin{align*}
\mathcal{R}_{L,\mathrm{P}}(\bar{f}_{\mathrm{P},\mathrm{E}})-\mathcal{R}^*_{L,\mathrm{P}}
& = \int_{\mathcal{X}} \biggl( \frac{1}{T} \sum_{t=1}^T \bar{f}_{\mathrm{P}}^{p,t} - f_{L, \mathrm{P}}^* \biggr)^2 \, d\mathrm{P}_X
\\
& \leq \frac{1}{T} \sum_{t=1}^T \int_{\mathcal{X}} (\bar{f}_{\mathrm{P}}^{p,t} - f_{L, \mathrm{P}}^*)^2 \, d\mathrm{P}_X
\\
&\leq  \frac{1}{T} \sum_{t=1}^T \int_{\mathcal{X}} \frac{1}{K}\sum_{k=1}^K (f_{\mathrm{P}}^{p,t,k} - f_{L, \mathrm{P}}^*)^2 \, d\mathrm{P}_X
\\
& = \frac{1}{TK} \sum_{t=1}^T\sum_{k=1}^K\biggl( \mathcal{R}_{L, \mathrm{P}}(f_{\mathrm{P}}^{p,t,k}) - \mathcal{R}_{L,\mathrm{P}}^*\biggr).
\end{align*}
Combining this with Proposition \ref{pro::c0alphasingle}, we obtain
\begin{align*}
\mathbb{E}_{\mathrm{P}_{R,Z}} \bigl( \mathcal{R}_{L,\mathrm{P}}(f_\mathrm{P}^{t,k})-\mathcal{R}^*_{L,\mathrm{P}} \bigr)
& \leq (2r)^{2\alpha}d(2^{-p})^{\frac{1-4^{-\alpha}}{d\log 2}}.
\end{align*}
Consequently we get
\begin{align*}
\mathbb{E}_{\mathrm{P}_{R,Z}} \bar{A}(\lambda)
& = \mathbb{E}_{\mathrm{P}_{R,Z}}\biggl(\inf_{f \in \bar{E}} \lambda_1 \|f\|_{\bar{E}} + \lambda_2 \Omega(h) + \mathcal{R}_{L,\mathrm{P}}(f) - \mathcal{R}^*_{L,\mathrm{P}}\biggr)
\\
& \leq \lambda_1 \|\bar{f}_{\mathrm{P},\mathrm{E}}\|_{\bar{E}}+ \lambda_2 \Omega(p)           
+\mathbb{E}_{\mathrm{P}_{R,Z}}\bigl( \mathcal{R}_{L,\mathrm{P}}(\bar{f}_{\mathrm{P},\mathrm{E}}) - \mathcal{R}^*_{L,\mathrm{P}}\bigr)
\\
& \leq \lambda_1/T + c \cdot \lambda_2^{\frac{1-4^{-\alpha}}{2d\log 2+(1-4^{-\alpha})}}.
\end{align*}
Then, Theorem \ref{equ::OracalBoostlarge} implies that with probability $\mathrm{P}^n$ not less than $1-3/n^2$, there holds
\begin{align}
& \mathbb{E}_{\mathrm{P}_{R,Z}}  \bigl(
\lambda_1 \|f\|_{\bar{E}} + \lambda_2 \Omega(p) + \mathcal{R}_{L,\mathrm{D}}(\bar{f}_{\mathrm{D},B}) - \mathcal{R}^*_{L,\mathrm{P}} \bigr)
\nonumber\\
& \leq 6 \lambda_1/T + 6 c \lambda_2^{\frac{1-4^{-\alpha}}{2d\log 2+(1-4^{-\alpha})}}+ 3 c_0' T^{2\delta'} \lambda_1^{-2\delta'} \lambda_2^{-1} n^{-2} 
+ 3456 M^2 (2\log n)/ n,
\label{equ::c0lambda1large}
\end{align}
where $\delta=1/(d\cdot 2^p)$, $c$ and $c_0'$ are constants defined as in Proposition \ref{pro::c0alphasingle} and Theorem \ref{thm::OracalBoost}. Minimizing the right hand side of \eqref{equ::c0lambda1}, we get 
\begin{align}\label{equ::c0ratelarge}
\mathbb{E}_{\mathrm{P}_{R,Z}}\bigl(\mathcal{R}_{L,\mathrm{P}}(\bar{f}_{\mathrm{D},B}) - \mathcal{R}^*_{L,\mathrm{P}}\bigr)
\leq c'' n^{-\frac{2(1-4^{-\alpha})}{(4-2\delta)(1-4^{-\alpha})+2d\log 2}},
\end{align}
if we choose 
\begin{align*}
\lambda_{1,n} = n^{-\frac{2(1-4^{-\alpha})}{(4-2\delta)(1-4^{\alpha})+2d\log 2}}, \,
\lambda_{2,n} = n^{-\frac{2(1-4^{-\alpha}+2d\log 2)}{(4-2\delta)(1-4^{-\alpha})+2d\log 2}}, \,
p_n \asymp \frac{2d\log 2\log n}{(4-2\delta)(1-4^{-\alpha})+2d\log 2},
\end{align*}
where $c''$ is a constant depending on $c$, $c_0'$, $d$, $M$, $R$ and $T$. Consequently, by the Borel-Cantelli Lemma, \eqref{equ::c0rate} holds with probability $\mathrm{P}^n$ equal to one for sufficiently large $n$, which leads to the desired assertion.
\end{proof}

\subsection{Proofs for $f^*_{L,\mathrm{P}}\in C^{1,0}$}

\subsubsection{Proofs Related to Section \ref{sec::upperc1}}

\begin{proof}[of Proposition \ref{prop::biasterm}]
According to the rotated binary histogram splitting rule, the rotation transformation $\{H_t\}_{t=1}^T$ and split coordinates $\{Z_t\}_{t=1}^T$ are i.i.d. Therefore, for any $x \in B_{r,d}$, the expected approximation error term can be decomposed as follows:
\begin{align}
&\mathbb{E}_{\mathrm{P}_{R,Z}}  \bigl( f_{\mathrm{P},\mathrm{E}}(x)- f_{L, \mathrm{P}}^*(x) \bigr)^2 \nonumber\\
& =\mathbb{E}_{\mathrm{P}_{R,Z}} \bigl( 
(f_{\mathrm{P},\mathrm{E}}(x) - \mathbb{E}_{\mathrm{P}_{R,Z}}(f_{\mathrm{P},\mathrm{E}}(x)) )
+ \mathbb{E}_{\mathrm{P}_{R,Z}}(f_{\mathrm{P},\mathrm{E}}(x)) - f_{L,\mathrm{P}}^*(x)) \bigr)^2
\nonumber\\
& =  \mathrm{Var}_{\mathrm{P}_{R,Z}}(f_{\mathrm{P},\mathrm{E}}(x))
+ (\mathbb{E}_{\mathrm{P}_{R,Z}}(f_{\mathrm{P},\mathrm{E}}(x))-f_{L,\mathrm{P}}^*(x))^2
\nonumber\\
& = \frac{1}{T} \cdot \mathrm{Var}_{\mathrm{P}_{R,Z}}(f_{\mathrm{P}}^{p,1}(x))
+ \bigl(\mathbb{E}_{\mathrm{P}_{R,Z}}( f_{\mathrm{P}}^{p,1}(x) ) - f_{L, \mathrm{P}}^*(x) \bigr)^2.
\label{equ::biasvarianceDecom}
\end{align}

For the first term in \eqref{equ::biasvarianceDecom}, the assumption $f_{L,\mathrm{P}}^* \in C^{1,0}$ implies
\begin{align}\label{equ::first}
\mathrm{Var}_{\mathrm{P}_{R,Z}} \bigl( f_{\mathrm{P}}^p(x) \bigr)
& = \mathbb{E}_{\mathrm{P}_{R,Z}} \bigl( f_{\mathrm{P}}^p(x) - \mathbb{E}_{\mathrm{P}_{R,Z}}(f_{\mathrm{P}}^p(x)) \bigr)^2
\nonumber\\
& \leq \mathbb{E}_{\mathrm{P}_{R,Z}} \bigl( f_{\mathrm{P}}^p(x) - f_{L, \mathrm{P}}^*(x) \bigr)^2
\nonumber\\
& = \mathbb{E}_{\mathrm{P}_{R,Z}}\biggl( \frac{1}{\mathrm{P}_X(A_p(x))} \int_{A_p(x)} f_{L, \mathrm{P}}^*(x') \, d\mathrm{P}_X(x')
- f_{L, \mathrm{P}}^*(x) \biggr)^2
\nonumber\\
& = \mathbb{E}_{\mathrm{P}_{R,Z}} \biggl( \frac{1}{\mathrm{P}_X(A_p(x))} \int_{A_p(x)} \bigl( f_{L, \mathrm{P}}^*(x') - f_{L, \mathrm{P}}^*(x) \bigr) \, d\mathrm{P}_X(x') \biggr)^2
\nonumber\\
& \leq \mathbb{E}_{\mathrm{P}_{R,Z}} \bigl( c_L \mathrm{diam} \bigl( A_p(x) \bigr) \bigr)^2
\end{align}
According to Lemma \ref{lem::diamapx}, the first term is further bounded by
\begin{align}\label{equ::varfppx}
\mathrm{Var}_{\mathrm{P}_{R,Z}} \bigl( f_{\mathrm{P}}^p(x) \bigr)\leq c_L^2(2r)^4 d\exp \biggl(-\frac{0.75p}{d}\biggr).
\end{align}

We now consider the second term in \eqref{equ::biasvarianceDecom}. 
Taking the first-order Taylor expansion of $f_{L,\mathrm{P}}^*(x')$ at $x$, we get
\begin{align} \label{TaylorExpansion}
f_{L,\mathrm{P}}^*(x') - f_{L,\mathrm{P}}^*(x)
& = \int_0^1 \bigl( \nabla f_{L,\mathrm{P}}^*(x + t(x' - x)) \bigr)^{\top} (x' - x) \, dt. 
\end{align}
Therefore, there holds
\begin{align*}
& \bigl| f_{L,\mathrm{P}}^*(x') - f_{L,\mathrm{P}}^*(x) - \nabla f_{L,\mathrm{P}}^*(x)^{\top} (x' - x) \bigr|
\\
& = \biggl| \int_0^1 \bigl( \nabla f_{L,\mathrm{P}}^*(x + t(x' - x)) - \nabla f_{L,\mathrm{P}}^*(x) \bigr)^{\top} (x' - x) \, dt \biggr|
\\
& \leq \int^1_0 c_L (t \|x' - x\|_2)^{\alpha} \|x' - x\|_2 \, dt
\\
& \leq c_L \|x' - x\|^{1+\alpha}. 
\end{align*}
Now, by the triangle inequality, we have
\begin{align*}
&\biggl|\mathbb{E}_{\mathrm{P}_{R,Z}}\biggl(\frac{1}{\mathrm{P}_X(A_p(x))}\int_{A_p(x)}(f^*_{L,\mathrm{P}}(x') -f^*_{L,\mathrm{P}}(x))d\mathrm{P}_X(x')\biggr)\biggr|\\
&\quad-\biggl|\mathbb{E}_{\mathrm{P}_{R,Z}}\biggl(\frac{1}{\mathrm{P}_X(A_p(x))}\int_{A_p(x)} \nabla f^*_{L,\mathrm{P}}(x)^\top (x'-x)d\mathrm{P}_X(x')\biggr)\biggr|\\
&\leq \biggl|\mathbb{E}_{\mathrm{P}_{R,Z}}\biggl(\frac{1}{\mathrm{P}_X(A_p(x))}\int_{A_p(x)}(f^*_{L,\mathrm{P}}(x')-f^*_{L,\mathrm{P}}(x)-\nabla f^*_{L,\mathrm{P}}(x)^\top(x'-x))d\mathrm{P}(x')\biggr)\biggr|\\
&\leq \mathbb{E}_{\mathrm{P}_{R,Z}}\biggl(\frac{c_L}{\mathrm{P}_X(A_p(x))} \int_{A_p(x)}\|x'-x\|^{1+\alpha}d\mathrm{P}(x')\biggr)\\
&\leq c_L\mathbb{E}_{\mathrm{P}_{R,Z}}(\mathrm{diam}(A_p(x))^{1+\alpha}).
\end{align*}
Then we get
\begin{align}\label{equ::eprpzfpp1x}
&\big|\mathbb{E}_{\mathrm{P}_{R,Z}}( f_{\mathrm{P}}^{p,1}(x) ) - f_{L, \mathrm{P}}^*(x)\big|\nonumber \\
&\leq \biggl|\mathbb{E}_{\mathrm{P}_{R,Z}}\biggl(\frac{1}{\mathrm{P}_X(A_p(x))}\int_{A_p(x)} \nabla f^*_{L,\mathrm{P}}(x)^\top (x'-x)d\mathrm{P}_X(x')\biggr|+c_L\mathbb{E}_{\mathrm{P}_{R,Z}}(\mathrm{diam}(A_p(x))^{1+\alpha})\nonumber\\
&\leq  \mathbb{E}_{\mathrm{P}_{R,Z}}\biggl(\frac{1}{c_{r,d}\mathrm{P}_X(A_p(x))}\biggl|\int_{A_p(x)\cap B_{r,d}} \nabla f^*_{L,\mathrm{P}}(x)^\top (x'-x)dx'\biggr|\biggr) +c_L\mathbb{E}_{\mathrm{P}_{R,Z}}(\mathrm{diam}(A_p(x))^{1+\alpha}),
\end{align}
where $c_{r,d}:=d^{d/2}/((2r)^d)$ is a constant. By making the substitution $x'=R^{-1}\cdot \tilde{x}'$ and writing $\tilde{x}:=R\cdot x$, we obtain
\begin{align}\label{equ::apxbrd}
\biggl| \int_{A_p(x)\cap B_{r,d}} \nabla f^*_{L,\mathrm{P}}(x)^\top (x'-x)dx' \biggr|&=\biggl| \int_{H(A_p(x)\cap B_{r,d})} \nabla f^*_{L,\mathrm{P}}(x)^\top R^{-1} (\tilde{x}'-\tilde{x})d\tilde{x}' \biggr|\nonumber\\
&=\biggl| \int_{H(A_p(x)\cap B_{r,d})}(R\cdot \nabla f^*_{L,\mathrm{P}}(x))^\top
(\tilde{x}'-\tilde{x})d\tilde{x}' \biggr|.
\end{align}
Since $f^*_{L,\mathrm{P}}\in C^{1,0}$ we have $\|\nabla f^*_{L,\mathrm{P}}\|\leq c_L$, recall that $R$ is a orthogonal matrix, thus we find
\begin{align*}
\|R\cdot \nabla f^*_{L,\mathrm{P}}(x)\|_{\infty}\leq \|R\cdot \nabla f^*_{L,\mathrm{P}}(x)\|=\sqrt{(R \cdot \nabla f^*_{L,\mathrm{P}}(x))^{\top}H \nabla f^*_{L,\mathrm{P}}(x)}=\|\nabla f^*_{L,\mathrm{P}}(x)\|\leq c_L.
\end{align*}
Together with \eqref{equ::apxbrd}, we find
\begin{align}\label{equ::fppxflpx}
\biggl| \int_{A_p(x)\cap B_{r,d}} \nabla f^*_{L,\mathrm{P}}(x)^\top (x'-x)dx' \biggr|\leq c_L\sum^d_{j=1} \biggl|\int  _{H(A_p(x)\cap B_{r,d})} (\tilde{x}_j'-\tilde{x}_j)d\tilde{x}'\biggr|,
\end{align}
where $\tilde{x}_j$ and $\tilde{x}_j'$ denote the $j$-th entries of the vectors $\tilde{x}$ and $\tilde{x}'$ respectively.

Recall that $\underline{a}_p^j(x)$ and $\overline{a}_p^j(x)$ denotes the minimum and maximum values of the $j$-th entries of points in  $H(A_p(x))$. Moreover, by the construction of rotated binary histogram, there holds
\begin{align*}
H(A_p(x))
= [\underline{a}_p^1(x), \overline{a}_p^1(x)] \times \cdots \times [\underline{a}_p^d(x), \overline{a}_p^d(x)].
\end{align*}
Since $A_p(x) \cap B_{r,d} \subset A_p(x)$, we have $|\tilde{x}'_j - \tilde{x}_j| \leq \overline{a}_p^j(x) - \underline{a}_p^j(x)$ for any $\tilde{x}', \tilde{x} \in A_p(x) \cap B_{r,d}$ and $1 \leq j \leq d$. Consequently, we get
\begin{align}\label{equ::apxbrdmuap}
\biggl| \int _{H(A_p(x) \cap B_{r,d})} (\tilde{x}_j' - \tilde{x}_j) \, d\tilde{x}' \biggr|
& \leq \int_{H(A_p(x) \cap B_{r,d})} |\tilde{x}_j' - \tilde{x}_j| \, d\tilde{x}'
\nonumber\\
& \leq (\overline{a}_p^j(x) - \underline{a}_p^j(x)) \int_{H(A_p(x) \cap B_{r,d})} \, d\tilde{x}'
\nonumber\\
& = \mu(H(A_p(x) \cap B_{r,d})) (\overline{a}_p^j(x) - \underline{a}_p^j(x))
\nonumber\\
& = \mu(A_p(x) \cap B_{r,d}) A_p^j(x),
\end{align}
where we use the fact that an orthogonal transformation is volume-preserving. Combining \eqref{equ::fppxflpx} with \eqref{equ::apxbrdmuap}, we obtain
\begin{align}\label{equ::fppflpsumapj}
\biggl| \int_{A_p(x) \cap B_{r,d}} \nabla f^*_{L,\mathrm{P}}(x)^\top (x' - x) \, dx' \biggr|
\leq  c_L \mu(A_p(x) \cap B_{r,d}) \sum^d_{j=1} A_p^j(x).
\end{align}
Then \eqref{equ::apxbrdmuap} together with \eqref{equ::apxbrd} yields
\begin{align*}
& \frac{1}{c_{r,d} \mathrm{P}_X(A_p(x))} \biggl| \int_{A_p(x) \cap B_{r,d}} \nabla f^*_{L,\mathrm{P}}(x)^\top (x' - x) \, dx' \biggr|
\\
& = \frac{1}{c_{r,d} \mathrm{P}_X(A_p(x))} \biggl| \int_{H(A_p(x) \cap B_{r,d})} \nabla f^*_{L,\mathrm{P}}(x)^\top (x' - x) \, dx' \biggr|
\\
& \leq \frac{c_L \mu(A_p(x) \cap B_{r,d})}{c_{r,d} \mathrm{P}_X(A_p(x))} \sum^d_{j=1} A_p^j(x) 
   = c_L \sum^d_{j=1} A_p^j(x).
\end{align*}
Combining this with \eqref{equ::eprpzfpp1x}, we obtain
\begin{align}\label{equ::leqdsumapj}
\bigl| \mathbb{E}_{\mathrm{P}_{R,Z}}(f_{\mathrm{P}}^{p,1}(x)) - f_{L, \mathrm{P}}^*(x) \bigr|
\leq c_L \mathbb{E}_{\mathrm{P}_{R,Z}} \biggl( \sum^d_{j=1} A_p^j(x) \biggr)
       + \mathbb{E}_{\mathrm{P}_{R,Z}}({\mathrm{diam}(A_p(x))}^{1+\alpha}).
\end{align}
By \eqref{equ::apjr}, we have
\begin{align}\label{equ::sumapj1}
\mathbb{E}_{\mathrm{P}_{R,Z}} \biggl( \sum^d_{j=1} A_p^j(x) \biggr)
= \sum^d_{j=1} \mathbb{E}_{\mathrm{P}_R} \mathbb{E}_{\mathrm{P}_Z}(A_p^j(x) | R)
= 2 r d \biggl( 1 - \frac{1}{2d} \biggr)^p
\leq 2 r d \exp \biggl( - \frac{p}{2d} \biggr).
\end{align}
Moreover, Lemma \ref{lem::diamapx} implies
\begin{align}\label{equ::sumapj2}		
\mathbb{E}_{\mathrm{P}_{R,Z}}({\mathrm{diam}(A_p(x))}^{1+\alpha})
\leq (2r)^{1+\alpha}d \exp \biggl( \frac{(2^{-\alpha-1}-1)p}{d} \biggr).
\end{align}
Combining \eqref{equ::sumapj1}, \eqref{equ::sumapj2} with \eqref{equ::leqdsumapj}, we get
\begin{align}\label{equ::biasfppx}
\bigl| \mathbb{E}_{\mathrm{P}_{R,Z}}(f_{\mathrm{P}}^{p,1}(x)) - f_{L, \mathrm{P}}^*(x) \bigr|
& \leq 2 c_L r d \exp \biggl( - \frac{p}{2d} \biggr) + c_L (2r)^{1+\alpha} d \exp \biggl( \frac{(2^{-\alpha-1}-1)p}{d} \biggr)
\nonumber\\
& \leq 2 c_L (2r)^{\alpha+1} d \exp \biggl( - \frac{p}{2d} \biggr).
\end{align}
Combining \eqref{equ::varfppx}, \eqref{equ::biasfppx} with \eqref{equ::biasvarianceDecom}, we find
\begin{align*}
\mathbb{E}_{\mathrm{P}_{R,Z}}  \bigl( f_{\mathrm{P},\mathrm{E}}(x) - f_{L, \mathrm{P}}^*(x) \bigr)^2 
\leq \frac{c_L^2(2r)^4 d}{T} \exp \biggl( - \frac{0.75p}{d} \biggr) + 4 c_L^2 (2r)^{2\alpha+2} d^2 \exp \biggl( - \frac{p}{d} \biggr).
\end{align*}
Taking expectation with respect to $\mathrm{P}_X$, we get
\begin{align*}
\mathbb{E}_{\mathrm{P}_X} \mathbb{E}_{\mathrm{P}_{R,Z}}  \bigl( f_{\mathrm{P},\mathrm{E}}(x) - f_{L, \mathrm{P}}^*(x) \bigr)^2 
\leq \frac{c_L^2(2r)^4 d}{T} \exp \biggl( - \frac{0.75p}{d} \biggr) + 4 c_L^2 (2r)^{2\alpha+2} d^2 \exp \biggl( - \frac{p}{d} \biggr),
\end{align*}
which leads to the desired assertion by exchanging the order of integration.
\end{proof}

\subsubsection{Proofs Related to the Upper Bound in Section \ref{sec::c1}}

\begin{proof}[of Theorem \ref{thm::optimalForest}]
It is easy to see that $\bar{f}_{\mathrm{P},\mathrm{E}}$ defined by \eqref{fpeensemble} satisfies $\bar{f}_{\mathrm{P},\mathrm{E}}\in \bar{E}$ and $\lambda_1 \|f_{\mathrm{P},\mathrm{E}}\|_{\bar{E}}\leq \lambda_1/T$. Then we have
\begin{align*}
\mathbb{E}_{\mathrm{P}_{R,Z}} \bar{A}(\lambda)
& = \mathbb{E}_{\mathrm{P}_{R,Z}} \biggl( \inf_{f \in \bar{E}} \lambda_1 \|f\|_{\bar{E}} + \lambda_2 \Omega(h) + \mathcal{R}_{L,\mathrm{P}}(f) - \mathcal{R}^*_{L,\mathrm{P}} \biggr)
\\
& \leq \lambda_1 \|f_{\mathrm{P},\mathrm{E}}\|_{\bar{E}} + \lambda_2 \Omega(h)           
+ \mathbb{E}_{\mathrm{P}_{R,Z}} \bigl( \mathcal{R}_{L,\mathrm{P}}(\bar{f}_{\mathrm{P},\mathrm{E}}) - \mathcal{R}^*_{L,\mathrm{P}} \bigr).
\end{align*}
Applying Proposition \ref{prop::biasterm}, we find
\begin{align*}
\mathbb{E}_{\mathrm{P}_{R,Z}} \bar{A}(\lambda)
\leq  \lambda_1/T + \frac{c_L^2(2r)^4 d}{T} \exp \biggl( - \frac{3p}{4d} \biggr) + 4 c_L^2 (2r)^{2\alpha+2} d^2 \exp \biggl( - \frac{p}{d} \biggr).
\end{align*}
This together with Theorem \ref{thm::OracalBoost} implies that with probability at least $1-3/n^2$, there holds
\begin{align*}
& \mathbb{E}_{\mathrm{P}_{R,Z}} \bigl( \mathcal{R}_{L,\mathrm{P}}(f_{\mathrm{D},B}) - \mathcal{R}_{L, \mathrm{P}}^* \bigr)
\\
& \lesssim \frac{\lambda_1}{T} + \lambda_2 \cdot 4^p+ \exp \biggl( - \frac{p}{d} \biggr) + \frac{1}{T} \exp \biggl( - \frac{3p}{4d} \biggr) + T^{2\delta'} \lambda_1^{-2\delta'} \lambda_2^{-1} n^{-2} + 3456 M^2 \biggl( \frac{2\log n}{n} \biggr),
\end{align*}
where $\delta' := 1 - \delta$ and $\delta :=1/(d\cdot 2^p)$. Choosing 
\begin{align*}
\lambda_{1,n} := n^{-\frac{3}{4(2-\delta+d\log 2)}}, \,
\lambda_{2,n} := n^{-\frac{2d\log 2+1}{2-\delta+d\log 2}}, \, 
T_n := n^{\frac{1}{4(2-\delta+d\log 2)}}, \, 
p_n := \frac{d\log n}{2-\delta+d\log 2},
\end{align*}
we obtain that with probability $\mathrm{P}^n$ not less than $1-3/n^2$, there holds
\begin{align*}
\mathbb{E}_{\mathrm{P}_{R,Z}} \bigl( 
\mathcal{R}_{L,\mathrm{P}}(f_{\mathrm{D},B}) - \mathcal{R}_{L,\mathrm{P}}^* \bigr)
\lesssim n^{-\frac{1}{2-\delta+d\log 2}}.
\end{align*}
Consequently, by the Borel-Cantelli Lemma, \eqref{UpperBoundEnsemble} holds with probability $\mathrm{P}^n$ equal to one for sufficiently large $n$. This completes the proof.
\end{proof}

\subsubsection{Proofs Related to Section \ref{sec::c1large}}

\begin{proof}[of Theorem \ref{thm::optimalForestlarge}]
It is easy to see that $\bar{f}_{\mathrm{P},\mathrm{E}}$ defined by \eqref{equ::barfpe} satisfies $\bar{f}_{\mathrm{P},\mathrm{E}}\in \bar{E}$ and $\lambda_1 \|f\bar{f}_{\mathrm{P},\mathrm{E}}\|_{\bar{E}}\leq \lambda_1/T$. Then we have
\begin{align*}
\mathbb{E}_{\mathrm{P}_{R,Z}} \bar{A}(\lambda)
& = \mathbb{E}_{\mathrm{P}_{R,Z}} \biggl( \inf_{f \in \bar{E}} \lambda_1 \|f\|_{\bar{E}} + \lambda_2 \Omega(h) + \mathcal{R}_{L,\mathrm{P}}(f) - \mathcal{R}^*_{L,\mathrm{P}} \biggr)
\\
& \leq \lambda_1 \|f_{\mathrm{P},\mathrm{E}}\|_{\bar{E}} + \lambda_2 \Omega(h)           
+ \mathbb{E}_{\mathrm{P}_{R,Z}} \bigl( \mathcal{R}_{L,\mathrm{P}}(\bar{f}_{\mathrm{P},\mathrm{E}}) - \mathcal{R}^*_{L,\mathrm{P}} \bigr).
\end{align*}
Applying Proposition \ref{prop::biasterm}, we find
\begin{align*}
\mathbb{E}_{\mathrm{P}_{R,Z}} \bar{A}(\lambda)
\leq  \frac{\lambda_1}{T} + \frac{c_L^2(2r)^4 d}{TK} \exp \biggl( - \frac{3p}{4d} \biggr) + 4 c_L^2 (2r)^{2\alpha+2} d^2 \exp \biggl( - \frac{p}{d} \biggr).
\end{align*}
This together with Theorem \ref{equ::OracalBoostlarge} implies that with probability at least $1-3/n^2$, there holds
\begin{align*}
& \mathbb{E}_{\mathrm{P}_{R,Z}} \bigl(
\mathcal{R}_{L,\mathrm{P}}(\bar{f}_{\mathrm{D},B}) - \mathcal{R}_{L, \mathrm{P}}^* \bigr)
\\
& \lesssim \frac{\lambda_1}{T} + \lambda_2 \cdot 4^p + \exp \biggl( - \frac{p}{d} \biggr) + \frac{1}{TK} \exp \biggl( - \frac{3p}{4d} \biggr) + T^{2\delta'} \lambda_1^{-2\delta'} \lambda_2^{-1} n^{-2} + 3456 M^2 \biggl( \frac{2\log n}{n} \biggr),
\end{align*}
where $\delta' := 1 - \delta$ and $\delta :=1/(d\cdot 2^p)$. Choosing 
\begin{align*}
\lambda_{1,n} := n^{-\frac{3}{4(2-\delta+d\log 2)}}, \,
\lambda_{2,n} := n^{-\frac{2d\log 2+1}{2-\delta+d\log 2}},\, 
T_n K_n := n^{\frac{1}{4(2-\delta+d\log 2)}}, \, 
p_n := \frac{d\log n}{2-\delta+d\log 2},
\end{align*}
we obtain that with probability $\mathrm{P}^n$ not less than $1-3/n^2$, there holds
\begin{align*}
\mathbb{E}_{\mathrm{P}_{R,Z}} \bigl(
\mathcal{R}_{L,\mathrm{P}}(\bar{f}_{\mathrm{D},B}) - \mathcal{R}_{L,\mathrm{P}}^* \bigr)
\lesssim n^{-\frac{1}{2-\delta+d\log 2}}.
\end{align*}
Consequently, by the Borel-Cantelli Lemma, \eqref{UpperBoundEnsemblelarge} holds with probability $\mathrm{P}^n$ equal to one for sufficiently large $n$. This completes the proof.
\end{proof}

\subsubsection{Proofs Related to Section \ref{sec::upperc1large}}

\begin{proof}[of Proposition \ref{prop::barbiasterm}]
According to the rotated binary histogram splitting rule, the rotation transformation $\{H_t^k\}$ and split coordinates $\{Z_t^k\}$, $1 \leq t \leq T$, $1 \leq k \leq K$ are i.i.d. Therefore, for any $x \in B_{r,d}$, the expected approximation error term can be decomposed as follows:
\begin{align}
& \mathbb{E}_{\mathrm{P}_{R,Z}}  \bigl( \bar{f}_{\mathrm{P},\mathrm{E}}(x)- f_{L, \mathrm{P}}^*(x) \bigr)^2 
\nonumber\\
& = \mathbb{E}_{\mathrm{P}_{R,Z}} \bigl( (\bar{f}_{\mathrm{P},\mathrm{E}}(x) - \mathbb{E}_{\mathrm{P}_{R,Z}}(\bar{f}_{\mathrm{P},\mathrm{E}}(x)))
+ \mathbb{E}_{\mathrm{P}_{R,Z}}(\bar{f}_{\mathrm{P},\mathrm{E}}(x)) - f_{L,\mathrm{P}}^*(x)) \bigr)^2
\nonumber\\
& =  \mathrm{Var}_{\mathrm{P}_{R,Z}}(\bar{f}_{\mathrm{P},\mathrm{E}}(x))
+ (\mathbb{E}_{\mathrm{P}_{R,Z}}(\bar{f}_{\mathrm{P},\mathrm{E}}(x))-f_{L,\mathrm{P}}^*(x))^2
\nonumber\\
& = \frac{1}{KT} \cdot \mathrm{Var}_{\mathrm{P}_{R,Z}}(f_{\mathrm{P}}^{1,1}(x))
+ \bigl(\mathbb{E}_{\mathrm{P}_{R,Z}}( f_{\mathrm{P}}^{1,1}(x) ) - f_{L, \mathrm{P}}^*(x) \bigr)^2.
\label{equ::biasvarianceDecomlarge}
\end{align}
Combining \eqref{equ::biasvarianceDecomlarge} with \eqref{equ::varfppx} and \eqref{equ::biasfppx}, we find
\begin{align*}
\mathbb{E}_{\mathrm{P}_{R,Z}}  \bigl( f_{\mathrm{P},\mathrm{E}}(x) - f_{L, \mathrm{P}}^*(x) \bigr)^2 
\leq \frac{c_L^2(2r)^4 d}{KT} \exp \biggl( - \frac{0.75p}{d} \biggr) + 4 c_L^2 (2r)^{2\alpha+2} d^2 \exp \biggl( - \frac{p}{d} \biggr).
\end{align*}
Taking expectation with respect to $\mathrm{P}_X$, we get
\begin{align*}
\mathbb{E}_{\mathrm{P}_X} \mathbb{E}_{\mathrm{P}_{R,Z}} \bigl( f_{\mathrm{P},\mathrm{E}}(x) - f_{L, \mathrm{P}}^*(x) \bigr)^2 
\leq \frac{c_L^2(2r)^4 d}{KT} \exp \biggl( - \frac{0.75p}{d} \biggr) + 4 c_L^2 (2r)^{2\alpha+2} d^2 \exp \biggl( - \frac{p}{d} \biggr),
\end{align*}
which leads to the desired assertion by exchanging the order of integration.
\end{proof}

\subsubsection{Proofs Related to Section \ref{sec::lowerboundconve}}

We first show proofs for lower bound of approximation error for binary histogram regression.

\begin{proof}[of Proposition \ref{counterapprox}]
Recall that the regression model is defined as $Y = f(X) + \varepsilon$. Considering the case when $X$ has the uniform distribution, for any $j \in \mathcal{I}_p$ and $x\in A_j$, we have
\begin{align*}
f_{\mathrm{P}}^p(x) 
= \frac{1}{\mathrm{P}_X(A_j)}\int_{A_j} f(x') \, d\mathrm{P}_X(x')
= \frac{1}{\mu(A_j)}\int_{A_j} f(x') \, dx'.
\end{align*}
Since $f(x)\in C^{1,0}$, according to the mean-value theorem, there exists $x^j \in A_j$ such that 
\begin{align*}
f(x^j) 
= \frac{1}{\mu(A_j)} \int_{A_j} f(x') \, dx' 
= f_{\mathrm{P}}^p(x).
\end{align*}
Consequently, we have
\begin{align}\label{equ::fppxfx*}
\mathbb{E}_{\mathrm{P}_X}(f_\mathrm{P}^p(x) - f(x))^2 
& = \frac{1}{(2r)^d} \int_{B_r}(f_\mathrm{P}^p(x) - f(x))^2 \, dx
\nonumber\\
& = \frac{1}{(2r)^d} \sum_{j \in \mathcal{I}_p} \int_{A_j} (f_\mathrm{P}^p(x) - f(x))^2 \, dx
\nonumber\\
& = \frac{1}{(2r)^d} \sum_{j \in \mathcal{I}_p} \int_{A_j} (f(x^j) - f(x))^2 \, dx.
\end{align}

Let $g(t) := f(x^j + t(x - x^j)) - f(x^j)$, $0 \leq t \leq 1$. Since $f(x) \in C^{1,0}$, $g(t)$ is differentiable at every $t \in (0,1)$.  According to Lagrange's mean value theorem, there exists $t^* \in (0, 1)$ such that 
\begin{align*}
g(1) - g(0) 
= g'(t^*)
= \nabla f(x^j + t^*(x - x^j))^{\top} (x - x^j).
\end{align*}
Let $\xi_{j,x}^* := x^j + t^*(x - x^j)$. Then we have
\begin{align*}
(f(x^j) - f(x))^2
= (\nabla f(\xi_{j,x}^*) (x - x^j))^{\top} \nabla f(\xi_{j,x}^*) (x - x^j)
= \|\nabla f(\xi_{j,x}^*)\|^2 \|x - x^j\|^2.
\end{align*}
Since $\|\nabla f\|\geq \underline{c}_f$, we have
\begin{align*}
(f(x^j) - f(x))^2
\geq \underline{c}_f^2 \|x - x^j\|^2.
\end{align*}
Combining this with \eqref{equ::fppxfx*}, we get
\begin{align}\label{equ::epxfppxfx}
\mathbb{E}_{\mathrm{P}_X}(f_\mathrm{P}^p(x) - f(x))^2
& \geq \frac{\underline{c}_f^2}{(2r)^d} \sum_{j \in \mathcal{I}_p} \int_{A_j} \|x - x^j\|^2 \, dx
\nonumber\\
& = \frac{\underline{c}_f^2}{(2r)^d} \sum_{j \in \mathcal{I}_p} \sum^d_{i=1} \int_{A_j} |x_i - x^j_i|^2 \, dx
\nonumber\\
& = \frac{\underline{c}_f^2}{(2r)^d} \sum^d_{i=1} \sum_{j \in \mathcal{I}_p} \int_{A_j} (x_i - x^j_i)^2 \, dx,
\end{align}
where $x_i^j$ denotes the $i$-th entry of the vector $x^j$. Let $\underline{a}_j^i$ and $\overline{a}_j^i$ be the minimum and maximum values of the $i$-th coordinates of points in $A_j$. Since $H(x) = x$ is an identity map, by the construction of the binary histogram partition, we have
\begin{align*}
A_j= [\underline{a}_j^1, \overline{a}_j^1] \times \cdots \times [\underline{a}_j^d, \overline{a}_j^d].
\end{align*}
Moreover, let $h(t) := \int_{A_j} (x_i-t)^2 \, dx$. Then by the iterated integral rule, we have
\begin{align*}
h(t)
& = \prod_{s \neq i} (\overline{a}_j^s - \underline{a}_j^s) \int_{\underline{a}_j^i}^{\overline{a}_j^i} (x_i - t)^2 \, dx_i
\\
& = \prod_{s \neq i} (\overline{a}_j^s - \underline{a}_j^s) \biggl( (\overline{a}_j^i - \underline{a}_j^i) t^2 - 2 t \int_{\underline{a}_j^i}^{\overline{a}_j^i} x_i \, dx_i + \int_{\underline{a}_j^i}^{\overline{a}_j^i} x_i^2 \, dx_i \biggr)
\\
& \geq h \biggl( \frac{\underline{a}_j^i + \overline{a}_j^i}{2} \biggr).
\end{align*}
Consequently, we get
\begin{align*}
\int_{A_j} (x_i - x^j_i)^2 \, dx 
= h(x_i^j)
\geq h \biggl( \frac{\underline{a}_j^i + \overline{a}_j^i}{2} \biggr)
= \int_{A_j} \biggl( x_i - \frac{\underline{a}_j^i+\overline{a}_j^i}{2} \biggr)^2 \, dx.
\end{align*}
This together with \eqref{equ::epxfppxfx} implies
\begin{align*}
\mathbb{E}_{\mathrm{P}_X}(f_\mathrm{P}^p(x) - f(x))^2
& \geq \frac{\underline{c}_f^2}{(2r)^d} \sum^d_{i=1} \sum_{j \in \mathcal{I}_p} \int_{A_j} \biggl( x_i - \frac{\underline{a}_j^i + \overline{a}_j^i}{2} \biggr)^2 \, dx
\\
& = \frac{\underline{c}_f^2}{(2r)^d} \sum^d_{i=1} \sum_{j \in \mathcal{I}_p} \int_{A_j} \biggl( x_i - \frac{\underline{a}_p^i(x) + \overline{a}_p^i(x)}{2} \biggr)^2 \, dx
\\
& = \underline{c}_f^2 \mathbb{E}_{\mathrm{P}_X} \sum^d_{i=1} \biggl( x_i - \frac{\underline{a}_p^i(x) + \overline{a}_p^i(x)}{2} \biggr)^2,
\end{align*}
where $\underline{a}_p^i(x)$ and $\overline{a}_p^i(x)$ are the minimum and maximum values of the $i$-th coordinates of points in $A_p(x)$. Therefore, we obtain
\begin{align}\label{equ::ezepx}
\mathbb{E}_{\mathrm{P}_Z} \mathbb{E}_{\mathrm{P}_X}(f_\mathrm{P}^p(x) - f(x))^2
\geq \underline{c}_f^2 \mathbb{E}_{\mathrm{P}_X} \sum^d_{i=1} \mathbb{E}_{\mathrm{P}_Z} \biggl( x_i - \frac{\underline{a}_p^i(x) + \overline{a}_p^i(x)}{2} \biggr)^2.
\end{align}
Let $S_p^i(x)$ be the number of times that $A_p(x)$ is split on the $i$-th coordinate. According to Lemma \ref{lem::basic2}, if $S_p^i(x)=k$, $0\leq k\leq q$, then we have
\begin{align*}
\biggl| x_i - \frac{\underline{a}_p^i(x) + \overline{a}_p^i(x)}{2} \biggr|
= \min_{q \in Q_k} |x_i - q|,
\end{align*}
where
\begin{align*}
Q_k = \biggl\{ \frac{r(2j-1)}{2^k}  \, \bigg| \, -2^{k-1}+1\leq j\leq 2^{k-1} \biggr\}.
\end{align*}
Therefore, we obtain
\begin{align*}
\mathbb{E}_{\mathrm{P}_Z} \biggl( x_i - \frac{\underline{a}_p^i(x) + \overline{a}_p^i(x)}{2} \biggr)^2
= \sum^p_{k=0} \mathrm{P}_{Z}(S_p^i(x) = k) \min_{q \in Q_k} (x_i - q)^2.
\end{align*}
This together with \eqref{equ::ezepx} implies
\begin{align}\label{equ::epzpxlower}
\mathbb{E}_{\mathrm{P}_Z} \mathbb{E}_{\mathrm{P}_X}(f_\mathrm{P}^p(x)-f(x))^2
& \geq \underline{c}_f^2 \mathbb{E}_{\mathrm{P}_X} \sum^d_{i=1} \sum^p_{k=0} \mathrm{P}_{Z}(S_p^i(x)=k) \min_{q \in Q_k} (x_i-q)^2 
\nonumber\\
& = \frac{\underline{c}_f^2}{(2r)^d} \int_{B_r} \sum^d_{i=1} \sum^p_{k=0} \mathrm{P}_Z(S_p^i(x)=k) \min_{q \in Q_k} (x_i - q)^2 \, dx
\nonumber\\
& = \frac{\underline{c}_f^2}{(2r)^d} \sum^d_{i=1} \biggl( \sum^p_{k=0} f(k,p,1/d) \int_{B_r} \min_{q \in Q_k} (x_i - q)^2 \, dx \biggr),
\end{align}
where $f(k,p,1/d) = \binom{p}{k} (\frac{1}{d})^k (1 - \frac{1}{d})^{n-k}$. By the definition of $Q_k$, we have
\begin{align*}
\int_{B_r} \min_{q \in Q_k} (x_i - q)^2 \, dx
& = (2r)^{d-1} \int_{-r}^r \min_{q \in Q_k} (x_i - q)^2 \, d x_i
\\
& = (2r)^{d-1} \cdot 2^{k+1} \int_{r-r/2^k}^r (x_i - (r - r/2^k))^2 \, dx_i
\\
& = \frac{3(2r)^{d-1}}{2} \cdot \frac{r^3}{2^{2k}}.
\end{align*}
This together with \eqref{equ::epzpxlower} implies
\begin{align*}
\mathbb{E}_{\mathrm{P}_Z} \mathbb{E}_{\mathrm{P}_X}(f_\mathrm{P}^p(x)-f(x))^2
\geq \frac{\underline{c}_f^2}{(2r)^d} \sum^d_{i=1} \biggl( \sum^p_{k=0} 2^{-2k} \cdot f(k,p,1/d) \biggr)
= \frac{3 \underline{c}_f^2 r^2d}{4} \biggl( 1 - \frac{3}{4d} \biggr)^p,
\end{align*}
which completes the proof.
\end{proof}

Then we present proofs for lower bound of sample error for binary histogram regression.

\begin{proof}[of Proposition \ref{counterapprox2}]
Since $H(x)=x$ is a identity map,
for any fixed split coordinates $Z=\{Z_{i,j},1\leq i\leq p, 1\leq j\leq 2^{i-1}\}$, $\{A_j\}_{j\in \mathcal{I}_p}$ forms a partition of $B_r$, then for $j \in \mathcal{I}_p$ we define the random variable $N_j$ by
\begin{align*}
N_j := \sum_{i=1}^n \eins_{A_j}(X_i).
\end{align*} 
Since the random variables $\{ \eins_{A_j}(X_i) \}_{i=1}^n$ are i.i.d.~Bernoulli distributed with parameter $\mathrm{P}_X(x\in A_j)$, it is clear to see that the random variable $N_j$ is Binomial distributed with parameters $n$ and $\mathrm{P}_X(x\in A_j)$. Therefore, for any $j \in \mathcal{I}_p$, we have
\begin{align*}
\mathbb{E}(N_j) = n \cdot \mathrm{P}_X(x\in A_j). 
\end{align*}
Moreover, the GBBH regressor $f_{\mathrm{D}}^p$ can be defined by
\begin{align*}
f_{\mathrm{D}}^p(x) =
\begin{cases}
\displaystyle \frac{\sum_{i=1}^n Y_i \eins_{A_j}(X_i)}{\sum_{i=1}^n \eins_{A_j}(X_i)} \cdot \eins_{A_j}(x) & \text{ if } N_j > 0,
\\
0 & \text{ if } N_j = 0.
\end{cases}
\end{align*}
By the law of total probability, we get 
\begin{align}
& \mathbb{E}_{D \sim \mathrm{P}^n} \mathbb{E}_{\mathrm{P}_X} \bigl( f_{\mathrm{D}}^p(x) - f_{\mathrm{P}}^p(x) \bigr)^2
\nonumber\\
& = \mathbb{E}_{D \sim \mathrm{P}^n} \biggl( \sum_{j \in \mathcal{I}_p} \mathbb{E}
\bigl( \bigl( f_{\mathrm{D}}^p(x) - f_{\mathrm{P}}^p(x) \bigr)^2 \big| x \in A_j \bigr)
\cdot \mathrm{P}_X(x\in A_j)\biggr)
\nonumber\\
& = \mathbb{E}_{D\sim \mathrm{P}^n}\biggl( \sum_{j \in \mathcal{I}_p} \mathbb{E}
\bigl( \bigl( f_{\mathrm{D}}^p(x) - f_{\mathrm{P}}^p(x) \bigr)^2 \big| x \in A_j, N_j > 0 \bigr)
\cdot \mathrm{P}(N_j > 0) \cdot \mathrm{P}_X(x\in A_j)\biggr)
\label{equ::term1}
\\
& \phantom{=}
+ \mathbb{E}_{D\sim \mathrm{P}^n}\biggl(\sum_{j \in \mathcal{I}_p} \mathbb{E}
\bigl( \bigl( f_{\mathrm{D}}^p(x) - f_{\mathrm{P}}^p(x) \bigr)^2 \big| x \in A_j, N_j = 0 \bigr)
\cdot \mathrm{P}(N_j = 0) \cdot \mathrm{P}_X(x\in A_j)\biggr).
\label{equ::term2}
\end{align}
For the term \eqref{equ::term1}, we have
\begin{align*}
& \sum_{j \in \mathcal{I}_p} \mathbb{E}\bigl( (f_{\mathrm{D}}^p(x) - f_{\mathrm{P}}^p(x))^2 \big| X \in A_j, N_j > 0 \bigr)
\cdot \mathrm{P}(N_j > 0) \mathrm{P}_X(x\in A_j)
\\
& = \sum_{j \in \mathcal{I}_p} \biggl( \frac{\sum_{i=1}^n Y_i \eins_{A_j}(X_i)}{\sum_{i=1}^n \eins_{A_j}(X_i)}
- \mathbb{E}(f_{L,\mathrm{P}}^*(X) | X \in A_j) \biggr)^2
\cdot \mathrm{P}(N_j > 0)\mathrm{P}_X(x\in A_j)
\\
& = \sum_{j \in \mathcal{I}_p} \biggl( \sum_{i=1}^n \eins_{A_j}(X_i) \bigl( Y_i - \mathbb{E}(f_{L,\mathrm{P}}^*(X) | X \in A_j) \bigr) \biggr)^2
\cdot \frac{\mathrm{P}_X(x\in A_j)}{(\sum_{i=1}^n \eins_{A_j}(X_i))^2} \cdot \mathrm{P}(N_j > 0).
\end{align*}
Since for a fixed $j \in \mathcal{I}_p$, there holds
\begin{align}
& \sum_{j \in \mathcal{I}_p} 
\biggl( \sum_{i=1}^n \eins_{A_j}(X_i) \bigl( Y_i - \mathbb{E}(f_{L,\mathrm{P}}^*(X) | X \in A_j) \bigr) \biggr)^2
\cdot \frac{\mathrm{P}_X(x\in A_j)}{(\sum_{i=1}^n \eins_{A_j}(X_i))^2}
\nonumber \\
& = \sum_{j \in \mathcal{I}_p} \sum_{i=1}^n \eins_{A_j}^2(X_i) \mathbb{E} \bigl( \bigl( Y - f_{\mathrm{P}}^p(X) \bigr)^2 \big| X \in A_j \bigr)
\cdot \frac{\mathrm{P}_X(x\in A_j)}{(\sum_{i=1}^n \eins_{A_j}(X_i))^2}  
\nonumber \\
& = \sum_{j \in \mathcal{I}_p} \frac{\mathrm{P}_X(x\in A_j)}{\sum_{i=1}^n \eins_{A_j}(X_i)} 
\cdot \mathbb{E} \bigl( \bigl( Y - f_{\mathrm{P}}^p(X))^2 \big| X \in A_j \bigr).
\label{ConditionalExpectationOnAj}
\end{align} 
Moreover, for any fixed $j \in \mathcal{I}_p$, there holds
\begin{align*}
\mathbb{E}(f_{\mathrm{P}}^p(X) | X \in A_j) = \mathbb{E}(f_{L,\mathrm{P}}^*(X) | X \in A_j). 
\end{align*}
Consequently, we obtain
\begin{align*}
& \mathbb{E} \bigl( (Y - f_{\mathrm{P}}^p(X))^2 \big| X \in A_j \bigr)
\\
& = \mathbb{E} \bigl( (Y - f_{L,\mathrm{P}}^*(X))^2 \big| X \in A_j \bigr)
       + \mathbb{E} \bigl( (f_{L,\mathrm{P}}^*(X) - f_{\mathrm{P}}^p(X))^2 \big| X \in A_j \bigr)
\\
& = \sigma^2 + \mathbb{E} \bigl( (f_{L,\mathrm{P}}^*(X) - f_{\mathrm{P}}^p(X))^2 \big| X \in A_j \bigr).
\end{align*}
Therefore, we get
\begin{align*}
& \mathbb{E}_{D \sim \mathrm{P}^n}\biggl( \sum_{j \in \mathcal{I}_p} \mathbb{E}
\bigl( \bigl( f_{\mathrm{D}}^p(x) - f_{\mathrm{P}}^p(x) \bigr)^2 \big| x \in A_j, N_j > 0 \bigr)
\cdot \mathrm{P}(N_j > 0) \cdot \mathrm{P}_X(x\in A_j)\biggr)
\\
& = \bigl( \sigma^2 + \mathbb{E}(f_{L,\mathrm{P}}^*(X) - f_{\mathrm{P}}^p(X))^2 \bigr)
\\
& \phantom{=} \cdot \sum_{j \in \mathcal{I}_p} \biggl( \mathrm{P}_X(x\in A_j)
\mathbb{E}_{D \sim \mathrm{P}^n} \biggl( \biggl( \sum_{i=1}^n \eins_{A_j}(X_i) \biggr)^{-1} \bigg| N_j > 0 \biggr) \biggr)
\cdot \mathrm{P}(N_j > 0)
\\
& = \bigl( \sigma^2 + \mathbb{E}(f_{L,\mathrm{P}}^*(X) - f_{\mathrm{P}}^p(X))^2 \bigr)
\cdot \sum_{j \in \mathcal{I}_p} \bigl( \mathrm{P}_X(x\in A_j) \mathbb{E}_{D \sim \mathrm{P}^n} (N_j^{-1} | N_j > 0) \bigr)
\mathrm{P}(N_j > 0)
\\
& = n^{-1} \bigl( \sigma^2 + \mathbb{E}(f_{L,\mathrm{P}}^*(X) - f_{\mathrm{P}}^p(X))^2 \bigr)
\cdot \sum_{j \in \mathcal{I}_p}  \bigl( \mathbb{E}(N_j) \cdot \mathbb{E}(N_j^{-1} | N_j > 0) \bigr) \mathrm{P}(N_j > 0).
\end{align*}
Clearly, $x^{-1}$ is convex for $x > 0$. Therefore, by Jensen's inequality, we get
\begin{align*}
\mathbb{E}(N_j) \cdot \mathbb{E}(N_j^{-1} | N_j > 0) \mathrm{P}(N_j > 0)
& \geq \mathbb{E}(N_j) \cdot \mathbb{E}(N_j | N_j > 0)^{-1} \mathrm{P}(N_j > 0)
\\
& = \mathbb{E}(N_j) \cdot \mathbb{E}(N_j \eins_{\{N_j > 0\}})^{-1} \mathrm{P}(N_j > 0) \mathrm{P}(N_j > 0)
\\
& = \mathrm{P}(N_j > 0)^2 = (1 - \mathrm{P}(N_j=0))^2
\\
& = \bigl( 1 - (1 - \mathrm{P}_X(x\in A_j))^n \bigr)^2
\\
& \geq 1 - 2 e^{- n \mathrm{P}_X(x\in A_j)},
\end{align*}
where the last inequality follows from $(1 - x)^n \leq e^{-nx}$, $x \in (0, 1)$.
Therefore, we have
\begin{align}\label{equ::lower1}
&\mathbb{E}_{D \sim \mathrm{P}^n}\biggl( \sum_{j \in \mathcal{I}_p} \mathbb{E}
\bigl( \bigl( f_{\mathrm{D}}^p(x) - f_{\mathrm{P}}^p(x) \bigr)^2 \big| x \in A_j, N_j > 0 \bigr)
\cdot \mathrm{P}(N_j > 0) \cdot \mathrm{P}_X(x\in A_j)\biggr)\nonumber\\
&\geq \frac{\sigma^2}{n}\sum_{j\in \mathcal{I}_p} (1 - 2 e^{- n \mathrm{P}_X(x\in A_j)})=\frac{\sigma^2}{n}\sum_{j\in \mathcal{I}_p} (1 - 2 e^{- n/ 2^{p}})=\frac{\sigma^2\cdot 2^p}{n}(1-2e^{-n/2^p}),
\end{align}
where we use the fact that $\mu(A_j)=(2r)^d/2^p$ for $j \in \mathcal{I}_p$ by \eqref{equ::muapx} and $\mathrm{P}_X$ is the uniform distribution on $B_r$.

We now turn to estimate the term \eqref{equ::term2}. By the definition of $f_{\mathrm{D}}^p$, we have
\begin{align*}
& \sum_{j \in \mathcal{I}_p} \mathbb{E}
\bigl( \bigl( f_{\mathrm{D}}^p(X) - f_{\mathrm{P}}^p(X))^2 \big| X \in A_j, N_j = 0 \bigr)
\cdot \mathrm{P}(N_j = 0) \cdot \mathrm{P}_X(x\in A_j)
\\
& = (2r)^{-d}\sum_{j\in \mathcal{I}_p}\biggl(\eins_{\{N_j=0\}} \int_{A_j}  f_{\mathrm{P}}^p(x)^2 \,  dx \biggr). 
\end{align*}
Since for $x\in A_j$, we have
\begin{align*}
f_{\mathrm{P}}^p(x)=\mu(A_j)^{-1}\int_{A_j} f^*_{L,\mathrm{P}}(x)dx,
\end{align*}
consequently, we obtain
\begin{align*}
\sum_{j\in \mathcal{I}_p}\biggl(\eins_{\{N_j=0\}} \int_{A_j}  f_{\mathrm{P}}^p(x)^2 \,  dx\biggr)=\sum_{j\in \mathcal{I}_p}\frac{\eins_{\{N_j=0\}}}{\mu(A_j)} \biggl(\int_{A_j}  f^*_{L,\mathrm{P}}(x)dx\biggr)^2.
\end{align*}
Together with \eqref{equ::muapx} in Fact \ref{fact2}, we find
\begin{align*}
(2r)^{-d} \sum_{j \in \mathcal{I}_p} \biggl( \eins_{\{N_j=0\}} \int_{A_j}  f_{\mathrm{P}}^p(x)^2 \, dx\biggr)
= 2^p \sum_{j \in \mathcal{I}_p} \biggl( \biggl( \int_{A_j}  f^*_{L,\mathrm{P}}(x) \, dx\biggr)^2 \cdot \eins_{\{ N_j = 0 \}} \biggr).
\end{align*}
Therefore, we get
\begin{align}\label{equ::lower2}
& \mathbb{E}_{D \sim \mathrm{P}^n} \biggl( \sum_{j \in \mathcal{I}_p} \mathbb{E}
\bigl( \bigl( f_{\mathrm{D}}^p(x) - f_{\mathrm{P}}^p(x) \bigr)^2 \big| x \in A_j, N_j = 0 \bigr)
\cdot \mathrm{P}(N_j = 0) \cdot \mathrm{P}_X(x\in A_j) \biggr)
\nonumber\\
& = 2^p \cdot \mathbb{E}_{D \sim \mathrm{P}^n} \sum_{j \in \mathcal{I}_p} \biggl( \biggl( \int_{A_j}  f^*_{L,\mathrm{P}}(x) \, dx \biggr)^2 \cdot \eins_{\{ N_j = 0 \}} \biggr)
\nonumber\\
& = 2^p \cdot \sum_{j \in \mathcal{I}_p} \biggl( \mathrm{P}(N_j = 0) \biggl( \int_{A_j} f^*_{L,\mathrm{P}}(x) \, dx \biggr)^2 \biggr)
\nonumber\\
& = \biggl( 1 - \frac{1}{2^p} \biggr)^n \biggl( 2^p \sum_{j \in \mathcal{I}_p} \biggl( \int_{A_j} f^*_{L,\mathrm{P}}(x) \, dx \biggr)^2 \biggr)
\nonumber\\
& \geq \biggl( 1 - \frac{1}{2^p} \biggr)^n \biggl( \sum_{j \in \mathcal{I}_p} \int_{A_j} f^*_{L,\mathrm{P}}(x) \, dx \biggr)^2 
   \geq (2r)^{2d} \underline{c}_f^2 \biggl( 1 - \frac{1}{2^p} \biggr)^n.
\end{align}
Combining \eqref{equ::lower1} and \eqref{equ::lower2}, we obtain
\begin{align*}
\mathbb{E}_{D \sim \mathrm{P}^n} \mathbb{E}_{\mathrm{P}_X} \bigl( f_{\mathrm{D}}^p(X) - f_{\mathrm{P}}^p(X) \bigr)^2
\geq \frac{\sigma^2 \cdot 2^p}{n} \biggl( 1 - 2 \exp \biggl( - \frac{n}{2^p} \biggr) \biggr) + (2r)^{2d} \underline{c}_f^2 \biggl( 1 - \frac{1}{2^p} \biggr)^n.
\end{align*} 
Taking expectation with respect to $\mathrm{P}_Z$, we prove the desired assertion.
\end{proof}

\subsubsection{Proofs Related to the Lower Bound in Section \ref{sec::c1}}

\begin{proof}[of Theorem \ref{prop::counter}]
Recall the error decomposition \eqref{equ::L2Decomposition}. Applying Propositions \ref{counterapprox} and \ref{counterapprox2}, we get 
\begin{align*} 
& \mathbb{E}_{\mathrm{P}^n\otimes \mathrm{P}_Z} \bigl( \mathcal{R}_{L,\mathrm{P}}(f_{\mathrm{D}}) - \mathcal{R}_{L,\mathrm{P}}^* \bigr)
\\
& = \mathbb{E}_{\mathrm{P}^n\otimes \mathrm{P}_Z} \mathbb{E}_{\mathrm{P}_X} \bigl( f_{\mathrm{D}}(X) - f_{L,\mathrm{P}}^*(X) \bigr)^2 
\\
& \geq \frac{3 \underline{c}_f^2 r^2 d}{4} \biggl( 1 - \frac{3}{4d} \biggr)^p + \frac{\sigma^2 \cdot 2^p}{n} \biggl( 1 - 2 \exp \biggl( - \frac{n}{2^p} \biggr) \biggr) + (2r)^{2d} \underline{c}_f^2 \biggl( 1 - \frac{1}{2^p} \biggr)^n.
\end{align*}
If $\exp(-n/2^p)\leq 1/4$, we have
\begin{align}\label{equ::lowerbound1}
\mathbb{E}_{\mathrm{P}^n \otimes \mathrm{P}_Z} \bigl( \mathcal{R}_{L,\mathrm{P}}(f_{\mathrm{D}}) - \mathcal{R}_{L,\mathrm{P}}^* \bigr)
\geq \frac{3 \underline{c}_f^2 r^2 d}{4} \biggl( 1 - \frac{3}{4d} \biggr)^p + \frac{\sigma^2 \cdot 2^p}{2n}
\geq c_0 n^{\frac{\log (1-0.75/d)}{\log 2-\log(1-0.75/d)}},
\end{align}
where the constant $c_0 := (\sigma^2/2) \cdot (3 \underline{c}_f^2r^2d/(2\sigma^2))^{\frac{\log 2}{\log 2-\log(1-0.75/d)}}$. On the other hand, we turn to the case $\exp(-n/2^p)> 1/4$. Let $g(x):=\log(1-x)+2\log 2x$, $x\leq 1/2$, we find $g'(x)=-1/(1-x)+2\log 2\leq -1-2\log 2$. It is easy to see that $g'(x)>0$ when $0\leq x< 1-1/(2\log 2)$ and $g'(x)<0$ when $1-1/(2\log 2)<x<1/2$, which yields that $g(x)$ is increasing on the interval $(0,1-1/(2\log 2))$ and decreasing on the interval $(1-1/(2\log 2),1/2)$. Therefore,
for $0<x\leq 1/2$, there holds
\begin{align*}
g(x) = \log(1-x) + 2 \log 2x 
\geq \min \{ g(0), g(1/2) \}
= 0.
\end{align*}
Consequently, we get
\begin{align}\label{equ::lowerbound2}
\mathbb{E}_{\mathrm{P}^n \otimes \mathrm{P}_Z} \bigl( \mathcal{R}_{L,\mathrm{P}}(f_{\mathrm{D}}) - \mathcal{R}_{L,\mathrm{P}}^* \bigr)
& \geq (2r)^{2d} \underline{c}_f^2 (1 - 1/2^p)^n	
\nonumber\\
& = (2r)^{2d} \underline{c}_f^2 \exp \bigl( n \log (1 - 1/2^p) \bigr)
\nonumber\\
& \geq (2r)^{2d} \underline{c}_f^2 \exp \bigl( - (2 n \log 2) / 2^p \bigr)
\nonumber\\
& \geq (2r)^{2d} \underline{c}_f^2 (1/4)^{2\log 2} := c_1.
\end{align}
Combining \eqref{equ::lowerbound1} with \eqref{equ::lowerbound2}, we find
\begin{align*}
\mathbb{E}_{\mathrm{P}^n \otimes \mathrm{P}_Z} \bigl( 
\mathcal{R}_{L,\mathrm{P}}(f_{\mathrm{D}}) - R^*_{L,\mathrm{P}} \bigr)
\geq c_0 n^{\frac{\log (1-0.75/d)}{\log 2-\log(1-0.75/d)}}\vee c_1,
\end{align*}
which leads to the desired assertion.
\end{proof}

\bibliography{GBBHE}

\end{document}